\documentclass[conf]{new-aiaa}
\usepackage[utf8]{inputenc}

\usepackage{graphicx}
\usepackage{amsmath}
\usepackage{booktabs}
\usepackage[version=4]{mhchem}
\usepackage{siunitx}
\usepackage{longtable,tabularx}
\usepackage{booktabs}
\usepackage{subcaption}
\usepackage{float}
\usepackage{multirow} % For \multirow
\usepackage{multicol} 

\title{Bird Movement Prediction Using Long Short-Term Memory Networks to Prevent Bird Strikes with Low Altitude Aircraft}

\author{Elaheh Sabziyan Varnousfaderani \footnote{Graduate Research Assistant, College of Aeronautics and Engineering, esabziya@kent.edu} and Syed A. M. Shihab\footnote{Assistant Professor, College of Aeronautics and Engineering, AIAA Member, sshihab@kent.edu}}

\begin{document}

\maketitle

\begin{abstract}
The number of collisions between aircraft and birds in the airspace has been increasing at an alarming rate over the past decade due to increasing bird population, air traffic and usage of quieter aircraft. Bird strikes with aircraft are anticipated to further increase dramatically in the near future when emerging Advanced Air Mobility aircraft start operating in the low altitude airspace where probability of bird strikes is the highest. Not only do such bird strikes can result in human and bird fatalities, but they also cost the aviation industry millions of dollars in damages to aircraft annually. To better understand the causes and effects of bird strikes, research to date has mainly focused on analyzing factors which increase the probability of bird strikes, identifying high risk birds in different locations, predicting the future number of bird strike incidents, and estimating cost of bird strike damages. However, research on bird movement prediction for use in flight planning algorithms to minimize the probability of bird strikes is very limited. To address this gap in research, we implement four different types of Long Short-Term Memory (LSTM) models, a type of recurrent neural network capable of learning order dependence in sequence prediction problems, to predict bird movement latitudes and longitudes. A publicly available data set on movement of pigeons, one of the top ten high risk birds with respect to bird strikes in aviation, is utilized to train the models and evaluate their performances. Using the bird flight track predictions, aircraft departures from Cleveland Hopkins airport are simulated to be delayed by varying amounts to avoid potential bird strikes with aircraft during takeoff. Results demonstrate that the LSTM models can predict bird movement with high accuracy, achieving a Mean Absolute Error of less than 100 meters, outperforming linear and nonlinear regression models. Our findings indicate that incorporating bird movement prediction into flight planning can be highly beneficial. Specifically, by implementing a modest delay in aircraft departures based on predicted bird movements, the occurrence of bird strikes during takeoff can be prevented.

\end{abstract}

\section{Introduction}
\label{sec:Introduction}

\subsection{Bird Strike Challenge}

\lettrine{C}{ollisions} between birds and aircraft have been occurring since the beginning of aviation. Such collisions, which can happen either during the take-off or landing roll or the cruise flight of the aircraft, are referred to as \textit{bird strikes} \cite{BS4}. Bird strikes threaten the safety of flight operations, putting the lives of both passengers onboard the aircraft and birds at risk, and causes financial loss to the aviation industry in the form of millions of dollars in damages to aircraft annually.  

Bird strikes have been the cause of many nonfatal and fatal accidents in aviation. The first reported nonfatal bird strike, as recorded by Orville Wright in his diary, occurred in 1905 when his aircraft, Wright Flyer III, hit a bird as he flew over a cornfield near Dayton, Ohio, USA \cite{dolbeer2013history}. On April 3, 1912, the first bird strike which caused human fatality occurred when Calbraith Rodgers's aircraft struck a gull while flying along the coast of southern California \cite{cleary2005wildlife}. In recent years, three different accidents further demonstrated the catastrophic nature of bird strikes in aviation. On January 15, 2009, US Airways Flight 1549 (Airbus 320) carrying 155 passengers aboard struck a flock of geese in the aircraft's engine shortly after take-off from New York City on its way to Charlotte in North Carolina, losing all engine power. Unable to reach any nearby airports for an emergency landing due to their low altitude, the pilots were forced to glide the plane to a ditching in the Hudson River off Midtown Manhattan \cite{BS2-2}. Three years later, on September 28, 2012, 16 passengers and three crew members were killed when a Dornier 228 crashed after striking a black kite (Milvus migrans) on take-off from Kathmandu, Nepal \cite{dennisbird}. More recently, Ural Airlines Flight 178 (an Airbus A321) carrying 226 passengers and seven crew members had to make a forced landing in a cornfield three miles from Zhukovsky International Airport on August 15, 2019 in Moscow, Russia after ingesting a flock of gulls into both engines during takeoff, which resulted in insufficient thrust to maintain flight \cite{BS3}.

 Collisions between birds and aircraft are a serious issue that can have significant financial damages for aviation every year. The financial costs arise due to expenses related to repairs, replacement of damaged aircraft components, and the delays or cancellations of flights caused by bird strikes. According to \cite{costs}, the cost of damages as well as delays applied is approximately US\$1.2 billion per year globally. Total accumulated cost of bird strikes during a 32-year period from 1990 to 2020 in the US has been reported to be US\$871,453,177 \cite{dolbeer2021wildlife}.
 
\subsection{Motivation}
\begin{figure}[b]
\centering
\includegraphics[width=16cm, height = 6 cm]{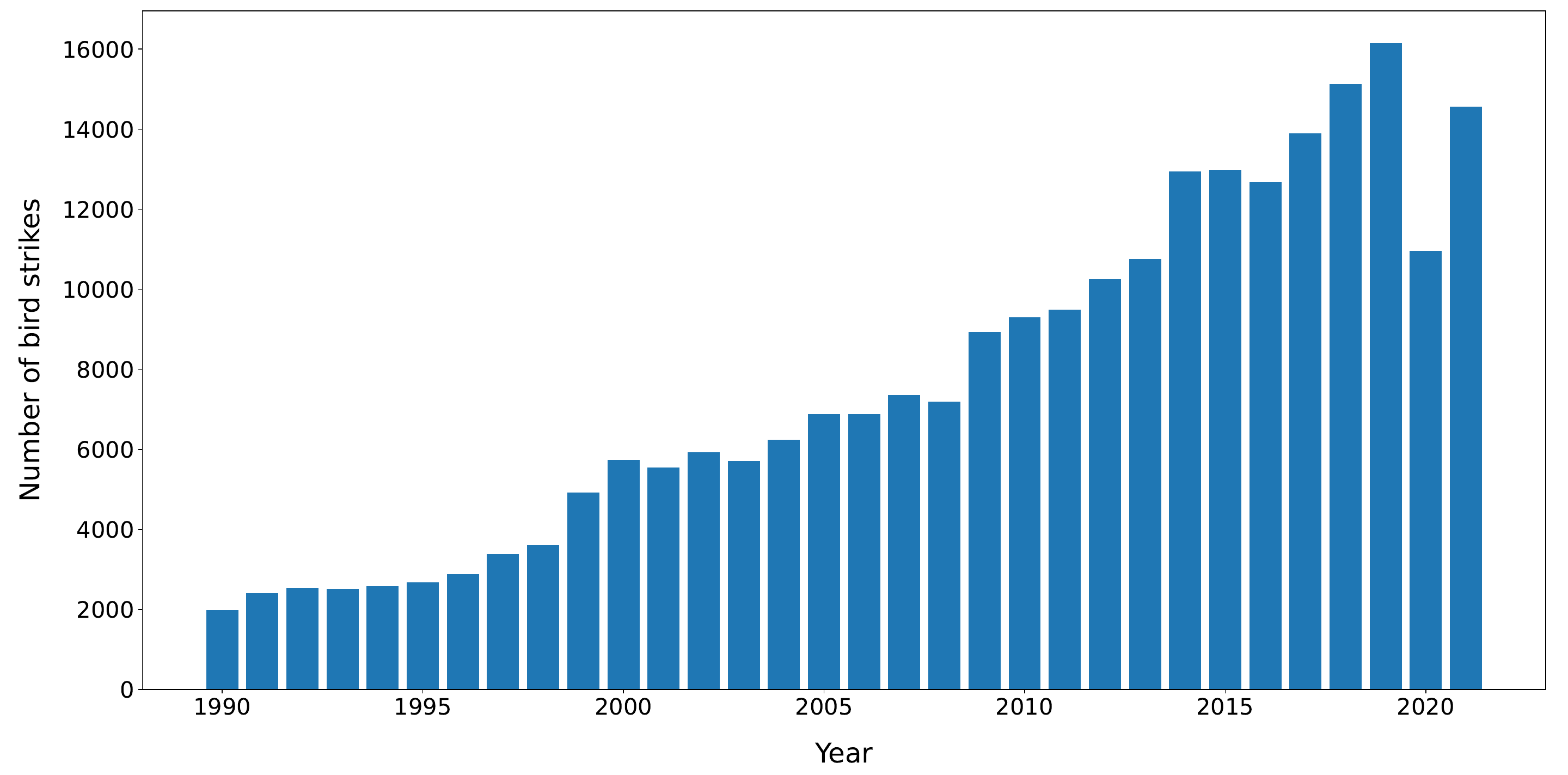}
\caption{Total number of annual bird strikes with civil aircraft in the United States (adapted from \cite{dolbeer2021wildlife})}
\label{Total annual}
\end{figure}

In recent times, tens of thousands of bird strikes with civil aircraft are reported annually by pilots, airports, and airlines in the United States \cite{dolbeer2021wildlife} to the FAA wildlife strike database, as depicted in Fig. \ref{Total annual}. An alarmingly increasing trend of bird strikes can be observed in the bar chart, which can be attributed to increasing bird population, air traffic and usage of quieter aircraft \cite{Robin}. As shown in the Fig. \ref{Total annual}, the number of bird strikes was trending upwards before 2020. But in 2020 it went down a lot because of COVID-19  which led to widespread cancellations of flights and a significant reduction in air traffic.  This may have resulted in fewer bird strikes occurring. In 2021, approximately 15000 bird strikes occurred with civil aircraft in the United States. These numbers suggest that bird strikes are a growing threat to aviation safety. 
\floatplacement{figure}{!b}

Most bird strike accidents share a common factor: they occur during aircraft takeoff when the aircraft is close to the ground. According to statistics and research, the highest risk or likelihood of occurrence of bird strikes in aviation is during the takeoff and landing phases of the flight when the aircraft is below 1000 m \cite{mckee2016approaches}. The research also shows that as altitude increases, the probability of bird strikes decreases. The severity of damage from bird strikes increases at higher altitudes, where aircraft have higher velocities \cite{dolbeer2011increasing}, as well as during takeoff when engines run at full power \cite{avrenli2014statistical}. It is anticipated that highly dense Advanced Air Mobility (AAM) operations will take place in the airspace below 400 m in the near future, where potential bird strikes will pose a severe challenge for the safety of AAM operations. Despite the safety and financial concerns surrounding bird strikes, the problem of preventing bird strikes with emerging low altitude AAM aircraft has not received sufficient attention from the AAM community thus far.

Previous research on bird strikes has mostly focused on identifying high risk birds in different locations; analyzing factors affecting the risk of bird strikes \cite{ning2014bird, metz2020bird} ; cost estimation of damages caused by bird strikes \cite{costs}; and predicting the future number of bird strike incidents in a given area \cite{dolbeer2021wildlife}. While previous research helps us to better understand the characteristics of bird strikes, it does little to minimize the probability of bird strike occurrences with aircraft \cite{mckee2016approaches}. Bird strike probability can be minimized by predicting future bird movement based on historical and real time bird movement detected and tracked by surveillance sensors such as avian radars and weather radars \cite{metz2020bird}. These predictions can then be taken into account in flight planning so that the space-time or 4D trajectory of the aircraft does not intersect with that of the birds during its flight. So far in the literature, only a simple linear regression model was developed by \cite{metz2021air} to predict bird movement. The model was reported to not have sufficient levels of accuracy for flight planning purposes in the real world. 

\subsection{Summary of Contributions}

To address the problem of bird strikes in aviation, the present study proposes the use of an AI-based bird movement predictive model to forecast bird movement with high accuracy. Such forecasts can then be used in flight planning to prevent collisions of aircraft with birds \cite{metz2021air, metz2021analysis}. More specifically, LSTM models are considered as suitable predictive models for this purpose. They are a special kind of artificial recurrent neural networks, which are capable of learning long term dependencies in sequential data, such as those present in bird movement data. They have been found to have high prediction accuracy in numerous practical applications involving, for example, trajectory prediction of moving objects \cite{Moving-objects}; language modeling \cite{Language-modeling}; natural language processing \cite{natural-l-processing}; time series forecasting to predict stock prices \cite{stock-market} and weather patterns \cite{weather-pattern}; image and video analysis for object recognition, tracking, and captioning \cite{video-captioning}; and autonomous vehicle control and navigation \cite{autonomous}. To implement and test the proposed approach, the study makes the following contributions: 

\begin{enumerate}
    \item Four different types of LSTM models have been developed and tested using real-world bird movement data of pigeons, which is one of the top ten high risk bird species in the context of bird strikes. The different LSTM models are vanilla LSTM, stacked LSTM, bidirectional LSTM and encoder-decoder LSTM. Their performances were benchmarked against linear and nonlinear regression models. Results show that the accuracy levels of LSTM models in forecasting bird movement are significantly higher than those of linear and nonlinear regression models. The LSTM models achieved a Mean Absolute Error (MAE) of less than 100 meters, while the MAE for linear and nonlinear regression models were much higher, indicating their insufficiency for predicting bird movement. Specifically, the MAE achieved by the linear regression model in predicting latitude was over 5600 meters, and in predicting longitude it was over 796 meters. The nonlinear regression model performed even worse than the linear regression model, achieving an MAE higher than 26281 and 1877 meters for latitude and longitude prediction, respectively.
    
    \item Several practical insights related to training, testing, and deploying the LSTM models in practice have been generated. The first observation is that the training data must be representative of bird movement behavior, so that the trained model can perform accurately during different test data sets. Models were tested on various test data sets to demonstrate that they can make accurate predictions on different time windows without requiring retraining on the preceding sequential data. It was found that the model's performance tends to deteriorate when the test data values are outside the range of the training data. However, this limitation does not have any implication for our specific application since an airport boundaries are enclosed within a narrow range of latitude and longitude values. Furthermore, if the LSTM models properly trained, they were observed to be able to forecast bird movement over different time windows with similarly high accuracy. The time windows do not need to sequentially follow the training data. The last observation is, although LSTM models have been successful in predicting bird movement over short periods of time, their performance may suffer when trying to make predictions over longer periods. This is because the models struggle to capture errors that accumulate over multiple time steps, leading to oscillations in performance. 
    
    \item In the real-world scenario, an avian radar can continuously track the bird for a limited period of time of about 3-6 minutes. Therefore, this study has incorporated a limited input horizon of five minutes in the LSTM models. The prediction horizon required for strategic deconfliction has been taken into account, which is the time required by an aircraft to reach its takeoff speed. This is because during this period, the aircraft is still on the ground. Therefore, if a bird strike is detected within this time, it is much cheaper and safer to delay the departure time to prevent the bird strike compared to tactical deconfliction maneuvers. The collision detected after this time will be prevented by tactical deconfliction algorithm. As the prediction horizon for tactical deconfliction does not need to be as long as strategic deconfliction, the chosen prediction horizon is sufficient for both deconfliction algorithms. 
    
    \item The use of the LSTM models in flight planning to ensure deconfliction between aircraft and birds has been demonstrated using a simple flight example. To this end, the bird movement trajectory obtained using the vanilla LSTM model and and the flight trajectory of Boeing 737 aircraft departing from the 06L/24R runway at Cleveland Hopkins International Airport have been considered. The departure of the flight was delayed by a few seconds to ensure that the trajectories of the aircraft and the bird did not intersect. The results show that such small delays in departure time based on the bird movement predictions can prevent potential bird strikes, thereby enhancing aviation safety. 
\end{enumerate}

\subsection{Outline of the Paper}

The rest of the paper is organized as follows. In \ref{sec:LitReview}, we discuss about previous research on bird strikes and their results as well as their limitations. Then, the formal bird strike problem statement considered in this study is described in \ref{sec:Problem Statement}, followed by the solution method used to address the problem in \ref{sec:Solution Method}. After that, the results are presented in \ref{sec:Results}. Finally, the paper concludes with a summary of the research findings and next steps in \ref{sec: Conclusion}.  

\section{Literature Review}
\label{sec:LitReview}

Prior research on bird strikes has focused more on analysis of contributing factors and impact of bird strikes and less on preventing or minimizing probability of bird strikes with aircraft by means of forecasting bird movement to inform flight planning and deconfliction algorithms. They range from studies on identifying high risk birds in different locations in the context of bird strikes and analyzing factors affecting the risk of bird strikes to cost estimation of damages caused by bird strikes and predicting the future number of bird strike incidents in a given region. Limited research has been conducted to develop predictive models to predict flight paths of birds and incorporate them in flight planning for strategic and tactical deconfliction between aircraft and birds. 

To understand and address the issue of bird strikes, it is important to identify high risk birds and study their flight characteristics and determine the underlying factors that contribute to bird strikes. Based on the study of 79 bird species, the highest risk of bird strikes to aircraft in the United States are posed by the following ten species: red-tailed hawks, Canada geese, turkey vultures, pigeons, mourning doves, gulls, American kestrels, barn swallows, European starlings, and killdeers. Many research studies have been conducted to determine the key factors that increase the probability of severe bird strikes \cite{metz2020bird, mathaiyan2021determination, shao2020key}. In \cite{metz2020bird}, various factors such as altitude, time of day, location and environmental conditions, season, and aircraft characteristics were analyzed to determine the probability of bird strikes with aircraft in a given region. Aircraft characteristics such as the size of aircraft and its engine can impact the probability of bird strikes. Large size aircraft with turbofan engines were found to be most likely to ingest birds.

Some studies have focused on analysis and prediction of damages caused by bird strikes using numerical methods. In \cite{smojver2011bird}, a numerical method is provided to predict bird strike induced damage on aeronautical components manufactured from different composite and metallic materials. \cite{riccio2016brief} focused on comparing the numerical approaches commonly adopted to simulate the high velocity bird impact effects on real aeronautical structures. They investigated three different numerical approaches: the first approach adopts rigid body elements to model the bird; the second approach adopts the Lagrangian theory to take into account bird deformation during impact; and the third approach uses the Smooth Particle Hydrodynamic formulation to model the bird body.

While the research discussed so far has provided us with a better understanding of bird strikes, the methods mentioned cannot be used to develop a deconfliction algorithm for preventing bird strikes with aircraft. This is because the main focus was on the analysis of historical data related to bird strikes and did not incorporate any bird strike prevention algorithms. To mitigate the risk of bird strikes using practical measures, three broad strategies are commonly used: bird habitat modification, bird behavior control and aircraft path modification. Habitat modification can discourage birds from roosting and feeding in unwanted areas. This can be achieved by removing food sources, covering ponds, and removing nesting sites through trimming vegetation \cite{bird-habitat}. Using hazing techniques such as pyrotechnics, long range acoustic devices and lasers can modify bird behavior and discourage them from entering a perceived threatening area. Furthermore, changing the appearance of the aircraft, such as using specific colors in its design, is another way to control bird behavior and prevent bird strikes \cite{B-H1,B-H2}. The effectiveness of bird control techniques may vary over time and is dependent on the specific bird species being targeted. Therefore, it is necessary to use a combination of different techniques and to periodically switch them up. Using the same technique repeatedly can lead to decreased effectiveness as the birds become habituated to it. One strategy for bird behavior control involves the use of herding drones, which attempt to drive birds away from a specific location. However, to effectively implement this method, bird movement prediction is necessary, as the herding drone must be in close proximity to the birds' movements to effectively disperse them \cite{paranjape2018robotic}. The last and most effective measure to mitigate bird strikes is modifying drone movement based on bird movement prediction. Accurate prediction of bird movements is crucial, as it enables adjustments to the drone's departure time, speed, and heading. By utilizing this technique, the risk of a bird strike can be minimized to a great extent.

A brief overview of the bird migration monitoring and warning systems used in aviation to reduce the probability of bird strikes during migration is provided in \cite{van2019aeroecology}. Avian radars, weather radars, and Global Positioning System (GPS) are examples of sensors that have been utilized to monitor bird tracks in real time. Based on the real time monitoring of bird movement activity in a given airspace, the warning system can alert the controller whether the airspace is safe for aircraft to enter or not at a given time. 

By supplementing real time bird movement data with bird movement predictions, effective and practical bird strike prevention systems can be developed. One such system involving pilots and controllers was proposed in \cite{metz2021analysis}, where the system takes bird movement predictions as input to assign departure delays to flights. The paper demonstrates via simulation that the number of bird strikes with aircraft can be minimized using this approach. The simple linear regression model was used in \cite{metz2021analysis} to predict bird movement. The results concluded that predicting bird movements based on simple linear regression is not sufficiently accurate to implement a safe flight planning algorithm in the real world. Hence, a strong need for a more accurate and advanced predictive model can be seen in the literature. In order to address this gap, this paper implements four different LSTM models to forecast the bird movement with sufficiently high accuracy for practical implementation. LSTM models are particularly effective in this context due to the various factors that can affect bird movement, such as food availability and weather conditions, finding a stationary pattern in their movement can be challenging. Consequently, statistical models are less effective in accurately predicting bird movement compared to LSTM models. The need for real-time bird movement prediction at an airport requires the airport to be equipped with avian or weather radars. However, avian radars have a limited capability to continuously track bird movements, which causes a limitation in the input horizon for predictive models. This limitation has been considered in this paper. In this paper, we establish a practical prediction horizon for strategic deconfliction using LSTM models and demonstrate their high accuracy in predicting bird movement.

Given an accurate model to predict bird movement, different collision avoidance algorithms that can handle non-cooperative aircraft such as those presented in \cite{zhao2021multiple, 4227576, 1633455, fasano2008multi} can be applied to prevent collision with birds. Note that birds can be treated as non-cooperative aircraft or obstacles in these deconfliction algorithms.

\section{Problem Statement}
\label{sec:Problem Statement}
\begin{figure}[h!]
\centering
\includegraphics[width=16cm, height = 4cm]{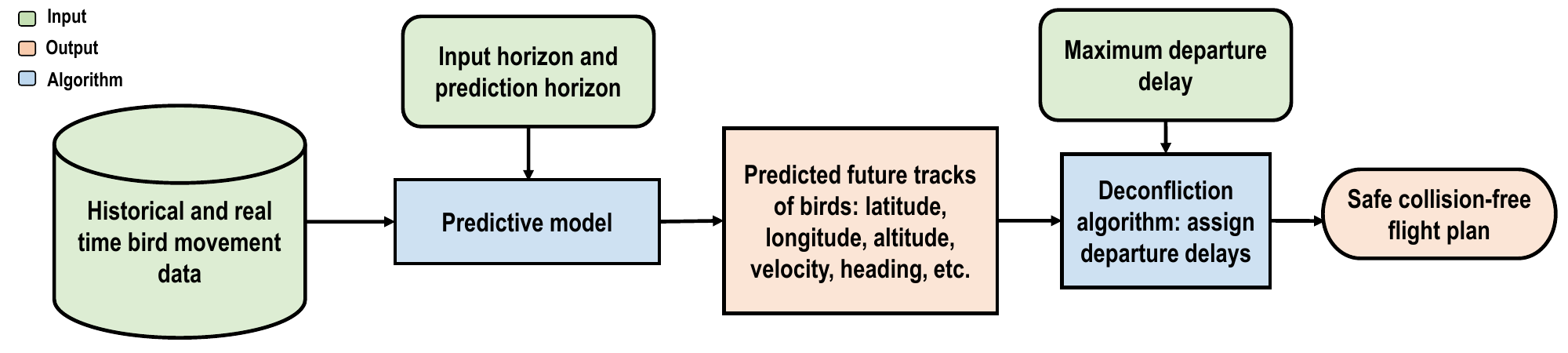}
\caption{Problem flowchart showing the steps involved in bird strike prevention}
\label{Problem flowchart}
\end{figure}

Bird strikes with aircraft poses a safety challenge for both emerging low altitude AAM aircraft as well as conventional aircraft. This problem is especially prevalent in the lower altitudes of the National Airspace System, which would be used by AAM aircraft for their entire flight and which is currently used by conventional aircraft during their takeoff and landing phases. The problem of preventing bird strikes involves two challenges, as shown in Fig. \ref{Problem flowchart}. The first challenge involves using historical and real time bird movement data along with other data such as weather and seasonality to develop predictive models to forecast the bird movement tracks over a given time horizon in terms of latitude, longitude, altitude, velocity and heading with high accuracy. The second challenge concerns using these predictions along with other air traffic and airspace information to make small changes to the flight plan of the aircraft by varying the departure time such that the 4D aircraft flight path does not intersect with the predicted bird path. 

This paper has mainly focused on addressing the first challenge shown in Fig. \ref{Problem flowchart} by collecting data from a high-risk bird in case of bird strikes, selecting appropriate models to predict bird movement, and considering a practical input and prediction horizons for predictive models. Therefore four different LSTM models: vanilla LSTM, stacked LSTM, bidirectional LSTM and encoder-decoder LSTM have been implemented. The models were trained and tested using a publicly available bird movement data set, which featured both the longitude and latitude time series data of birds of a given species. By doing so, the first challenge can be solved in a real-world situation. But the paper also demonstrates how bird movement predictions can be useful to develop a flight planning algorithm which can determine changes to departure time to an existing flight plan to prevent bird strikes with the aircraft. 

\section{Solution Method}
\label{sec:Solution Method}
\begin{figure}[t!]
\centering
\includegraphics[width=16cm, height = 6.5 cm]{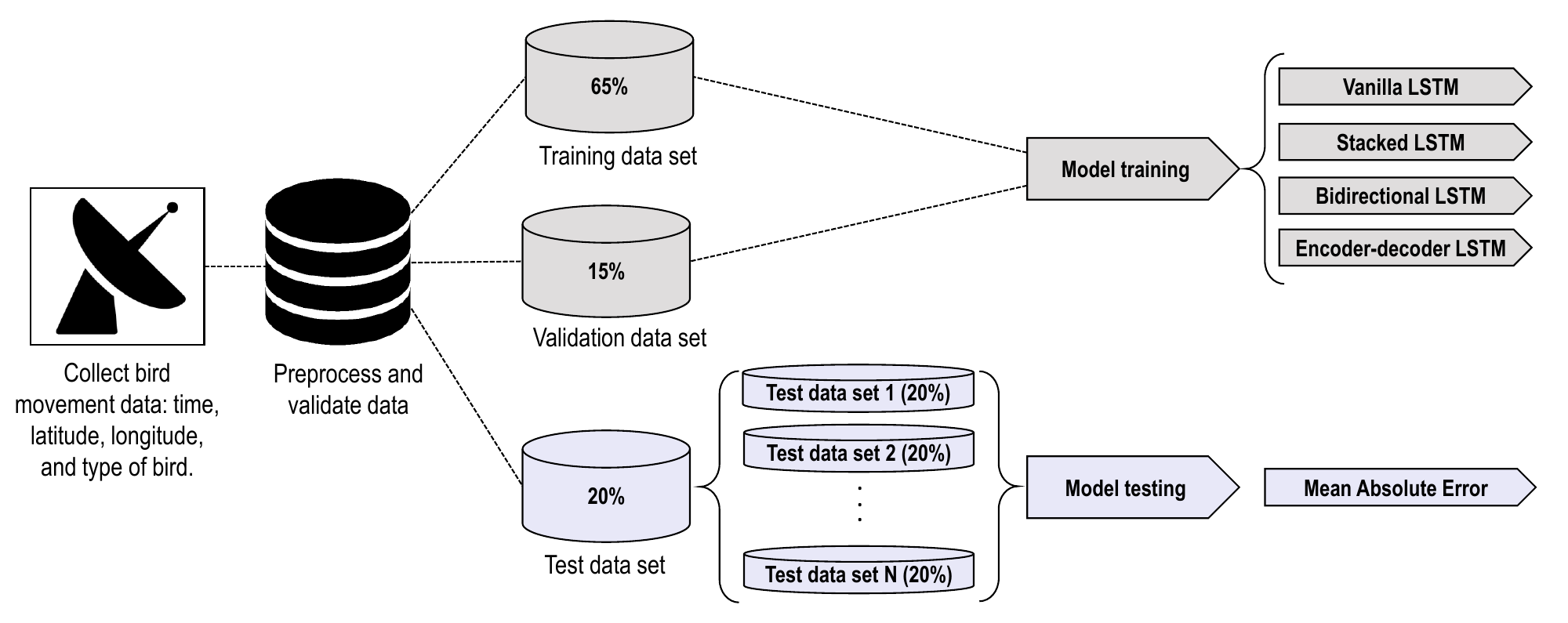}
\caption{Methodology for training and testing predictive models to forecast bird movement}
\label{flowchart}
\end{figure}

As discussed in the preceding section, the primary focus of this paper is to address the first challenge which involves collecting historical and real time bird movement data, selecting an appropriate predictive model and a practical prediction horizon accordingly. As shown in Fig. \ref{flowchart} the first step towards this goal involves collecting data of high-risk birds involved in bird strikes, which can be obtained using various types of sensors such as avian radar, weather radar, and GPS. The second step involves preprocessing the collected data, which includes addressing any missing data points through linear interpolation. The third step includes dividing the preprocessed data into three separate data sets: training, validation, and test data sets. This division involves allocating 65\% to the training data set, 15\% to the validation data set, and the remaining 20\% to the test data set. The training and validation data sets have been used to train four different LSTM models namely vanilla, stacked, bidirectional, and encoder-decoder LSTM. The validation data set has been utilized to prevent overfitting of the models. To evaluate the performance of the models, the MAE was calculated for each model using different test data sets. To assess the robustness of the models, various test data sets were used to evaluate their performance on data beyond the training and validation sets.

\subsection{Bird Movement Data}

To develop the models to predict future flight tracks of a given bird species, first of all, a data set consisting of latitudes and longitudes of birds of that species at different points in time is needed for training and testing the models. In this paper, a data set on the movement of pigeons, which is one of the top ten high-risk birds \cite{BS1}, is utilized for this purpose. The data set which is publicly available on MoveBank \cite{BS6} includes 14400 data points, each corresponding to one second over a four-hour period. Table \ref{new table} shows a snapshot of the data. Linear interpolation is used on the data to ensure that all latitudes and longitudes are evenly spaced apart in time.  
\floatplacement{figure}{!h}
\begin{table}[h]
\caption{A snapshot of the data}
\label{new table}
\centering
\begin{tabular}{@{}cccc@{}}
\toprule
\centering
\textbf{Time}                                 & \textbf{Latitude ($^{\circ}$)}                                      & \textbf{Longitude ($^{\circ}$)}                                     & \textbf{Bird type}                                    \\ \midrule
8:00:00 AM                                           & 43.215031                                        & 10.5719799                                      & Pigeon                                  \\
8:00:01 AM                                          & 43.2148682                                       & 10.5719911                                       & Pigeon                                  \\
8:00:02 AM                                           & 43.2147144                                        & 10.5720345                                       & Pigeon                                  \\
8:00:03 AM                                           & 43.2145836                                      & 10.5721471                                       & Pigeon                                  \\
.                                          & .                                      & .                                     & .                                 \\
.                                          & .                                      & .                                     & .                                 \\
.                                          & .                                      & .                                     & .                                 \\ \midrule
\end{tabular}
\end{table}
\subsection{Predictive models: Long Short-Term Memory Models}
Bird movement forecasting can be potentially carried out using different predictive models such as linear and nonlinear regression models, Autoregressive Integrated Moving Average (ARIMA) time series model, and LSTM models. To solve bird strike problem, LSTM models are a suitable choice since they can take in a sequence of input data and predict future data based on that sequence. Unlike linear and nonlinear regression models, which are limited by their fixed trend based on the training data, LSTM models are capable of adapting to varying trends in the output based on different input sequences. This makes them particularly useful for predicting trends in situations where there is a lot of variation in the data over time. In addition, due to the non-stationary nature of bird movement data, it can be hard to be analyzed by statistical models like ARIMA. Therefore, LSTM models can be appropriate choice in the case of bird movement prediction. In this paper, we considered the use of four different types of LSTM models to predict bird movement:  vanilla LSTM, stacked LSTM, bidirectional LSTM and encoder-decoder LSTM. LSTM models are an extension of Recurrent Neural Networks (RNNs). RNN is a network that focuses on the present input and the previous outputs. It stores them in its memory for a short period of time. Consequently it cannot manage long dependencies. LSTM was introduced to handle the situations where RNN cannot manages \cite{BS7}.

\begin{figure}
     \centering
     \begin{subfigure}[h!]{0.497\textwidth}
         \centering
         \includegraphics[width=\textwidth]{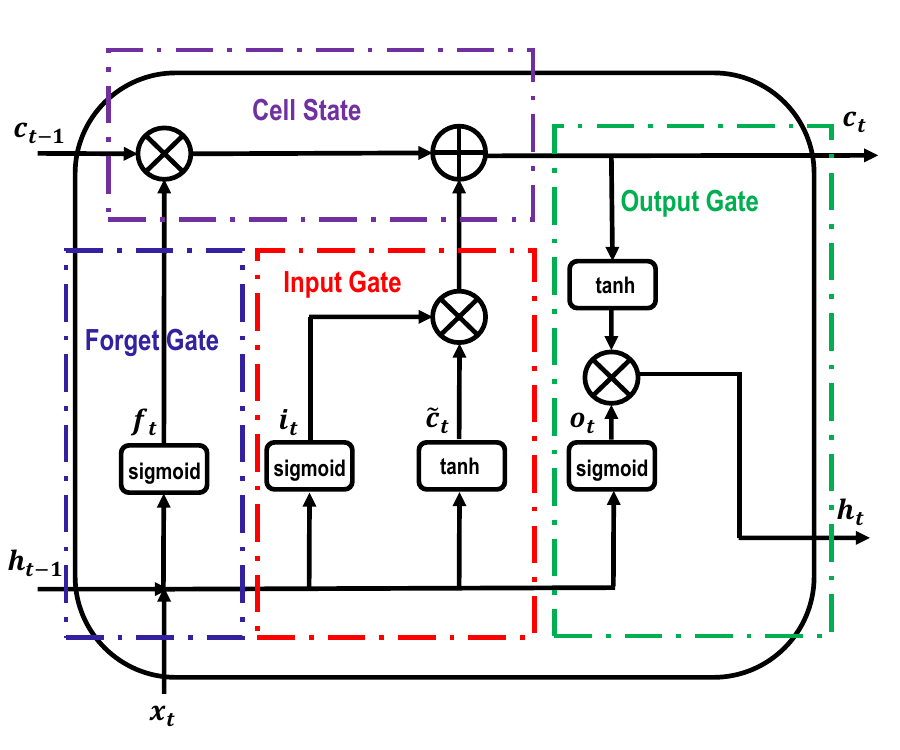}
         \caption{LSTM cell (adapted from  \cite{BS9})}
         \label{LSTM cell}
     \end{subfigure}
     \hfill
     \begin{subfigure}[h!]{0.497\textwidth}
         \centering
         \includegraphics[width=\textwidth]{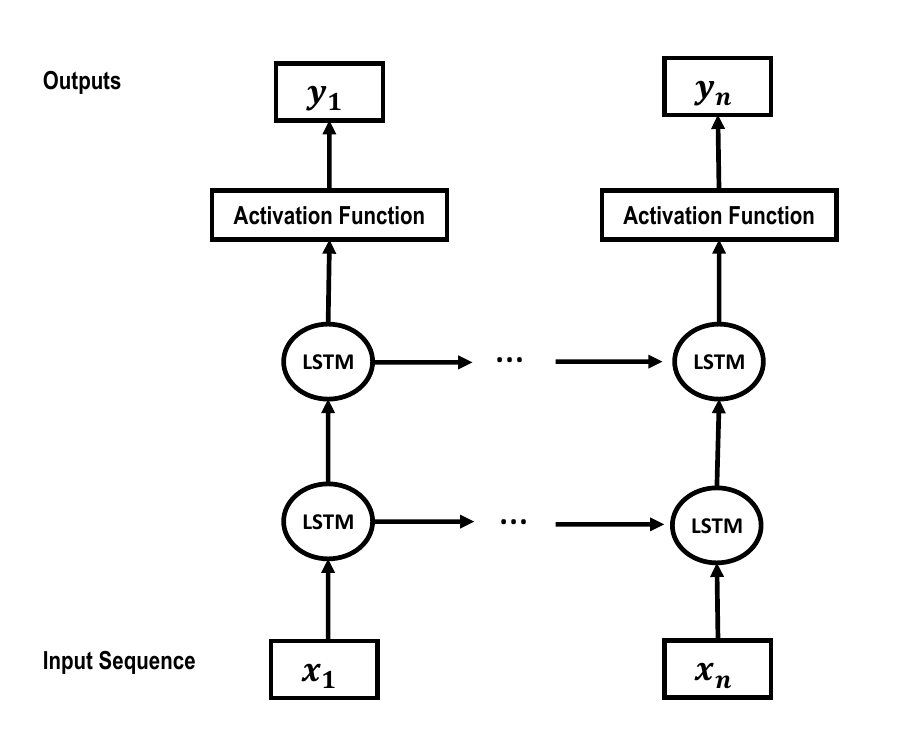}
         \caption{Stacked LSTM (adapted from  \cite{sahar2018lstm})}
         \label{Stacked LSTM}
     \end{subfigure}
%###########################
     \centering
     \begin{subfigure}[h!]{0.497\textwidth}
         \centering
         \includegraphics[width=\textwidth]{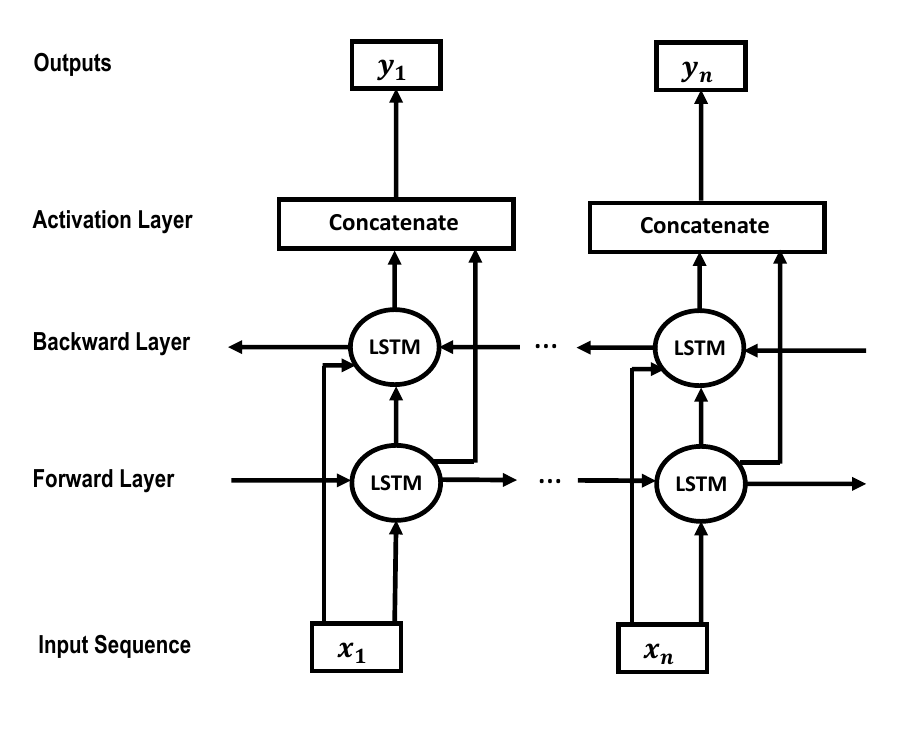}
         \caption{Bidirectional LSTM (adapted from  \cite{tavakoli2019modeling})}
         \label{Bidirectional LSTM}
     \end{subfigure}
     \hfill
     \begin{subfigure}[h!]{0.497\textwidth}
         \centering
         \includegraphics[width=\textwidth]{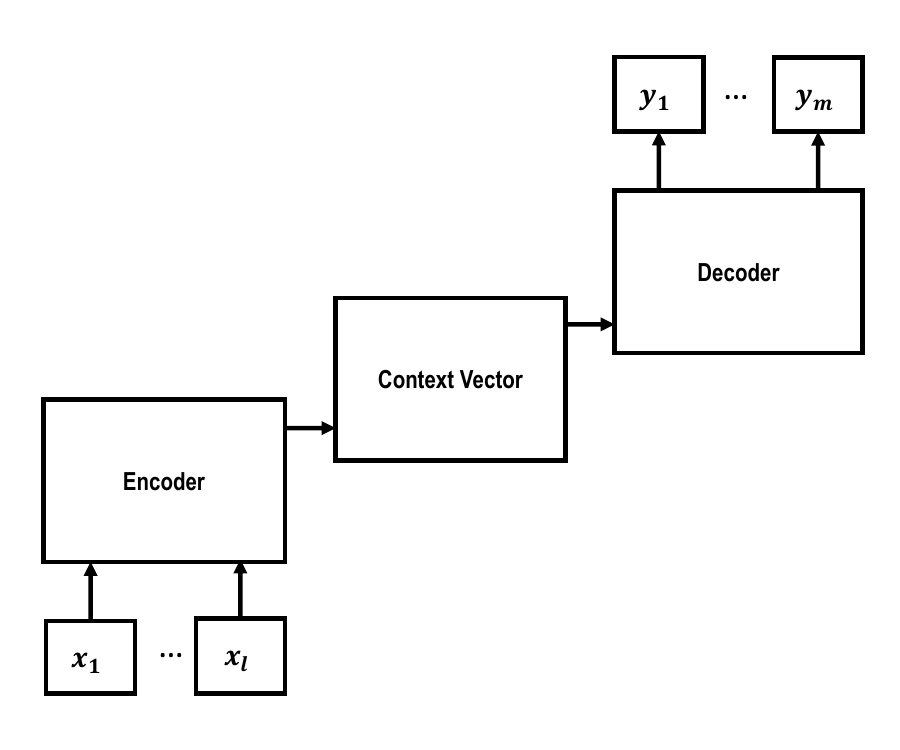}
         \caption{Encoder-Decoder LSTM (adapted from  \cite{jacoby2021short})}
         \label{Encoder Decoder LSTM}
     \end{subfigure}
     \caption{The architectures of four different LSTM models}
     \label{LSTM Models}
\end{figure}

 LSTMs are designed to handle both Long Term Memory (LTM) and Short-Term Memory (STM). They use gates to control the process of memorization of past data for simpler and more effective calculations. An LSTM cell, as shown in Fig. \ref{LSTM cell}, consists of three different gates: a forget gate, an input gate, and an output gate \cite{BS5}. The forget gate determines which specific set of data points needs attention and which can be ignored.
 
 The current input \(x_t\) (e.g., current bird position in the context of bird movement prediction problem) and the previous hidden state \(h_{t-1}\) are passed through an sigmoid function. Sigmoid generates values between 0 and 1 which determines whether the part of the old output is necessary (1) or not (0).  The sigmoid output \(f_t\) will later be used by the cell state for point by point multiplication. 

%###################################################################################
%################################################################################################
The next step involves determining what new information should be stored in the cell state through the input gate. It has two different parts. First, a sigmoid function decides which data points will be updated. Next, a tanh function creates a vector of new candidate data points, \(\Tilde{c}_{t}\), that could be added to the state \cite{BS8}. First, the current state \(x_t\) and previous hidden state \(h_{t-1}\) are passed into the sigmoid function to generate the input vector \(i_t\). The values are again transformed between 0 and 1 which determines which data points should be kept and which should be eliminated. Next, the same input for sigmoid function in input gate, the hidden state and current state is passed through the tanh function to regulate the network since the tanh eliminates the bias of the network. The output of the tanh function is \(\Tilde{c}_{t}\) which can be between -1 and 1. The output values generated from the activation functions are then multiplied point-by-point. 

Now, to update the old cell state \(c_{t-1}\), into the new cell state \(c_t\), the old state is multiplied by \(f_t\) forgetting the things decided to forget earlier. Then the value of \(i_t * \)\(\Tilde{c}_{t}\) will be added to the output of multiplication and new candidate values for cell state will be created. 

The final output of the LSTM cell, \(h_{t}\), is based on a filtered version of the cell state. First, the values of the current state \(x_t\) and previous hidden state \(h_{t-1}\)  are passed into the third sigmoid function. Then the new cell state  \(c_t\) generated from the cell state is passed through the tanh function. Both these outputs are multiplied point-by-point and create the output of the LSTM cell \(h_t\)\cite{BS8}.

As mentioned previously, four different types of LSTM models have been implemented and tested in this study. The architecture of these different types of LSTM models are depicted in Fig. \ref{LSTM Models}. 

\textit{Vanilla LSTM}: A vanilla LSTM has only a single hidden layer of LSTM units, making it a type of feed-forward neural network. In a vanilla LSTM, there can be many inputs, but the output is always a single value. Therefore, it is referred to as a "many-to-one" algorithm. 

\textit{Stacked LSTM}: Multiple hidden LSTM layers can be stacked on top of one another to improve prediction accuracy. The stacked LSTM is an extension of the LSTM that has multiple hidden LSTM layers, as shown in Fig. \ref{Stacked LSTM}. Each layer contains multiple LSTM cells. The stacked LSTM hidden layers make the model deeper. Like vanilla LSTM, in stacked LSTM, the input length is not limited, and it only outputs a single variable. Note that even though the input length is not limited in theory, it is limited by the memory capacity in practice. 

\textit{Bidirectional LSTM}: The process of bidirectional learning is implemented in some sequence prediction problems, which allows the LSTM model to learn the input sequence both forwards and backwards and concatenate both interpretations as shown in Fig. \ref{Bidirectional LSTM}. Like vanilla and stacked LSTM, bidirectional LSTM is also a "many-to-one" algorithm. 
 
\textit{Encoder-decoder LSTM}: Vanilla, stacked, and bidirectional LSTM models are designed for one-step prediction (i.e., prediction of value at the next time step). In contrast, the er LSTM is a type of recurrent neural network which is designed for multi-step prediction. Such multi-step prediction problems are also referred to as sequence-to-sequence problems. Sequence-to-sequence prediction problems are challenging since the length of the output sequences is more than one time step. In other words, there are multiple input time steps and multiple output time steps, which is also called a "many-to-many" type sequence prediction problem. An encoder-decoder LSTM is composed of two recurrent neural networks that function as an encoder and a decoder pair. The encoder maps a variable-length source sequence to a fixed-length vector, and the decoder maps the vector representation back to a variable-length target sequence. As shown in Fig. \ref{Encoder Decoder LSTM}, \(l\) denotes the length of input sequences and \(m\) represents the length of output sequences.

To increase the prediction horizon of vanilla, stacked, and bidirectional models, a recursive method can be implemented in which the output of each prediction is fed back into the model as input for the subsequent prediction. This recursive approach enables the generation of a sequence of predictions for each time step in the interval, which can then be merged to yield a prediction horizon greater than one time step.
\newpage
\subsection{Model Training and Testing}
As mentioned earlier, the processed data set of the pigeon movement has been partitioned to three different parts namely training, validation, and test data sets allocating 65\%, 15\%, and 20\% of the data set respectively. The model learns the relationship between inputs and outputs using a training data set. The number of inputs at each time step is called the input horizon, while the number of outputs at each time step is known as the prediction horizon. Vanilla, stacked, and bidirectional LSTM models are capable of predicting the next output. However, an encoder-decoder LSTM can predict multiple outputs, so the model must understand the relationship between multiple inputs and outputs during training. In each epoch, the model faces with all input and output examples in training data and tries to update its weights by making a comparison between predicted output and real output and applying backpropagation. To avoid overfitting and bias, the model also evaluates its performance on a separate validation data set during each epoch. The weights of the neural network are determined in the epoch with the best performance of the model on the validation data set. Five different data sets were used to evaluate the performance of the models, with MAE as the performance metric. Accurate prediction can be achieved when the range of values in the training data set is equal to or greater than the range of values in the test data set. In a real-world airport scenario, the training data should cover the bounds of the airport to ensure accurate prediction of bird movements.

\section{Results}
\label{sec:Results}
In this section, we present the results of our analysis, which are organized into three subsections. In subsection \ref{R-S-1}, we discuss the experimental configuration used in our study. Subsection \ref{R-S-2} presents the evaluation of the predictive models' performance according to MAE.  Finally, in subsection \ref{R-S-5}, we present a collision avoidance demonstration to prevent bird strikes during takeoff.

\subsection{Experimental Configuration}
\label{R-S-1}
 As discussed previously, we implemented four different LSTM models to predict bird movement as well as linear and nonlinear regression models to be as benchmark models. To consider a realistic prediction horizon for the LSTM models in the real world, we took into account the takeoff operation of a Boeing 737 aircraft in one of the runways at Cleveland Hopkins International Airport (6L/24R). In this paper, the prediction horizon is considered as the time it takes for the aircraft to take off and reach its takeoff speed in a standard day. The aircraft requires approximately 30 seconds to reach its optimal takeoff speed on this specific runway. The reason for using this prediction horizon is that within this time the aircraft is on the ground. Therefore, if a potential collision is detected within this time, it can be prevented by taking into account a delay in the departure time. Any collision occurring after this particular time can be prevented using a tactical deconfliction algorithm, and the chosen prediction horizon is sufficient to achieve this goal. 
 
 To forecast the trajectory of the pigeon movement over a 30-second period, different types of LSTM models can be utilized, including many-to-one models such as vanilla, bidirectional, and stacked models, as well as many-to-many models like the encoder-decoder model. Many-to-one models are specifically designed to predict one time step at a time, where each time step is equivalent to a fixed interval. In this study, we consider the time step length to be one second. In order to predict the entire 30-second interval with many-to-one models, we employed a recursive method, in which we fed the output of each prediction back into the model as input for the subsequent prediction. This recursive approach enables the generation of a sequence of predictions for each time step in the interval, which can then be merged to yield a complete trajectory of the bird's movement over time. The paper considered an input horizon of 300 steps, which is equivalent to five minutes, to track and collect data on birds. This time frame is applicable when using avian radar systems, which can continuously track a bird for a limited period of time, usually around 3-6 minutes. 

The experimental configuration for our LSTM models consisted of several key parameters that were carefully set to ensure accurate performance. As shown in Table \ref{tab:configuration}, the training data set comprised a total of 9300 samples that were used to train the LSTM model. We also utilized five different test data sets to assess the performance of our models, with the MAE serving as the primary performance metric for each experiment. The test data sets, which were separate from the training data set, contained 2880 samples. Additionally, a validation data set was used during training to monitor the model's performance on a data set separate from the training data set and prevent overfitting. The validation data set contained 2220 samples. Finally, the LSTM model was designed to predict 30 steps into the future based on 300 previous steps of the input sequence. 
\floatplacement{figure}{!h}
\begin{table}[h]
\caption{Experimental configuration}
\label{tab:configuration}
\centering
\begin{tabular}{@{}ll@{}}
\toprule
\textbf{Name}  & \textbf{Value} \\ \midrule
Training data set & \(9300\) \\
Validation data set & \(2220\) \\
Test data set & \(2880\) \\
Input horizon & 300 s \\
Prediction horizon & 30 s \\
Time resolution & 1 s \\\midrule
\end{tabular}
\end{table}
\newpage
To ensure accurate predictions in our LSTM models, we set various hyperparameters as shown in Table \ref{tab:hyperparameters}. We tested different settings for each of these hyperparameters and found that using Rectified Linear Unit (ReLU) as the activation function produced the best performance, compared to sigmoid, linear, and tanh. We did not observe a significant difference in the performance of models when we changed the batch size from 32 to 64, so we decided to keep it at 32. We also varied the loss function from Mean Squared Error (MSE) to MAE, but found that it did not significantly affect the model's error. The only hyperparameter that had a significant impact on the models' error was the number of neurons, which we varied from 8 to 64 in our experiments. The number of neurons that produced the least error varied from model to model. After determining the hyperparameter settings that produced the least error, we discovered that all models utilized ReLU as their activation function, MSE as their loss function, and a batch size of 32. We did not change the learning rate it was set as the default value of 0.001. Adam was used as the optimizer for all LSTM models. In addition, the number of epochs for all models was chosen to be 100, ensuring sufficient time for the models to converge and learn the patterns in the data.

\floatplacement{figure}{!h}
\begin{table}[h]
\caption{LSTM hyperparameters}
\label{tab:hyperparameters}
\centering
\begin{tabular}{@{}ll@{}}
\toprule
\textbf{Name}  & \textbf{Size} \\\midrule
Activation function & ReLU \\
Loss function & MSE\\
Batch size & 32 \\
Learning rate & 0.001\\ 
Optimizer & Adam\\ 
Number of epochs & 100\\ \midrule
\end{tabular}
\end{table}

\subsection{Performance of the Predictive Models}
\label{R-S-2}
In this subsection, we evaluate the performance of all predictive models implemented in this paper based on the MAE. We present the performance of predictive models in two subsequent subsections. \ref{R-S-3} shows the performance of LSTM models while \ref{R-S-4} presents the performance of linear and nonlinear regression models as benchmark models.

To compare the different predictive models effectively, we tested them on five distinct test data sets and calculated their MAE values. The results are summarized in Tables \ref{tab:Latitude} and \ref{tab:Longitude}.
% \floatplacement{table}{!h}
\begin{table}[!h]
\caption{Performance metric of the models (MAE in meters): Latitude prediction}
\label{tab:Latitude}
\centering
\begin{tabular}{@{}lllllll@{}}
\toprule
\textbf{Models} & \textbf{Test data set 1} & \textbf{Test data set 2} & \textbf{Test data set 3} & \textbf{Test data set 4} & \textbf{Test data set 5} \\\midrule
Vanilla LSTM  & \(73.72\)  & \(43.61\) & \(21.66\)  & \(50.18\) & \(57.66\) \\
Stacked LSTM & \(73.26\)  & \(51.66\) & \(54.30\)  & \(36.51\) & \(65.63\) \\
Bidirectional LSTM  & \(66.75\)  & \(42.72\) & \(12.81\)  & \(47.43\) & \(40.61\) \\
Encoder-decoder LSTM  & \(98.58\)  & \(84.53\) & \(37.39\)  & \(76.08\) & \(90.58\)  \\
Linear Regression  &  \(5603.86\)  & \(10560.13\) & \(12516.25\)  & \(10931.26\) & \(14875.31\) \\
Nonlinear Regression   & \(26281.11\)  & \(79797.16\) & \(137960.76\)  & \(194179.73\) & \(239258.84\) \\ \bottomrule
\end{tabular}
\end{table}

\begin{table}[!h]
\caption{Performance metric of the models (MAE in meters): Longitude prediction}
\label{tab:Longitude}
\centering
\begin{tabular}{@{}lllllll@{}}
\toprule
\textbf{Models} & \textbf{Test data set 1} & \textbf{Test data set 2} & \textbf{Test data set 3} & \textbf{Test data set 4} & \textbf{Test data set 5} \\\midrule
Vanilla LSTM  & \(74.49\)  & \(34.65\) & \(31.92\)  & \(61.35\) & \(44.91\) \\
Stacked LSTM & \(37.23\)  & \(74.75\) & \(36.21\)  & \(49.70\) & \(69.47\) \\
Bidirectional LSTM  & \(72.27\)  & \(19.68\) & \(33.48\)  & \(60.13\) & \(50.48\) \\
Encoder-decoder LSTM  & \(43.58\)  & \(28.22\) & \(24.96\)  & \(40.77\) & \(46.20\)  \\
Linear Regression  &  \(796.27\)  & \(3302.92\) & \(2206.29\)  & \(2964.36\) & \(4687.98\) \\
Nonlinear Regression   & \(1877.71\)  & \(23341.77\) & \(71230.81\)  & \(175091.65\) & \(362396.07\) \\ \bottomrule
\end{tabular}
\end{table}
\newpage
\subsubsection{LSTM Models}
\label{R-S-3}
\begin{figure}
     \centering
     \begin{subfigure}[p]{0.495\textwidth}
         \centering
         \includegraphics[width=\textwidth]{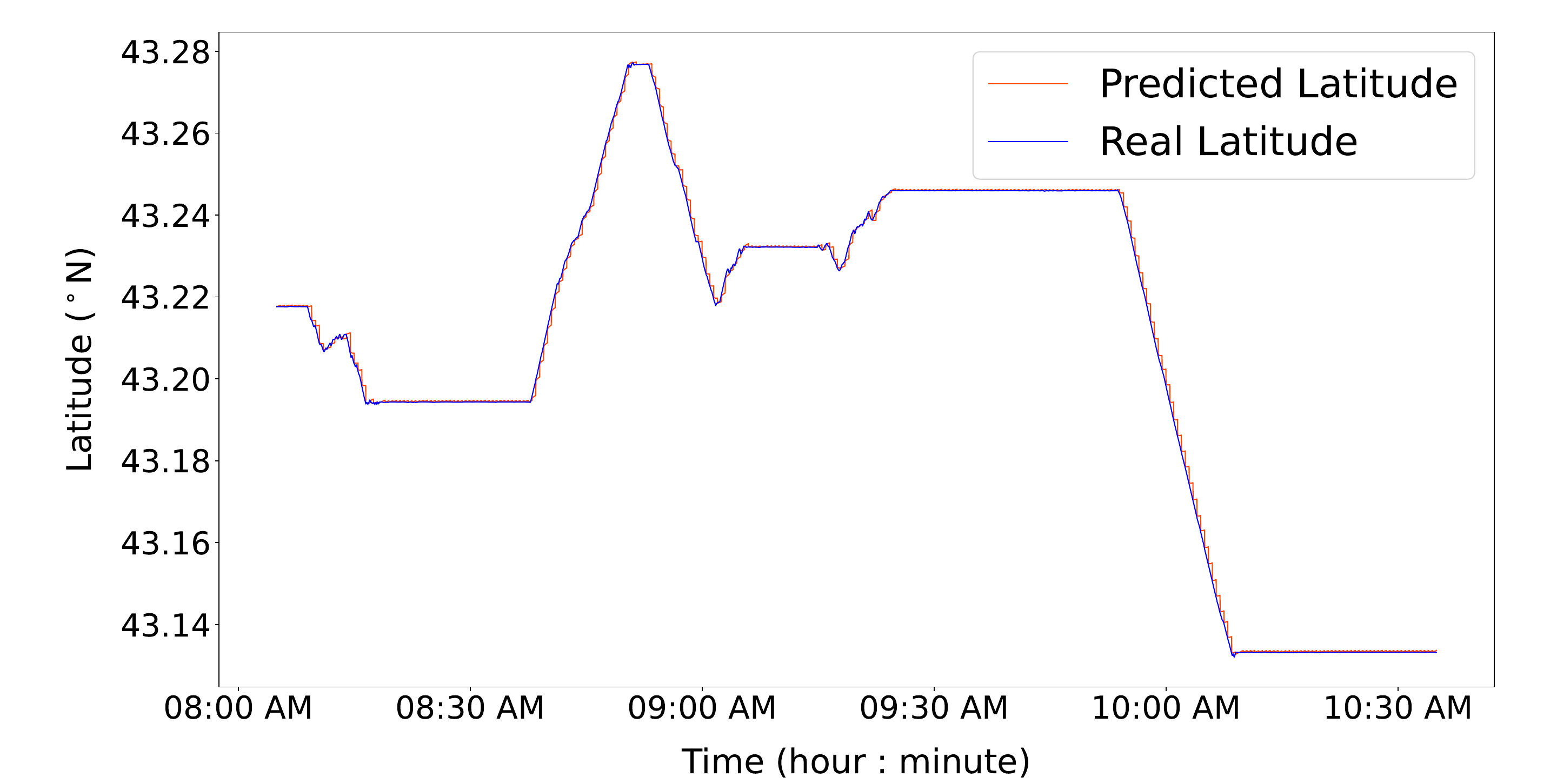}
         \caption{Latitude prediction during training}
         \label{V-Lat-T}
     \end{subfigure}
     \hfill
     \begin{subfigure}[p]{0.495\textwidth}
         \centering
         \includegraphics[width=\textwidth]{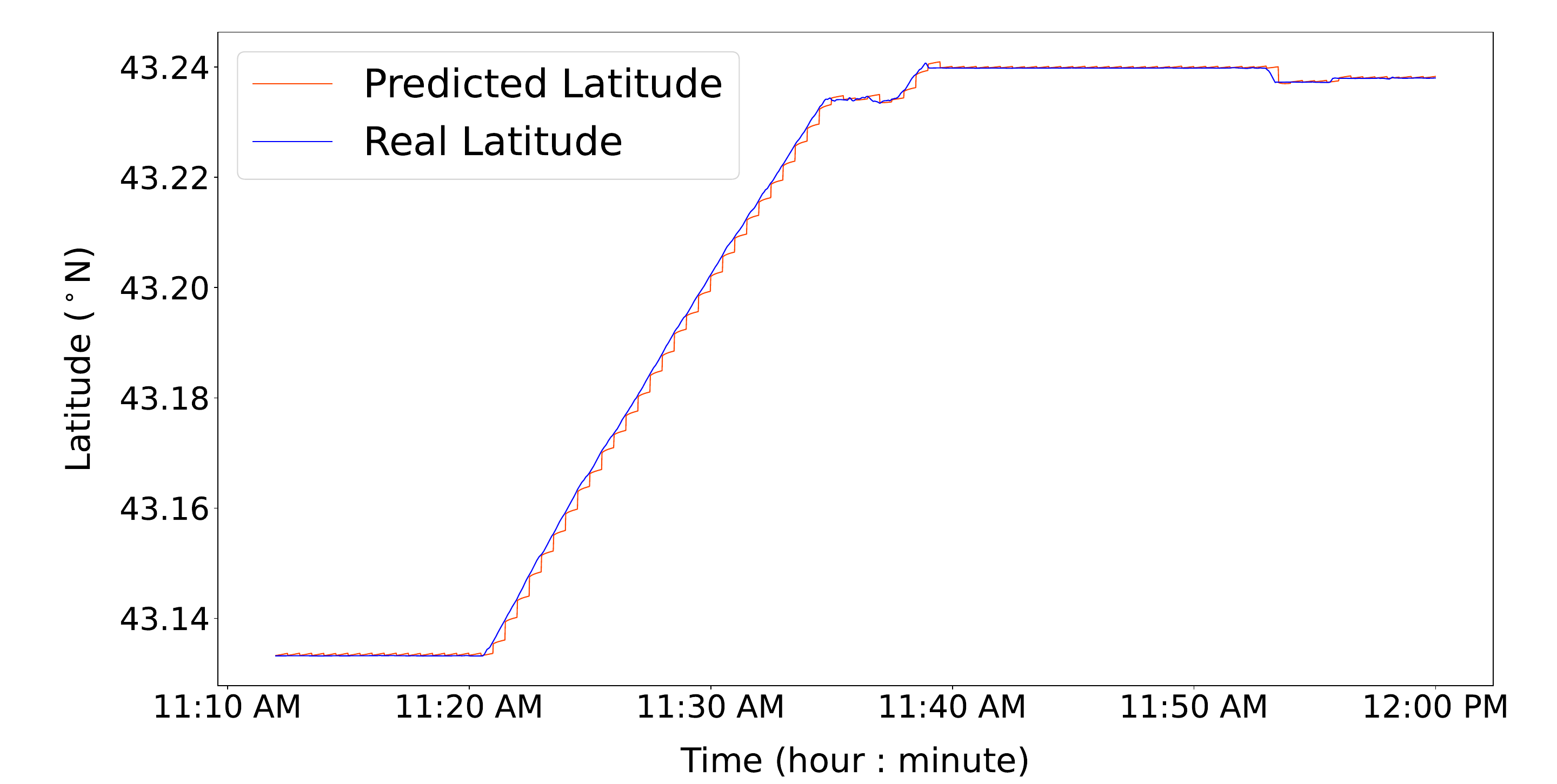}
         \caption{Latitude prediction during the first test data set}
         \label{V-Lat1}
     \end{subfigure}
%###########################
     \centering
     \begin{subfigure}[p]{0.495\textwidth}
         \centering
         \includegraphics[width=\textwidth]{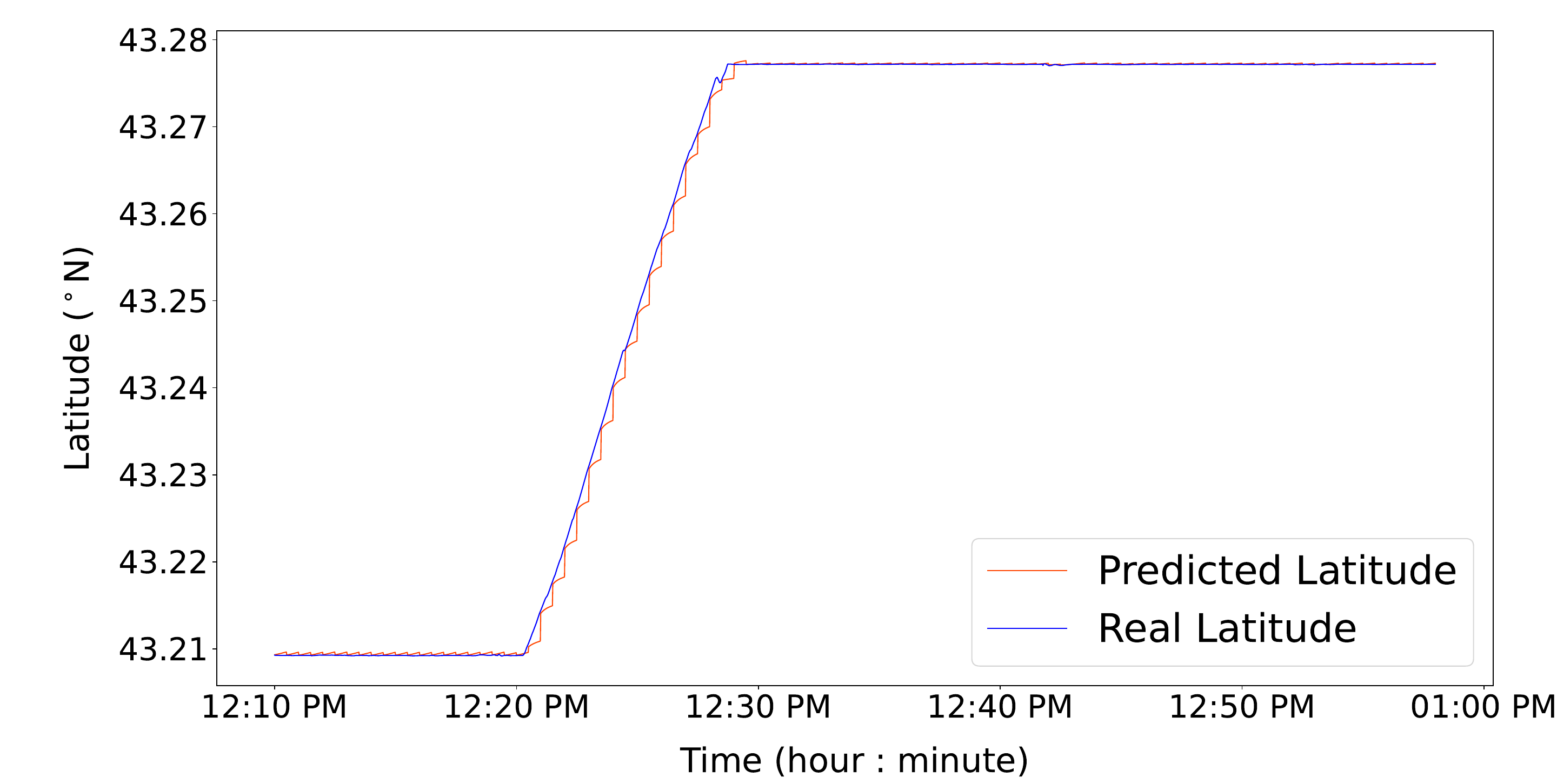}
         \caption{Latitude prediction during the second test data set}
         \label{V-Lat2}
     \end{subfigure}
     \hfill
     \begin{subfigure}[p]{0.495\textwidth}
         \centering
         \includegraphics[width=\textwidth]{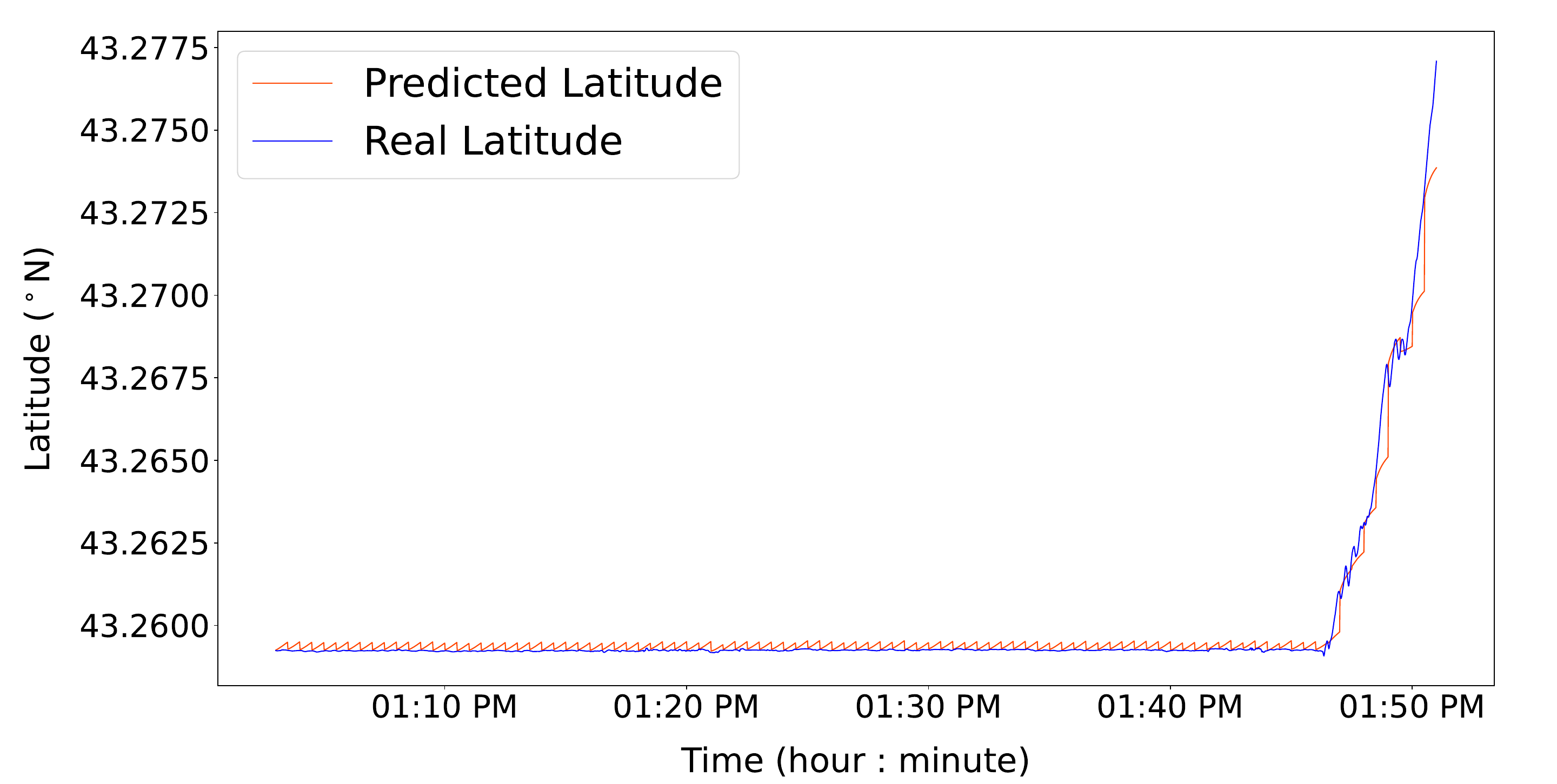}
         \caption{Latitude prediction during the third test data set}
         \label{V-Lat3}
     \end{subfigure}
%###########################
     \centering
     \begin{subfigure}[p]{0.495\textwidth}
         \centering
         \includegraphics[width=\textwidth]{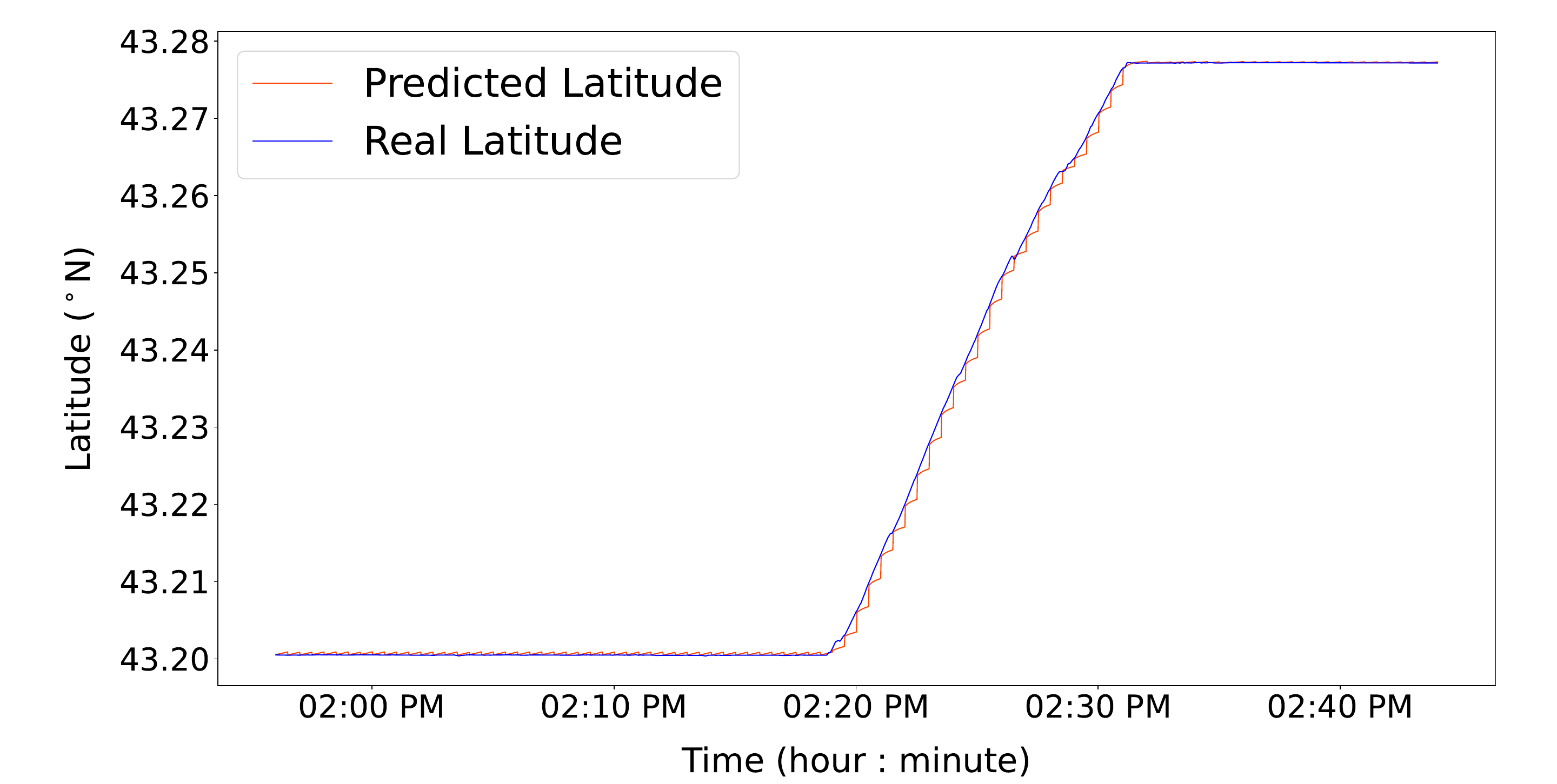}
         \caption{Latitude prediction during the forth test data set}
         \label{V-Lat4}
     \end{subfigure}
     \hfill
     \begin{subfigure}[p]{0.495\textwidth}
         \centering
         \includegraphics[width=\textwidth]{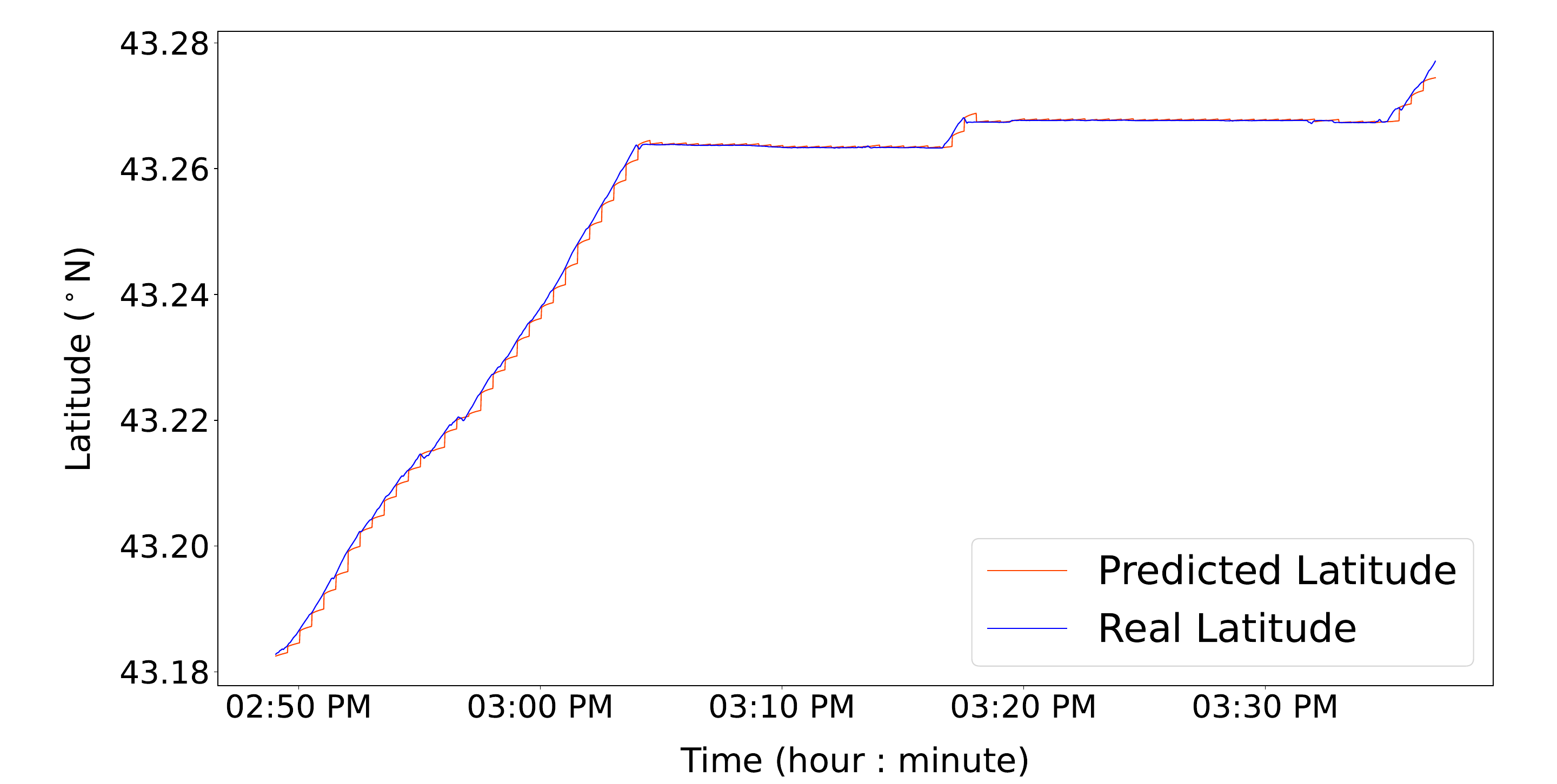}
         \caption{Latitude prediction during the fifth test data set}
         \label{V-Lat5}
     \end{subfigure}
     \caption{Latitude predicted by vanilla LSTM}
     \label{Fig: Latitude-Vanilla}
\end{figure}

\begin{figure} 
     \centering
     \begin{subfigure}[p]{0.495\textwidth}
         \centering
         \includegraphics[width=\textwidth]{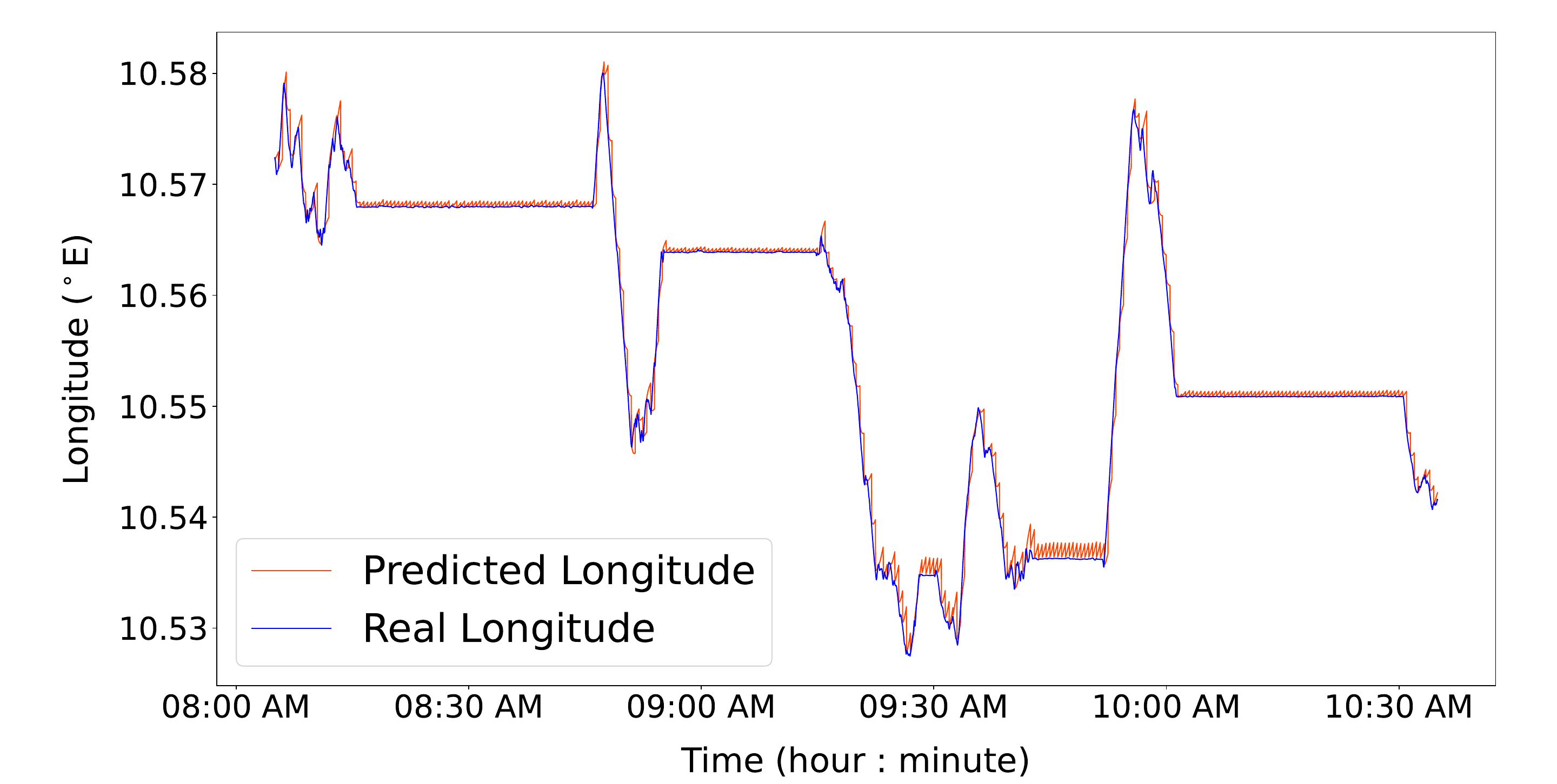}
         \caption{Longitude prediction during training}
         \label{V-Long-T}
     \end{subfigure}
     \hfill
     \begin{subfigure}[p]{0.495\textwidth}
         \centering
         \includegraphics[width=\textwidth]{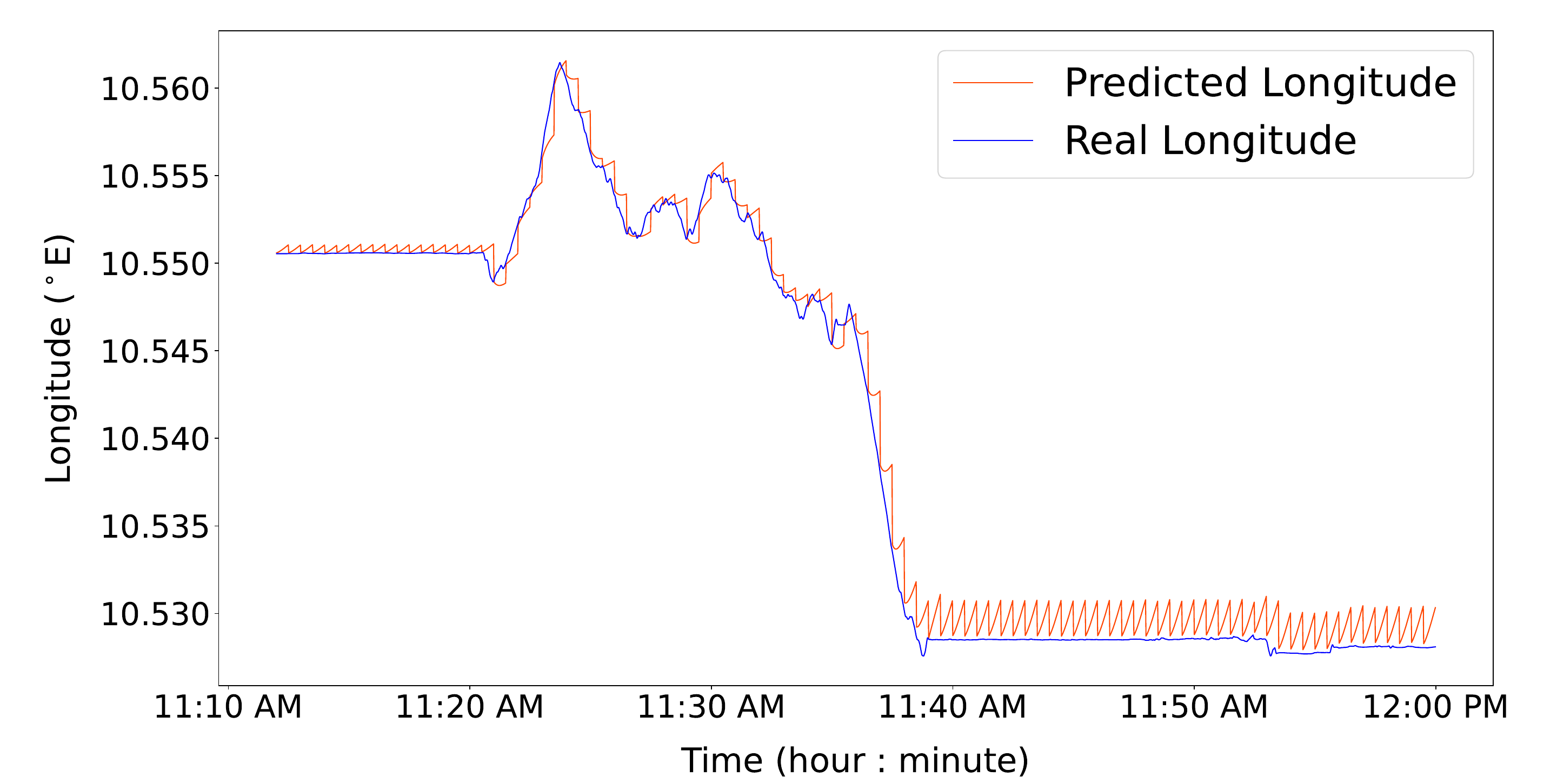}
          \caption{Longitude prediction during the first test data set}
         \label{V-Long1}
     \end{subfigure}
%###########################
     \centering
     \begin{subfigure}[p]{0.495\textwidth}
         \centering
         \includegraphics[width=\textwidth]{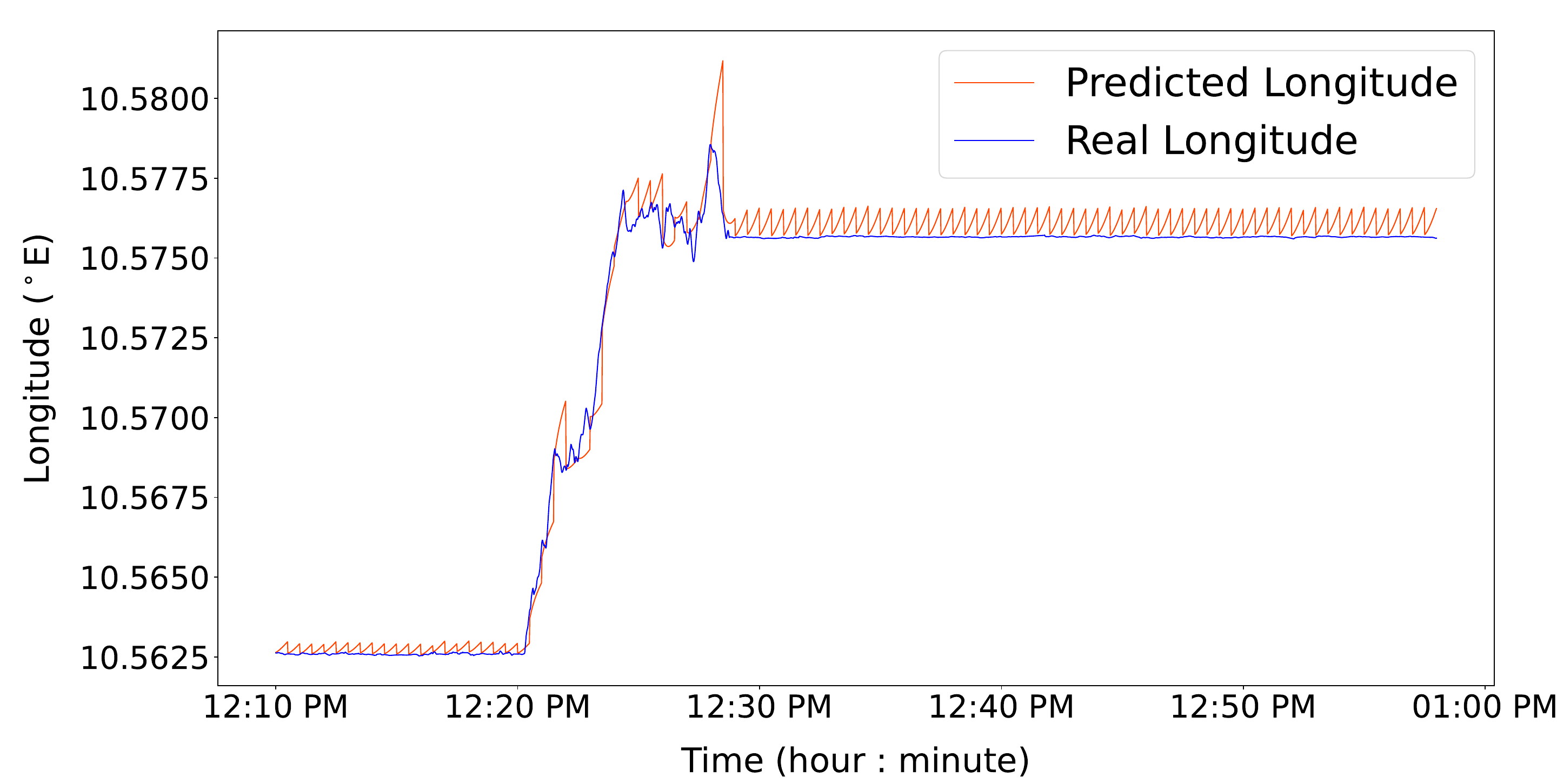}
         \caption{Longitude prediction during the second test data set}
         \label{V-Long2}
     \end{subfigure}
     \hfill
     \begin{subfigure}[p]{0.495\textwidth}
         \centering
         \includegraphics[width=\textwidth]{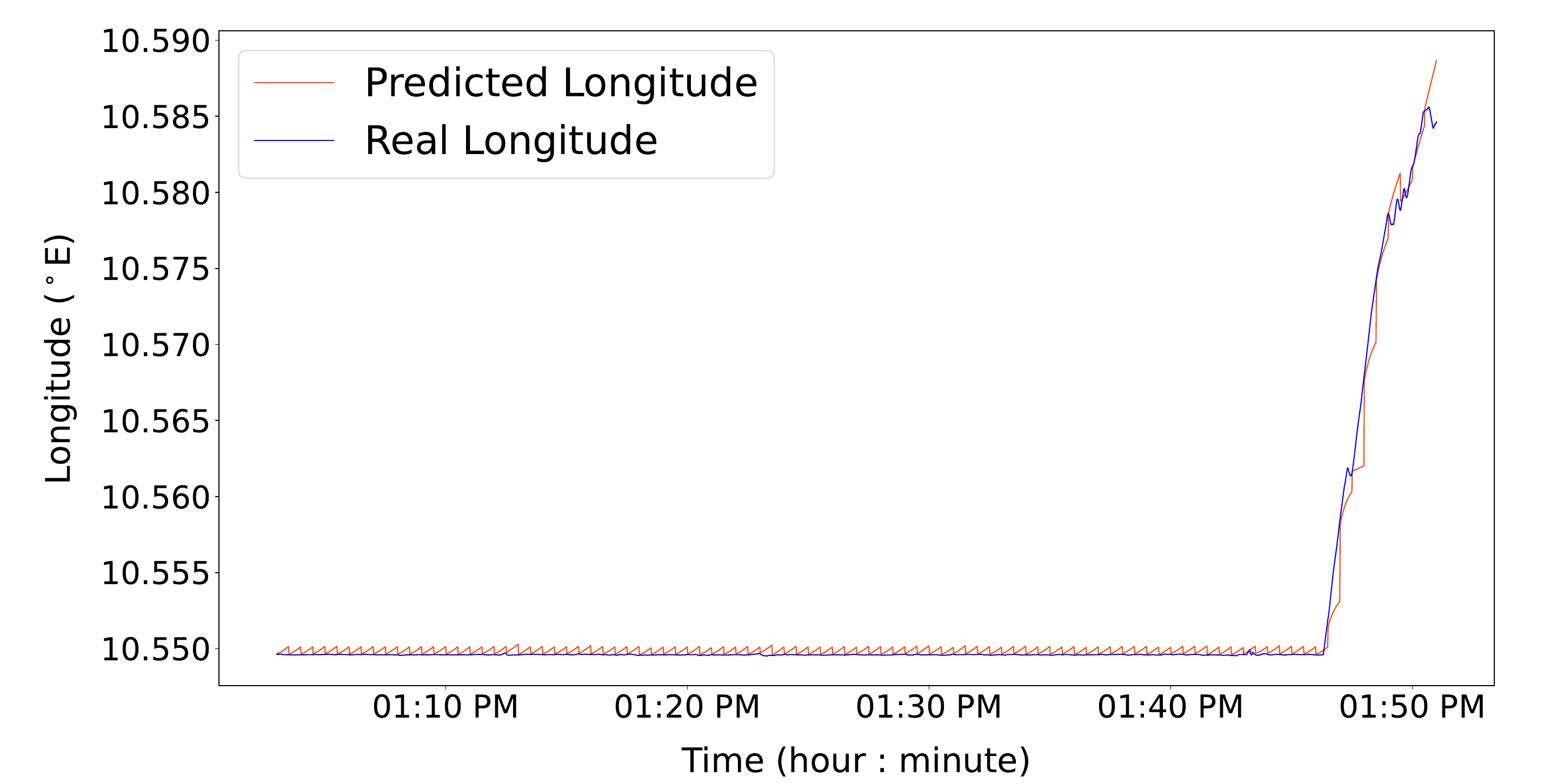}
         \caption{Longitude prediction during the third test data set}
         \label{V-Long3}
     \end{subfigure}
%###########################
     \centering
     \begin{subfigure}[p]{0.495\textwidth}
         \centering
         \includegraphics[width=\textwidth]{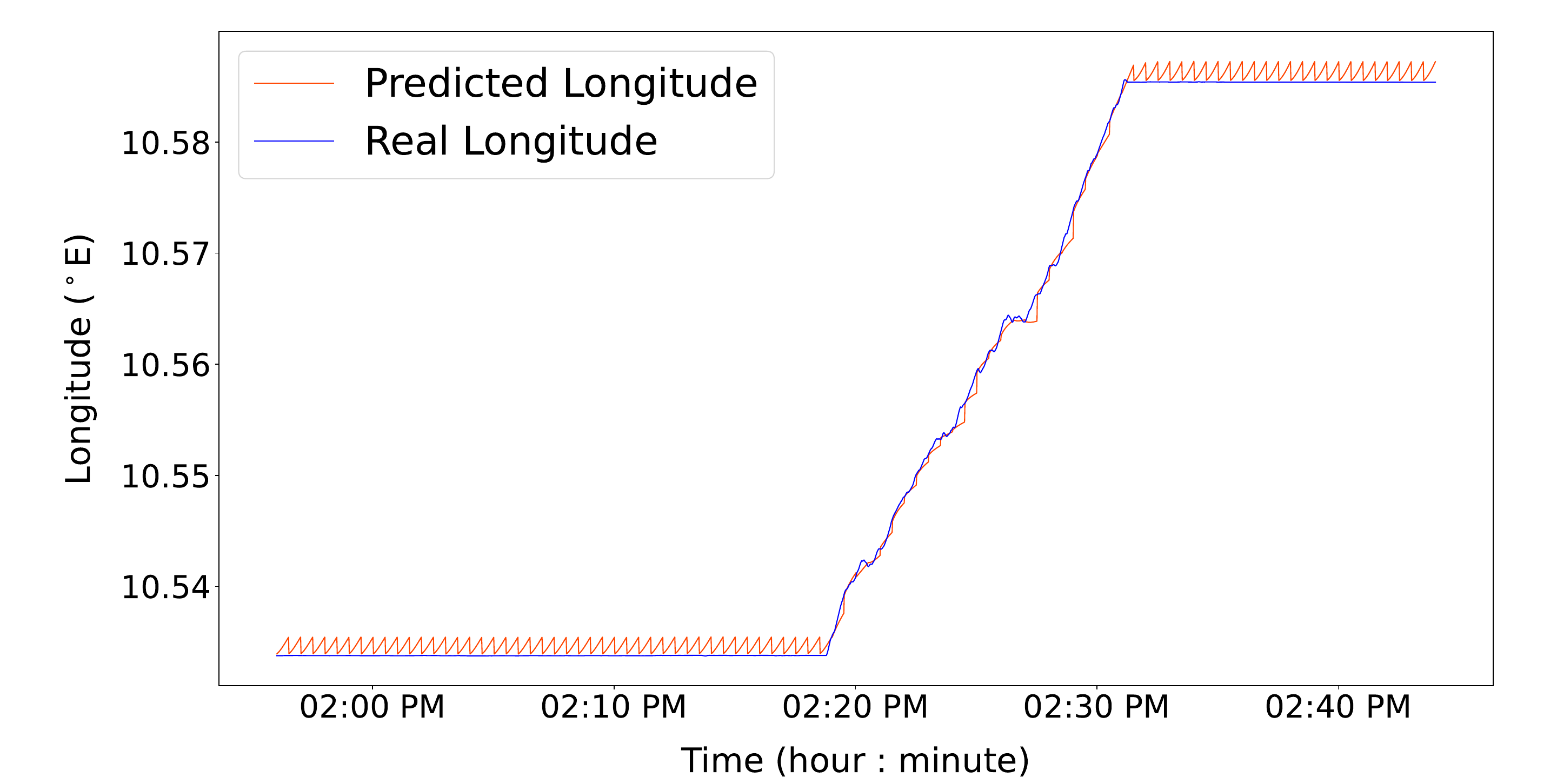}
         \caption{Longitude prediction during the forth test data set}
         \label{V-Long4}
     \end{subfigure}
     \hfill
     \begin{subfigure}[p]{0.495\textwidth}
         \centering
         \includegraphics[width=\textwidth]{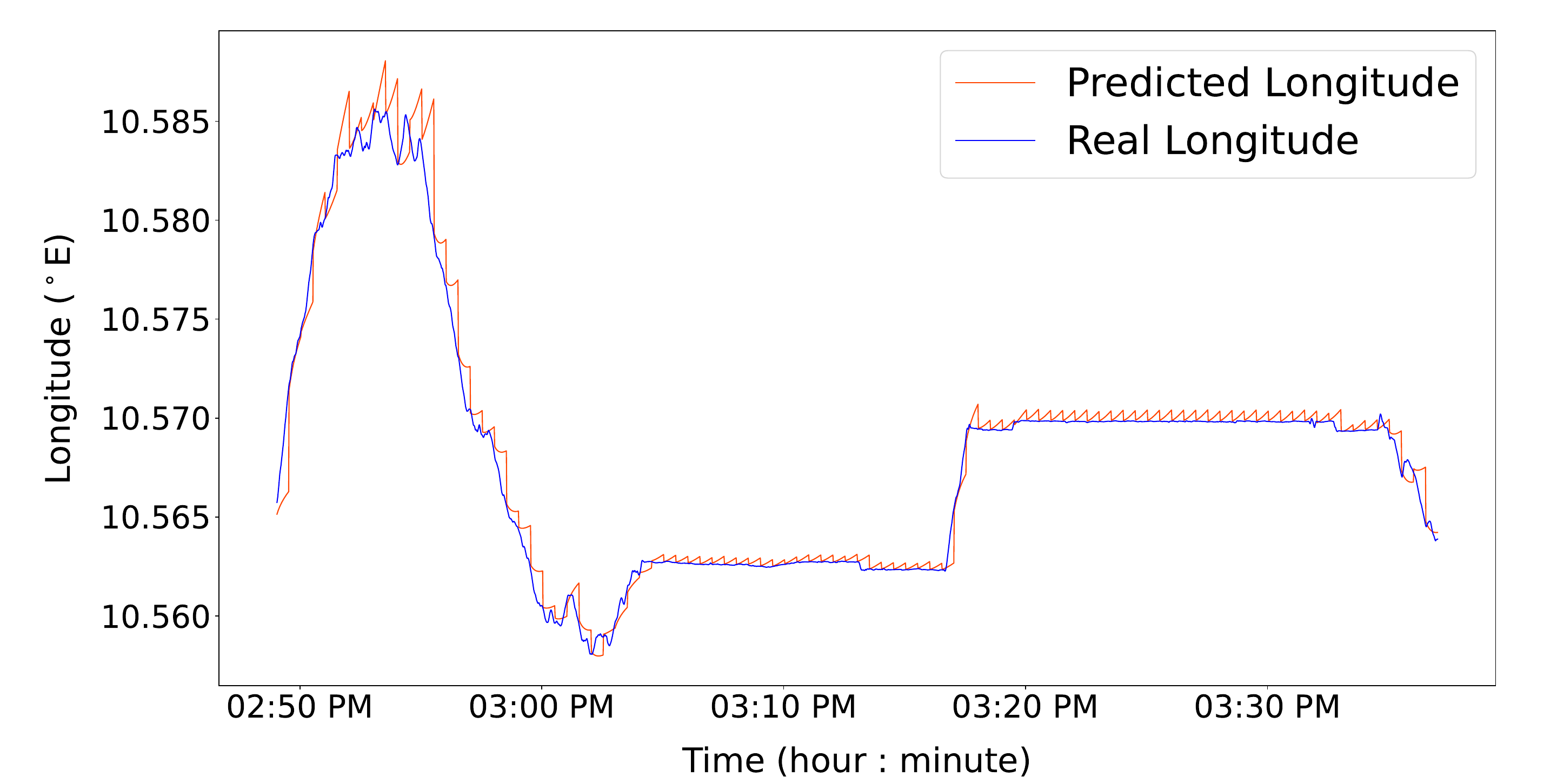}
         \caption{Longitude prediction during the fifth test data set}
         \label{V-Long5}
     \end{subfigure}
     \caption{Longitude predicted by vanilla LSTM}  
     \label{Fig: Longitude-Vanilla} 
\end{figure}

To predict bird movement using different LSTM models, we employed both many-to-one and many-to-many models to predict the next 30 time steps. We present the results of the vanilla LSTM model here, while the results of the other LSTM models (stacked, bidirectional, and encoder-decoder) are included in \ref{appendix} for completeness. For predicting latitude and longitude, we implemented two different vanilla LSTMs with 30 neurons for latitude prediction and 50 neurons for longitude prediction. The vanilla LSTMs were trained on the training data, which comprised numerous examples of 300 input data points as input horizon and one output data point as prediction horizon since it is a many-to-one model. We increased the prediction horizon of the vanilla LSTM through a recursive process. This approach involved using the model's output as input for the subsequent prediction in a step-by-step manner. By repeating this process recursively, we were able to generate a sequence of predictions for each time step in the interval, thus enabling us to create a complete trajectory of the bird's movement over time. The results indicate that increasing the prediction horizon led to some oscillations in the predictions of the vanilla LSTMs, as illustrated in Fig. \ref{Fig: Latitude-Vanilla} for latitude prediction and Fig. \ref{Fig: Longitude-Vanilla} for longitude prediction. Despite the oscillations, both models accurately predict patterns in latitude and longitude, even when significant changes occur within those patterns. For instance, in the third test data set related to latitude prediction, as shown in Fig. \ref{V-Lat3}, between 1:40:00 PM to 1:50:00 PM, the model recognized that a significant change was about to occur and accurately followed it. Similarly, in the first test data set for longitude prediction, as shown in Fig. \ref{V-Long1}, between 11:20:00 AM to 11:30:00 AM, there was a fluctuation that the model accurately captured. Tables \ref{tab:Latitude} and \ref{tab:Longitude} indicate the accuracy of all LSTM models were found to be less than 100 meters. 

For our training and test data sets, the performance of the bidirectional LSTM was better than the other LSTM models in latitude prediction. Although the stacked LSTM had two hidden layers, it did not perform more accurately compared to the vanilla LSTM in all test data sets for both latitude and longitude prediction. The performance of the encoder-decoder LSTM was the best compared to other LSTM models in longitude prediction, while it did not perform as well as others in latitude prediction.

Based on the results, we gathered the following insights. To predict bird movement accurately using the LSTM models, the training data should be representative of the bird's movement behavior. Otherwise, the LSTM models may perform worse while facing the data that they have not seen during training process. However, if LSTM models are trained properly, they can accurately predict movement over different time windows. 

\floatplacement{figure}{!p}
%################################################Longitude############################################
\floatplacement{figure}{!p}
%######################################################################################################
\newpage
\subsubsection{Regression Models}
\label{R-S-4}
\begin{figure}
     \centering
     \begin{subfigure}[h]{0.495\textwidth}
         \centering
         \includegraphics[width=\textwidth]{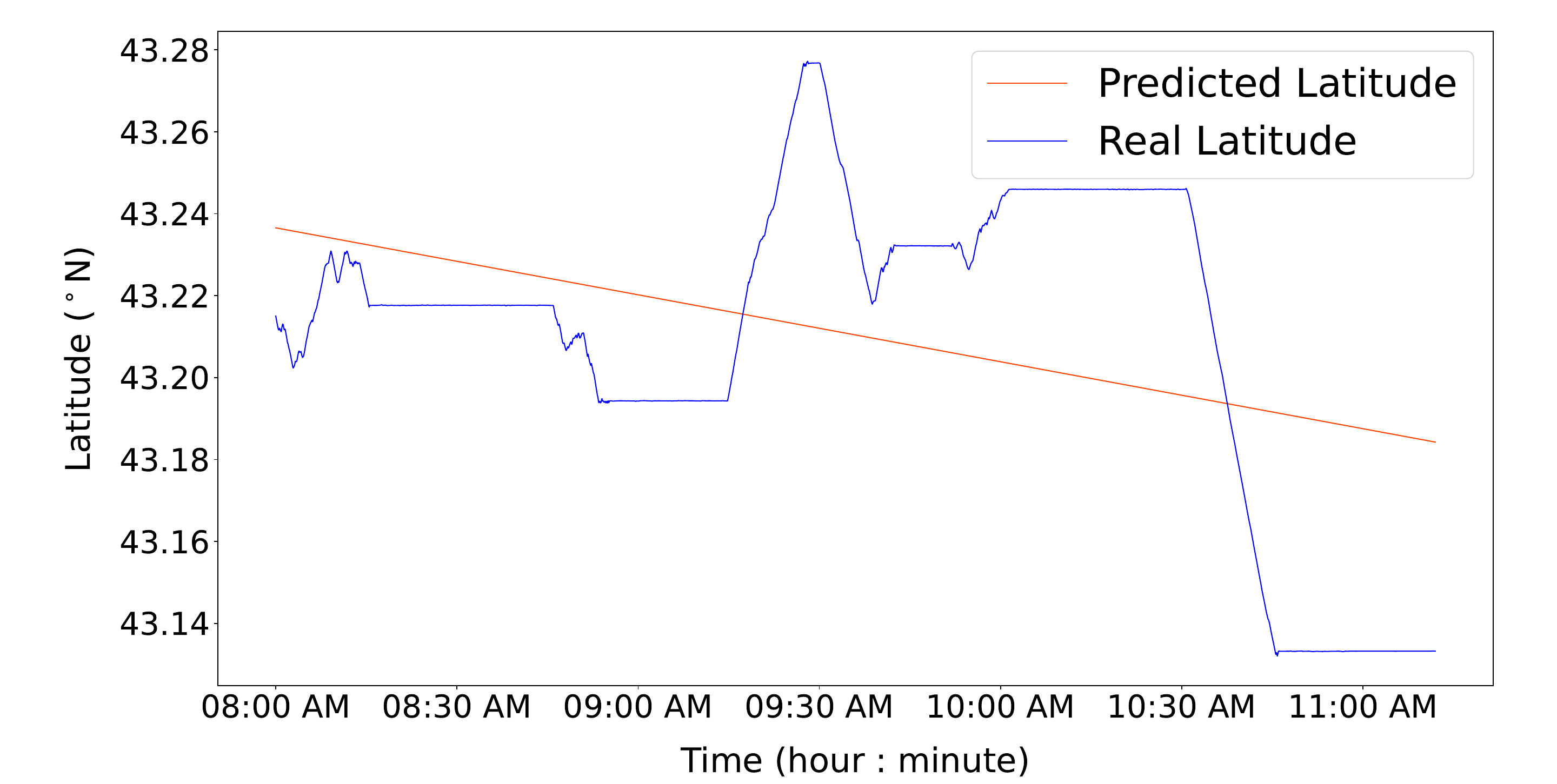}
         \caption{Latitude prediction during training}
         \label{L-Lat-T}
     \end{subfigure}
     \hfill
     \begin{subfigure}[h]{0.495\textwidth}
         \centering
         \includegraphics[width=\textwidth]{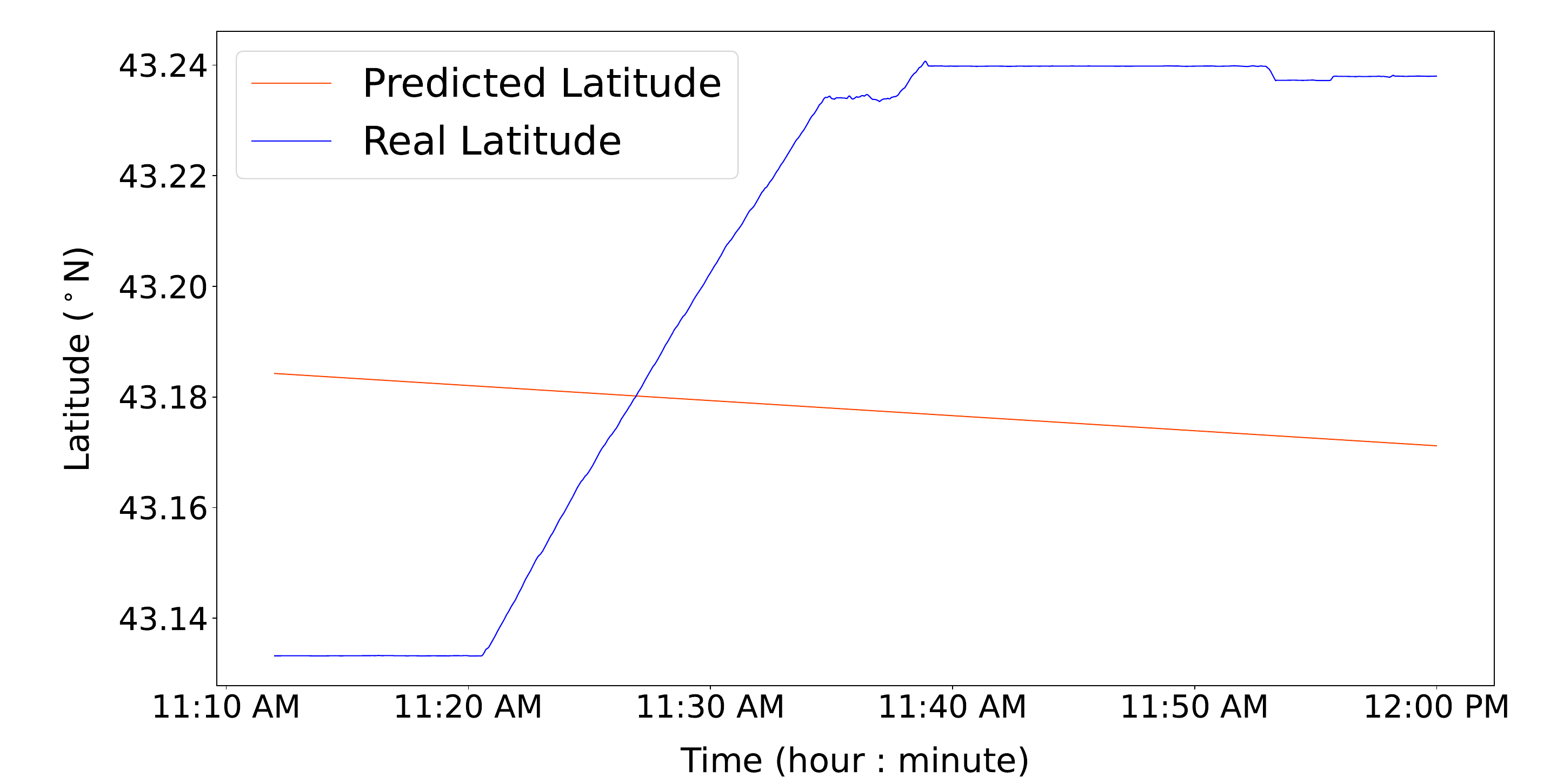}
         \caption{Latitude prediction during the first test data set}
         \label{L-Lat1}
     \end{subfigure}
%###########################
     \centering
     \begin{subfigure}[h]{0.495\textwidth}
         \centering
         \includegraphics[width=\textwidth]{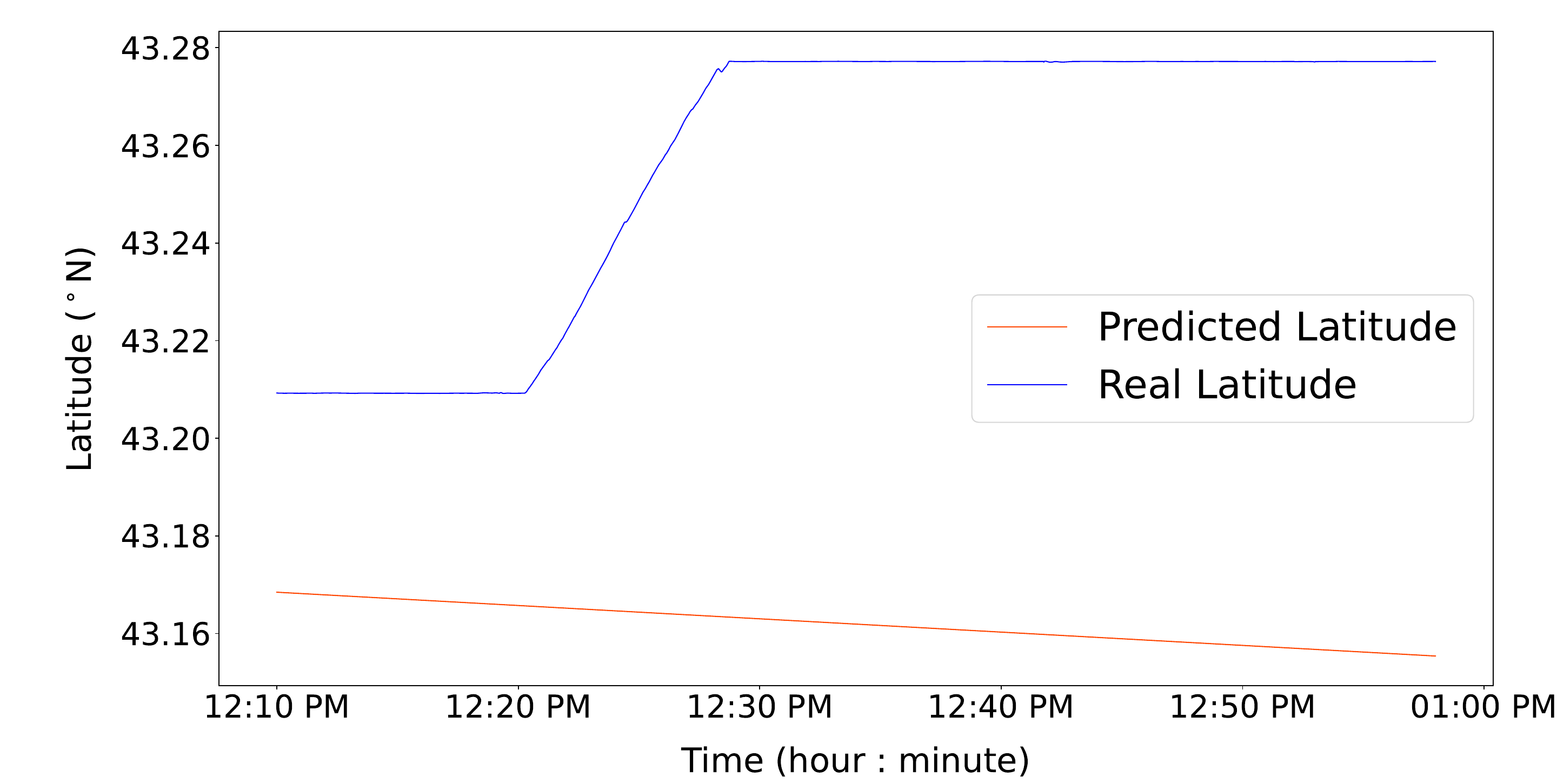}
         \caption{Latitude prediction during the second test data set}
         \label{L-Lat2}
     \end{subfigure}
     \hfill
     \begin{subfigure}[h]{0.495\textwidth}
         \centering
         \includegraphics[width=\textwidth]{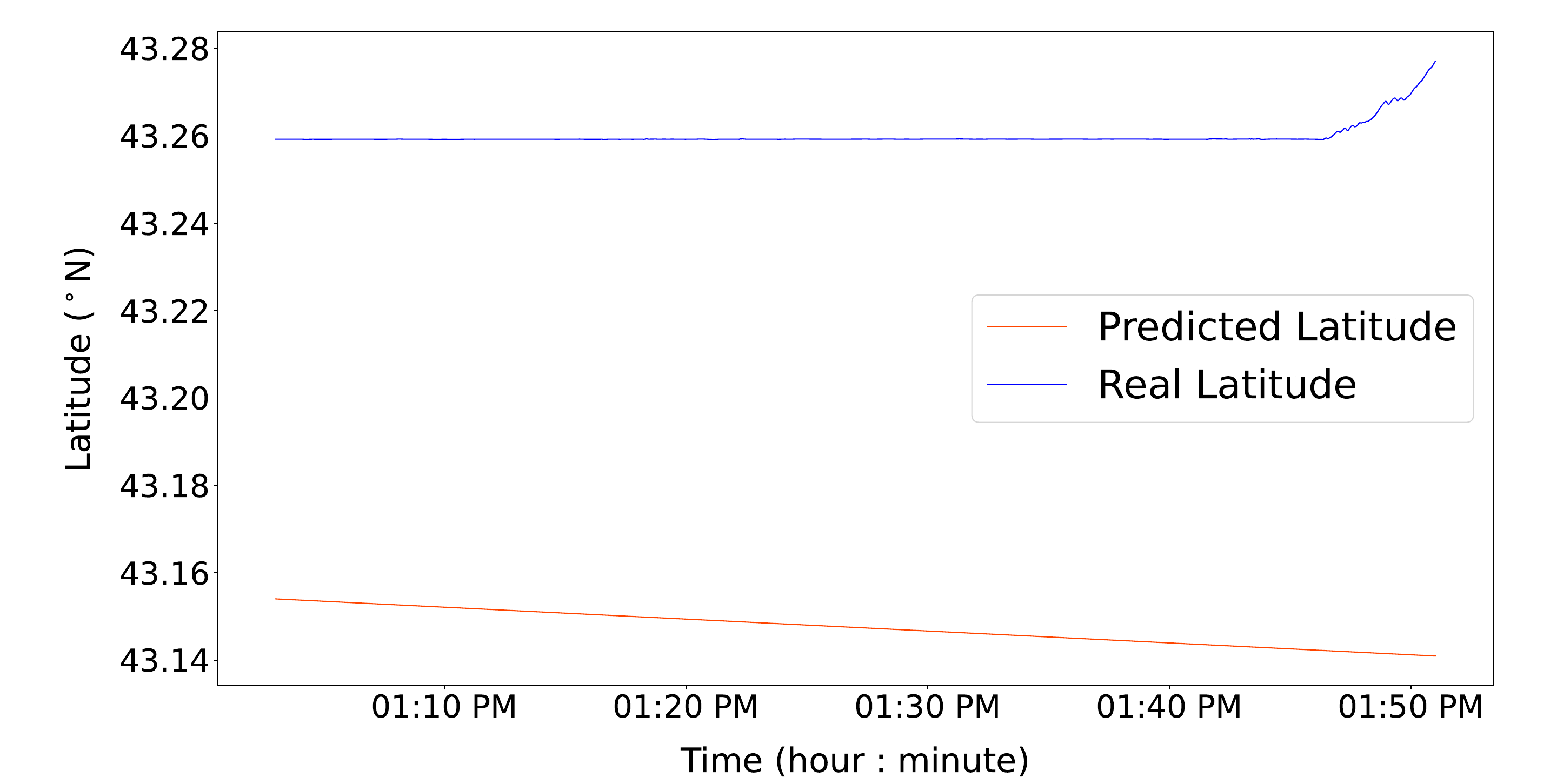}
         \caption{Latitude prediction during the third test data set}
         \label{L-Lat3}
     \end{subfigure}
%###########################
     \centering
     \begin{subfigure}[h]{0.495\textwidth}
         \centering
         \includegraphics[width=\textwidth]{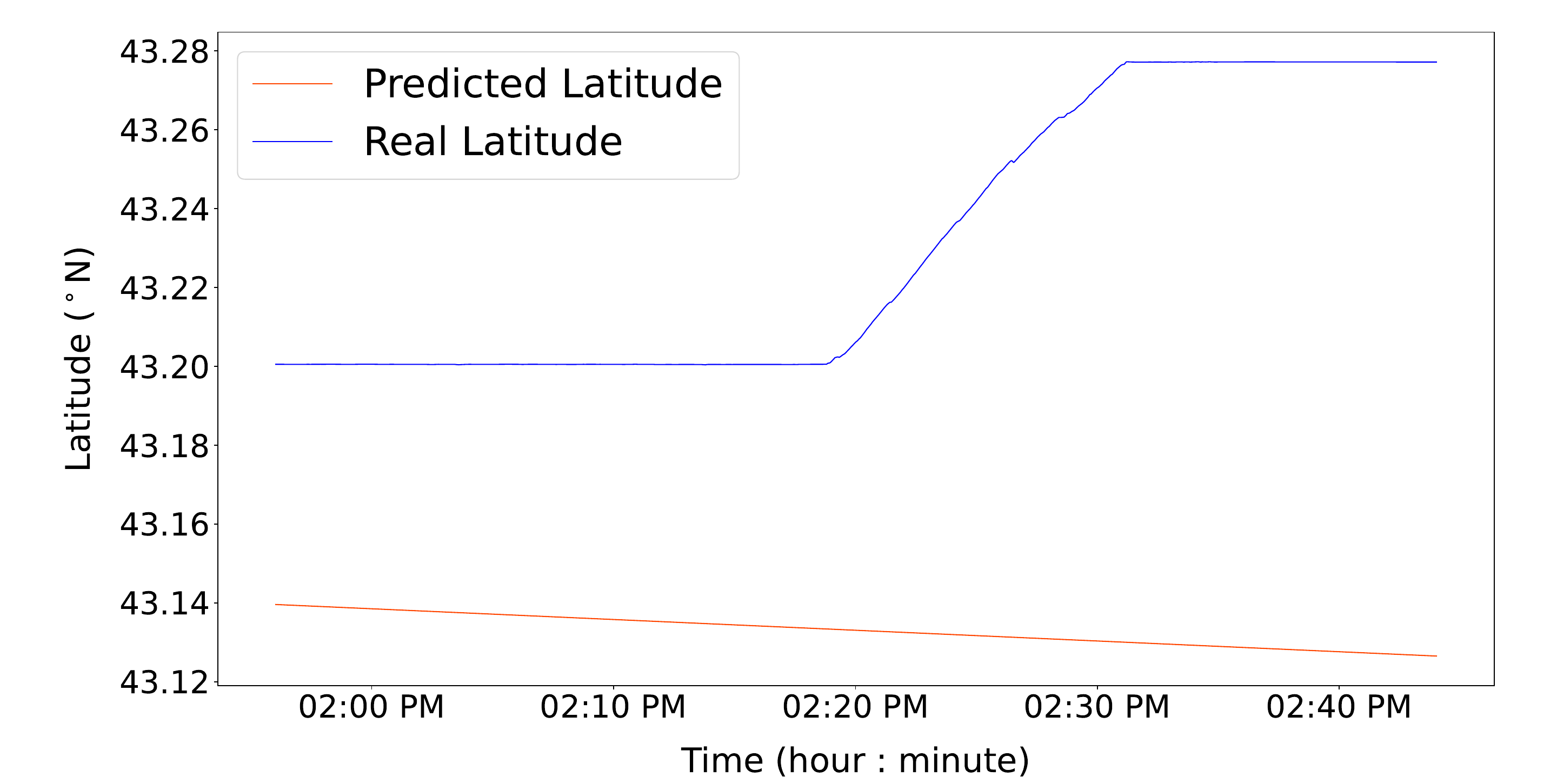}
         \caption{Latitude prediction during the forth test data set}
         \label{L-Lat4}
     \end{subfigure}
     \hfill
     \begin{subfigure}[h]{0.495\textwidth}
         \centering
         \includegraphics[width=\textwidth]{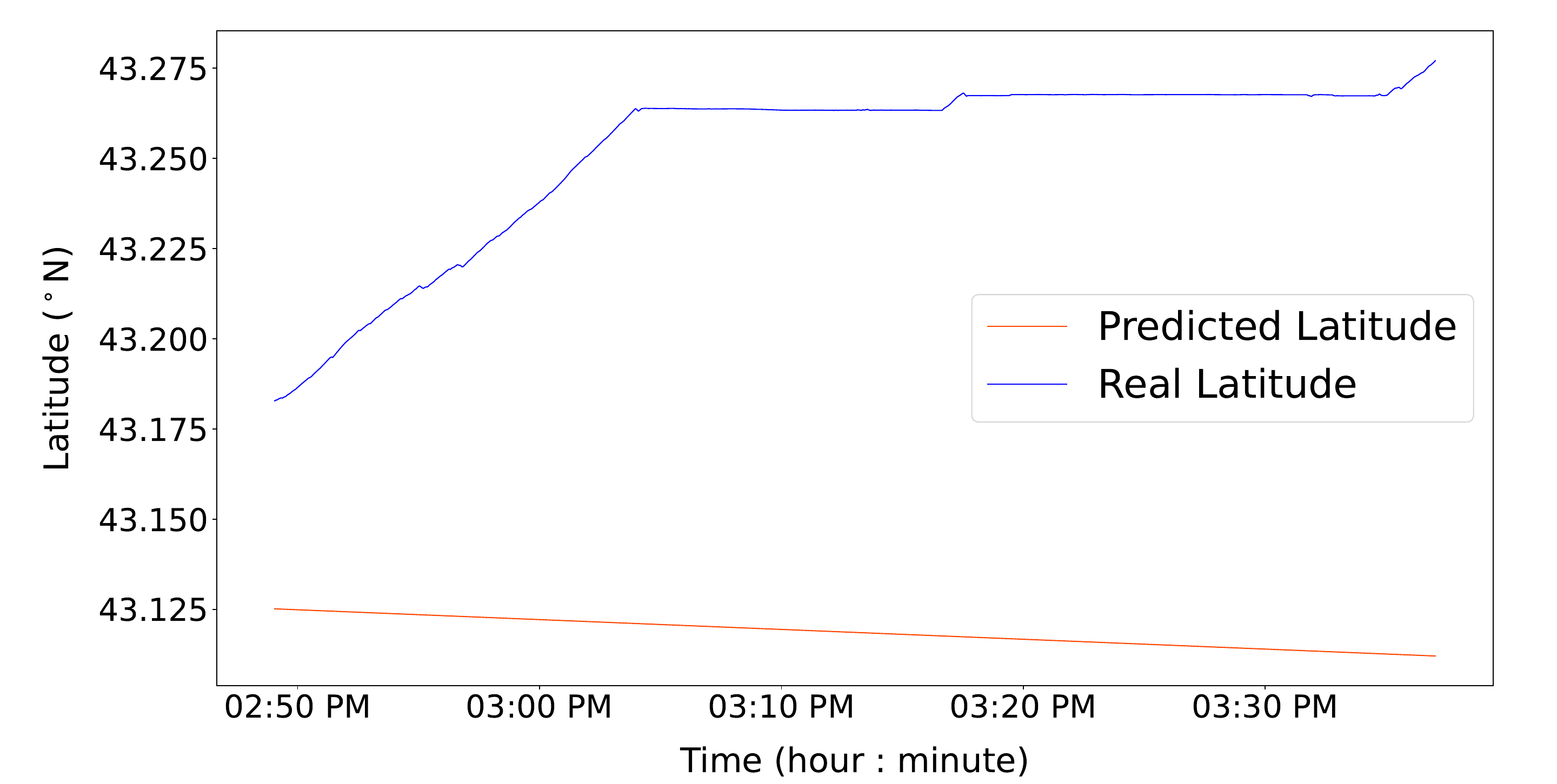}
         \caption{Latitude prediction during the fifth test data set}
         \label{L-Lat5}
     \end{subfigure}
     \caption{Latitude predicted by linear regression model}
     \label{Linear-Latitude}
\end{figure}

\begin{figure}
     \centering
     \begin{subfigure}[p]{0.495\textwidth}
         \centering
         \includegraphics[width=\textwidth]{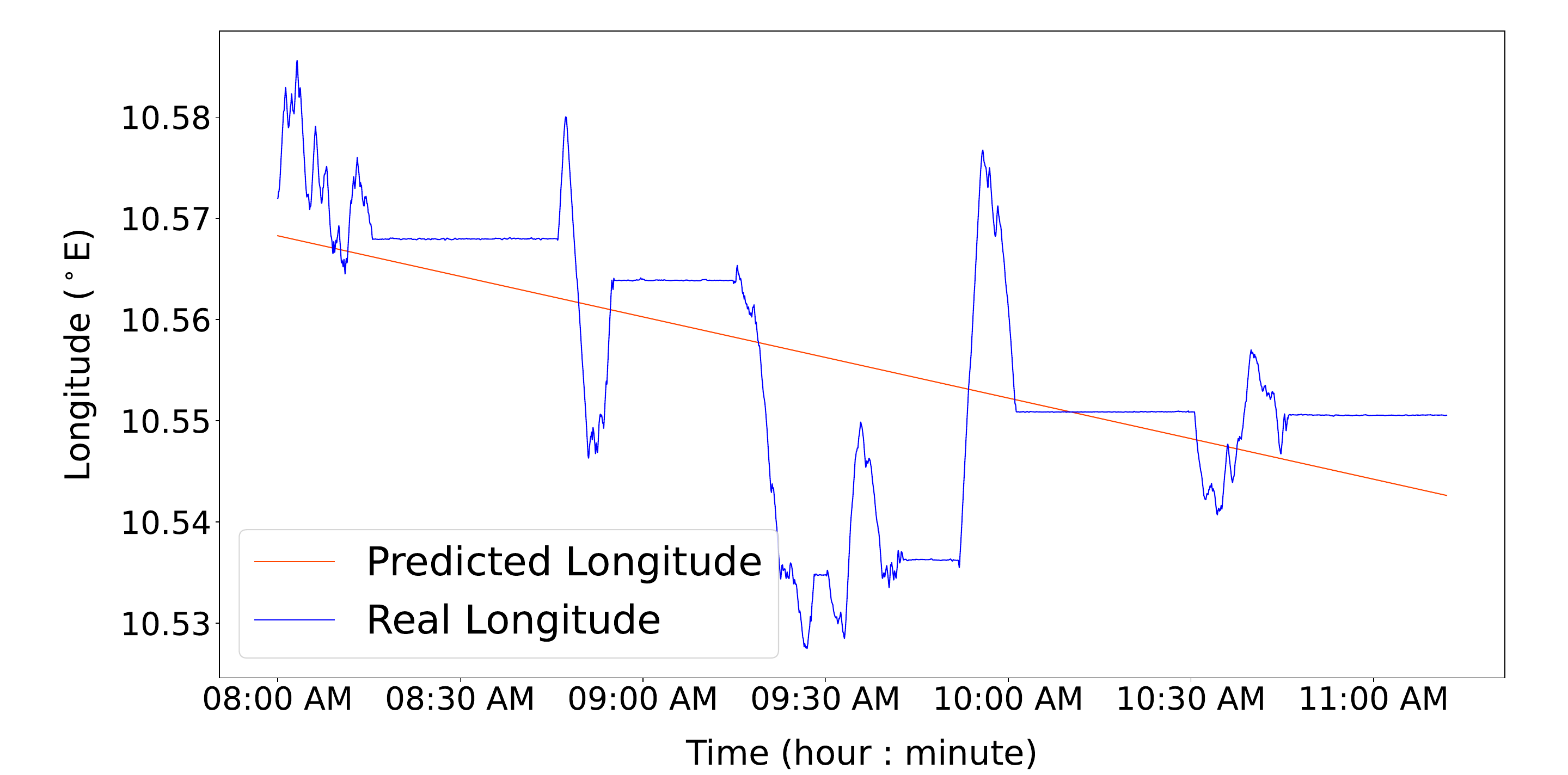}
         \caption{Longitude prediction during training}
         \label{L-Long-T}
     \end{subfigure}
     \hfill
     \begin{subfigure}[p]{0.495\textwidth}
         \centering
         \includegraphics[width=\textwidth]{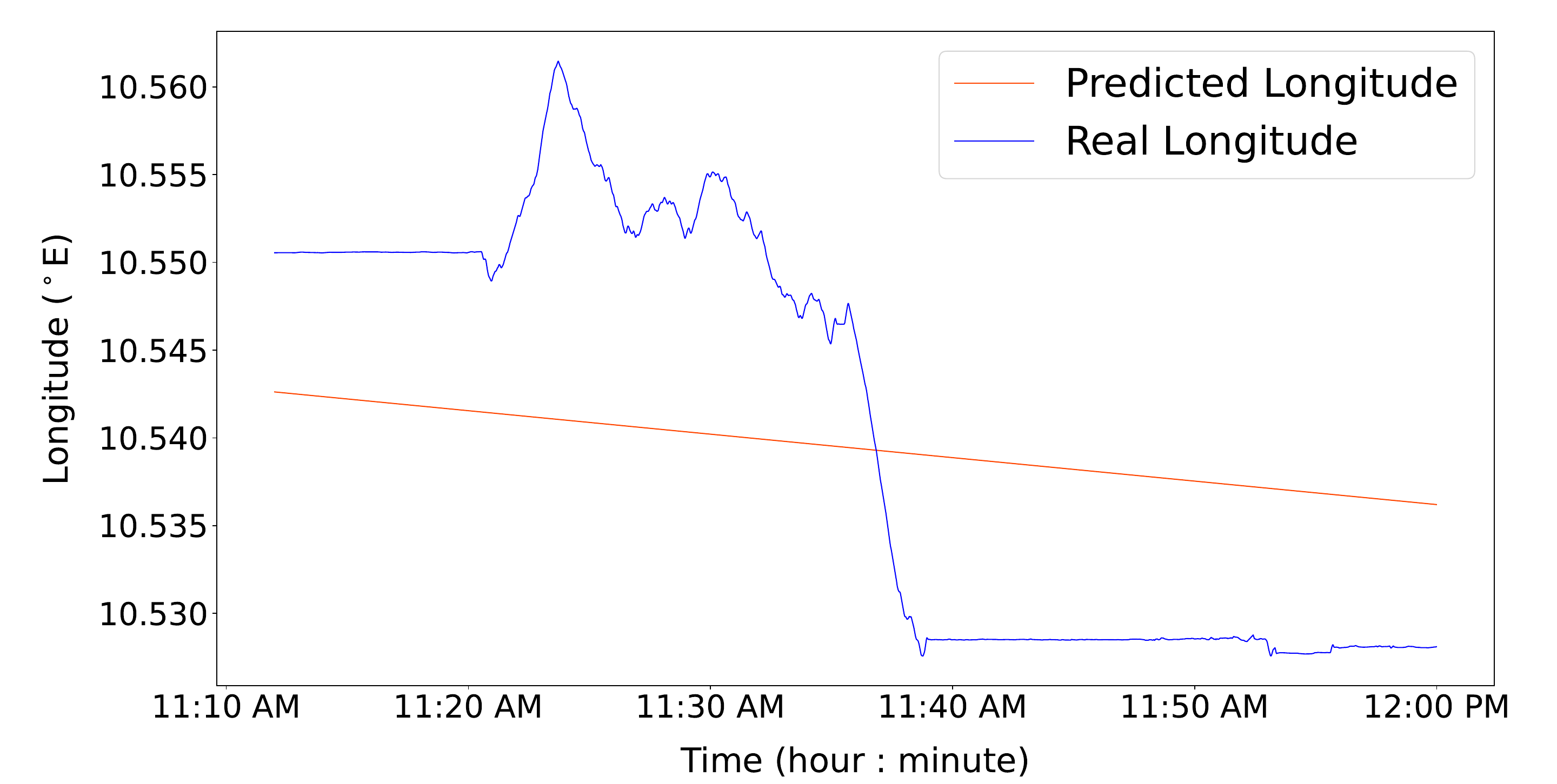}
         \caption{Longitude prediction during the first test data set}
         \label{L-Long1}
     \end{subfigure}
%###########################
     \centering
     \begin{subfigure}[p]{0.495\textwidth}
         \centering
         \includegraphics[width=\textwidth]{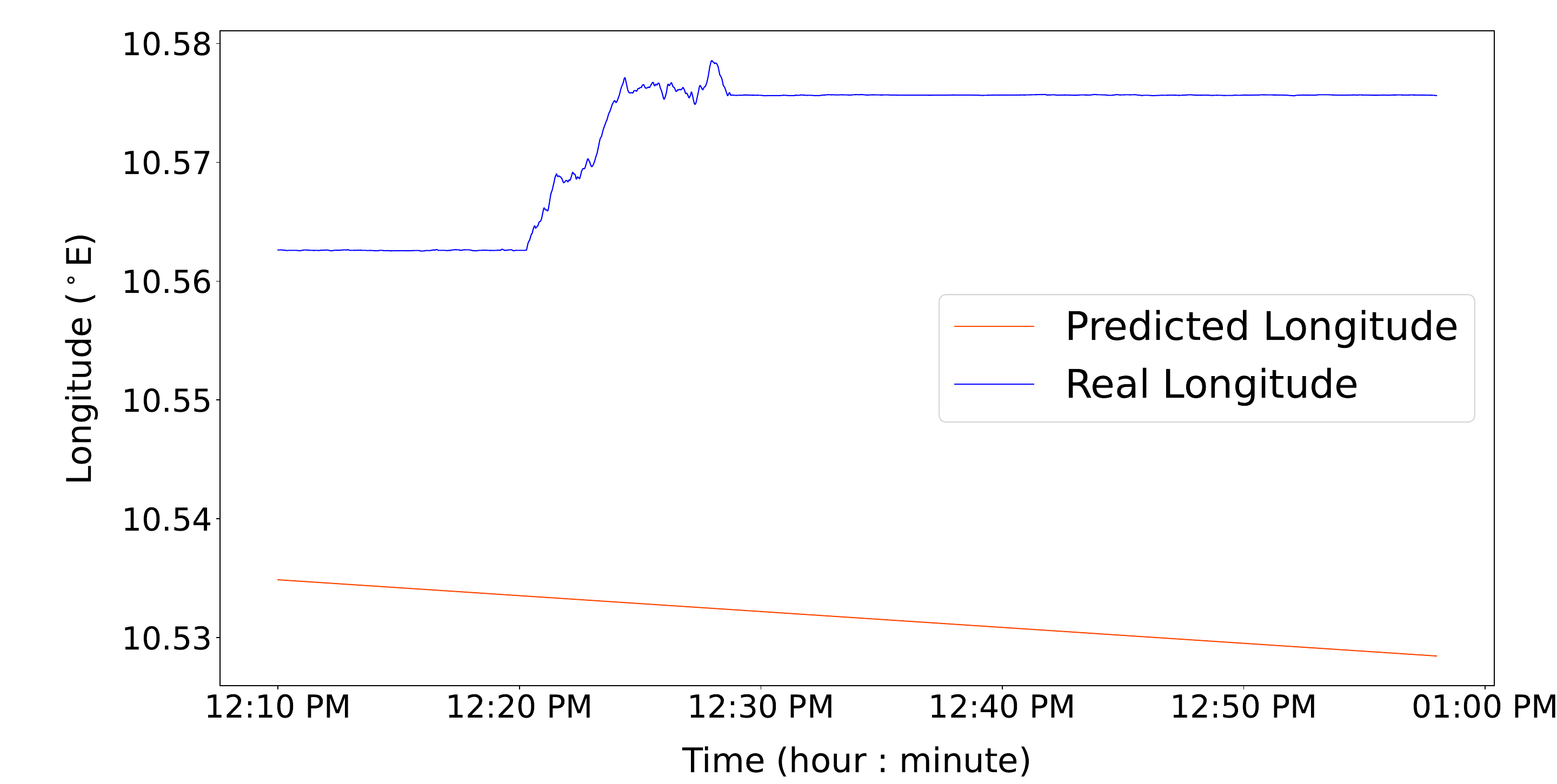}
         \caption{Longitude prediction during the second test data set}
         \label{L-Long2}
     \end{subfigure}
     \hfill
     \begin{subfigure}[p]{0.495\textwidth}
         \centering
         \includegraphics[width=\textwidth]{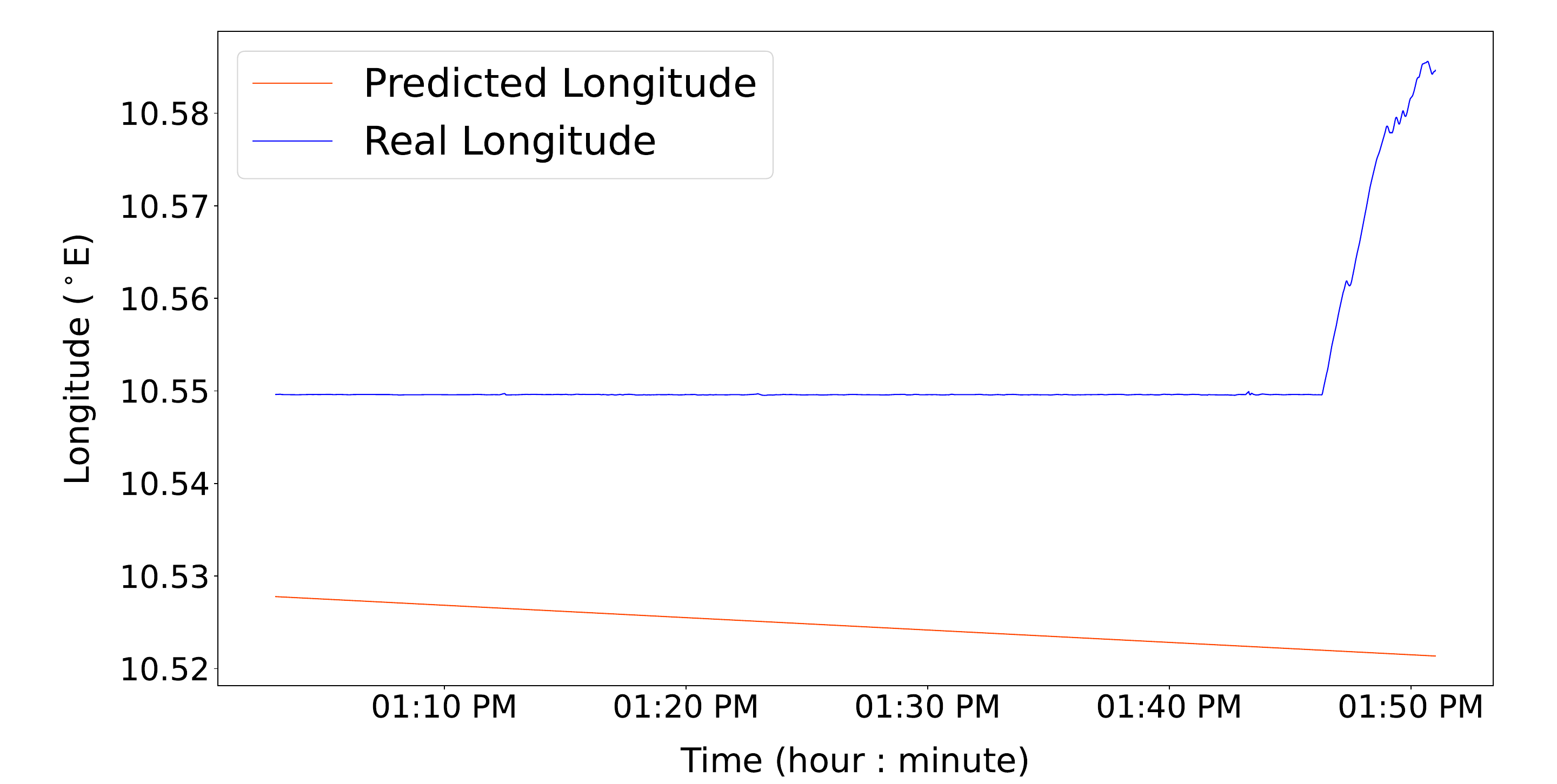}
         \caption{Longitude prediction during the third test data set}
         \label{L-Long3}
     \end{subfigure}
%###########################
     \centering
     \begin{subfigure}[p]{0.495\textwidth}
         \centering
         \includegraphics[width=\textwidth]{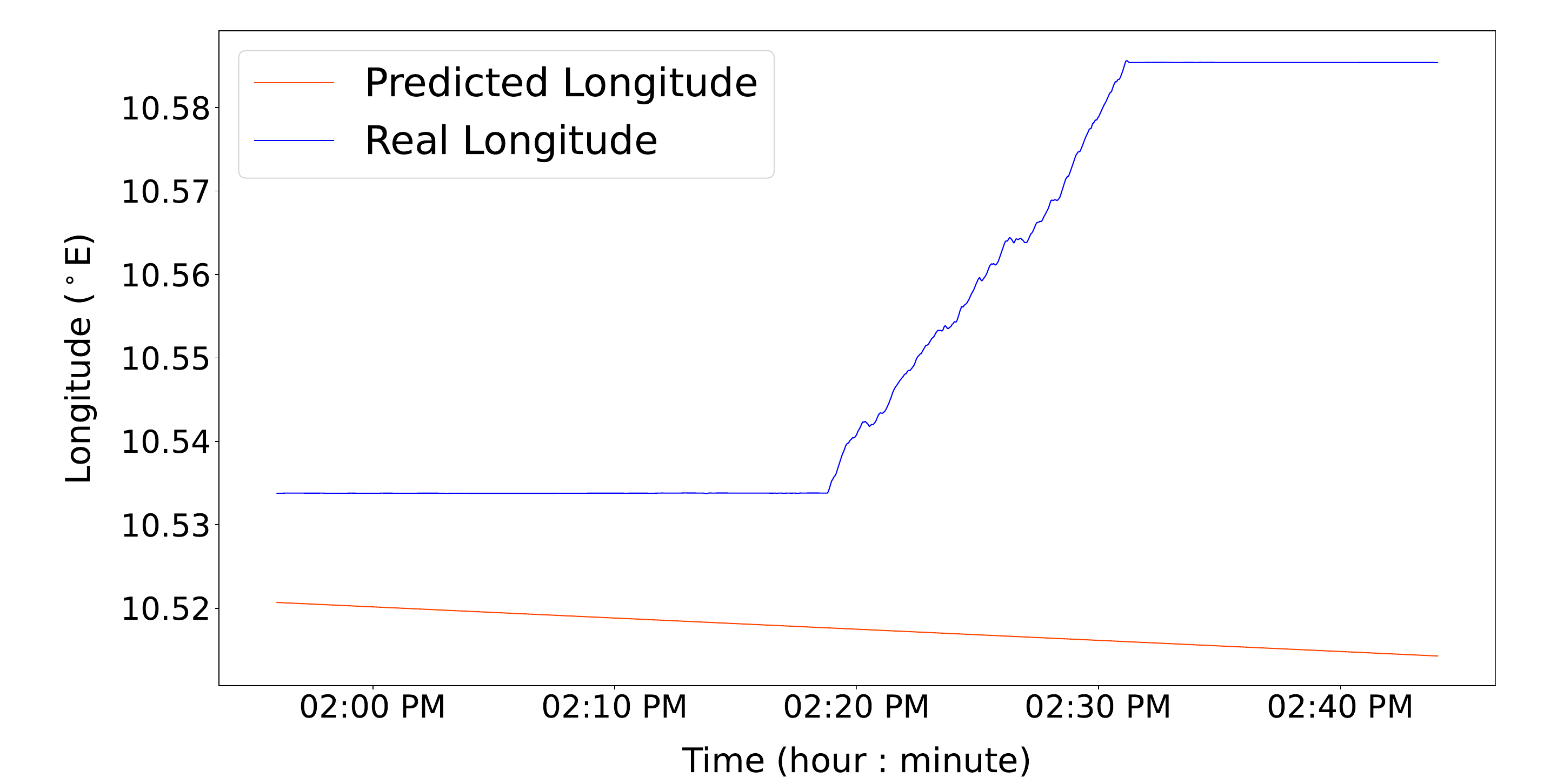}
         \caption{Longitude prediction during the forth test data set}
         \label{L-Long4}
     \end{subfigure}
     \hfill
     \begin{subfigure}[p]{0.495\textwidth}
         \centering
         \includegraphics[width=\textwidth]{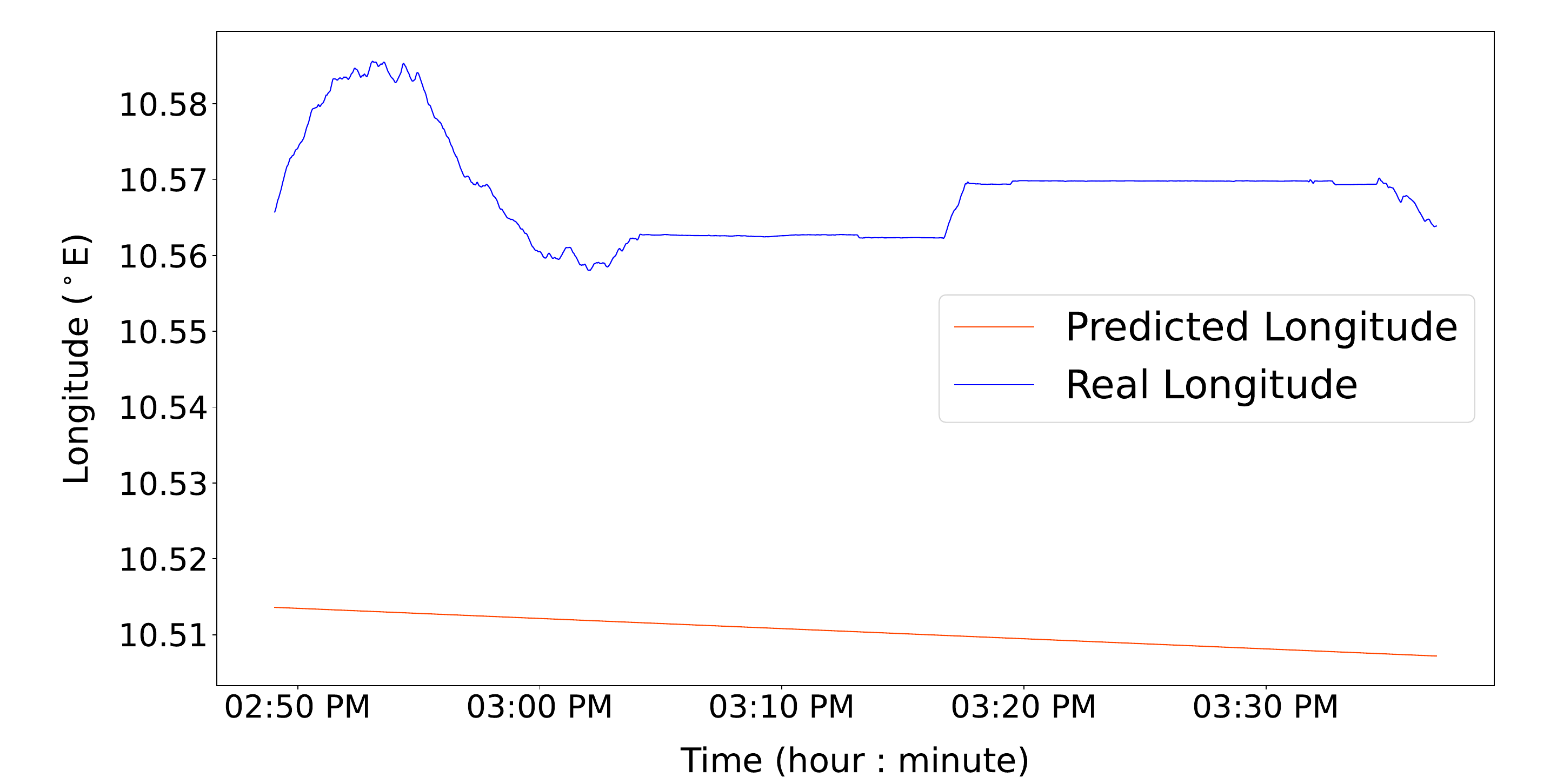}
         \caption{Longitude prediction during the fifth test data set}
         \label{L-Long5}
     \end{subfigure}
     \caption{Longitude predicted by linear regression model}
     \label{Linear-Longitude}
\end{figure}

A linear regression model has been used in the literature to predict bird movement, and it could serve as a good benchmark model. It is a basic model that is commonly used for prediction problems. Although it is simple to implement, it may not be able to accurately predict variations and oscillations since it tries to fit the data to a straight line and may not account for changes in different test data sets. Figs. \ref{L-Lat-T} and \ref{L-Long-T} show the straight lines fitted to the training data for latitude and longitude, respectively. However, when the model is applied to the test data sets, a significant gap can be observed between the predicted and real latitude values in Fig. \ref{Linear-Latitude}. Table \ref{tab:Latitude} illustrates that the model's performance in predicting latitude is inaccurate, with an MAE of more than 5600 meters for all test data sets. As shown in Fig. \ref{Linear-Longitude}, a divergence can be observed in all test data sets except for the first one, in the case of longitude. Table \ref{tab:Longitude} confirms that the model's performance is better in the first test data set, with an MAE of less than 1000 meters, since it did not face any divergence. However, when compared to LSTM models, this error is not acceptable.
\floatplacement{figure}{!h}
%################################################Longitude#################################
\floatplacement{figure}{!p}
%######################################################################################################

\begin{figure}
     \centering
     \begin{subfigure}[h]{0.495\textwidth}
         \centering
         \includegraphics[width=\textwidth]{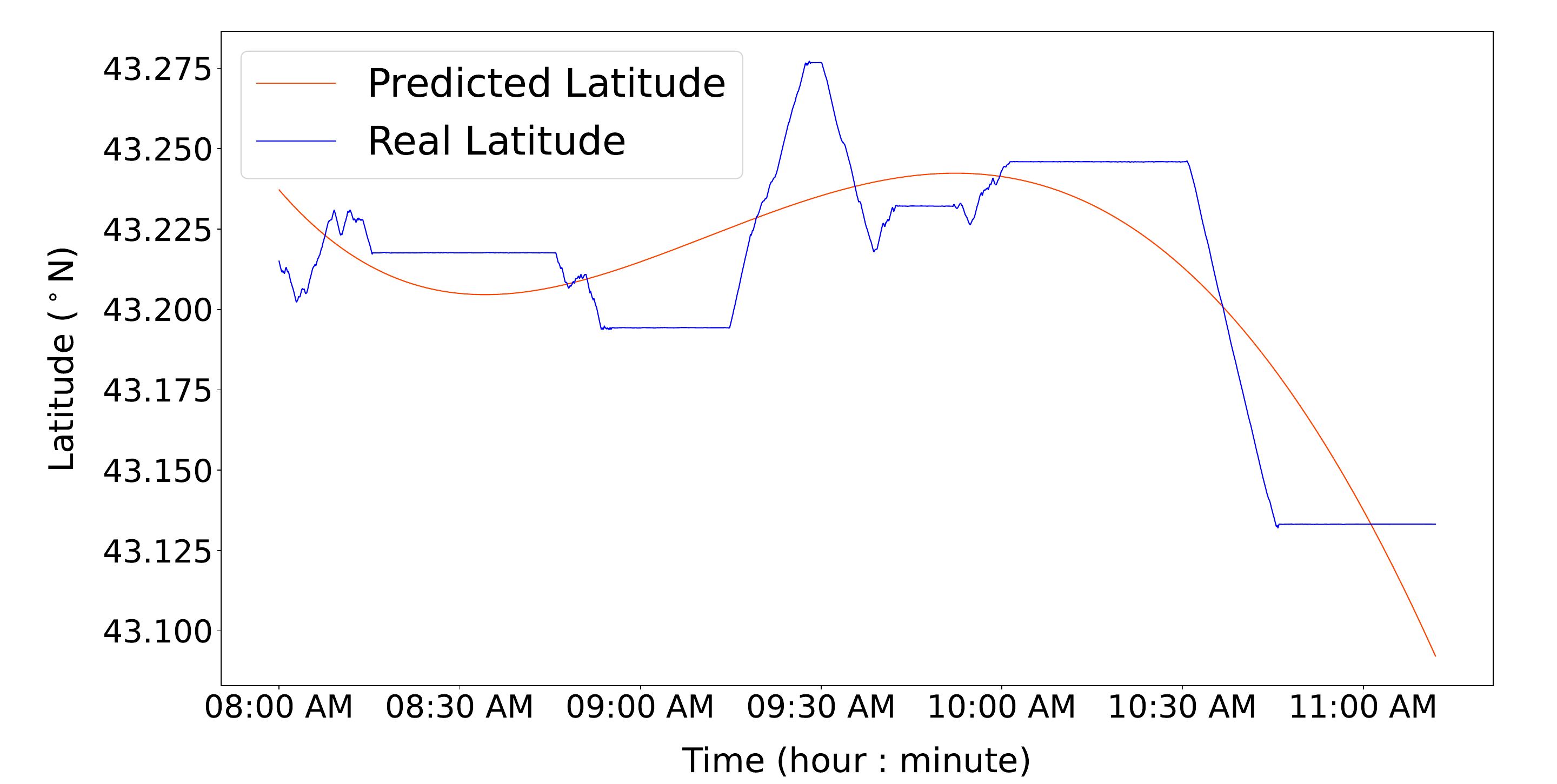}
         \caption{Latitude prediction during training}
         \label{N-Lat-T}
     \end{subfigure}
     \hfill
     \begin{subfigure}[h]{0.495\textwidth}
         \centering
         \includegraphics[width=\textwidth]{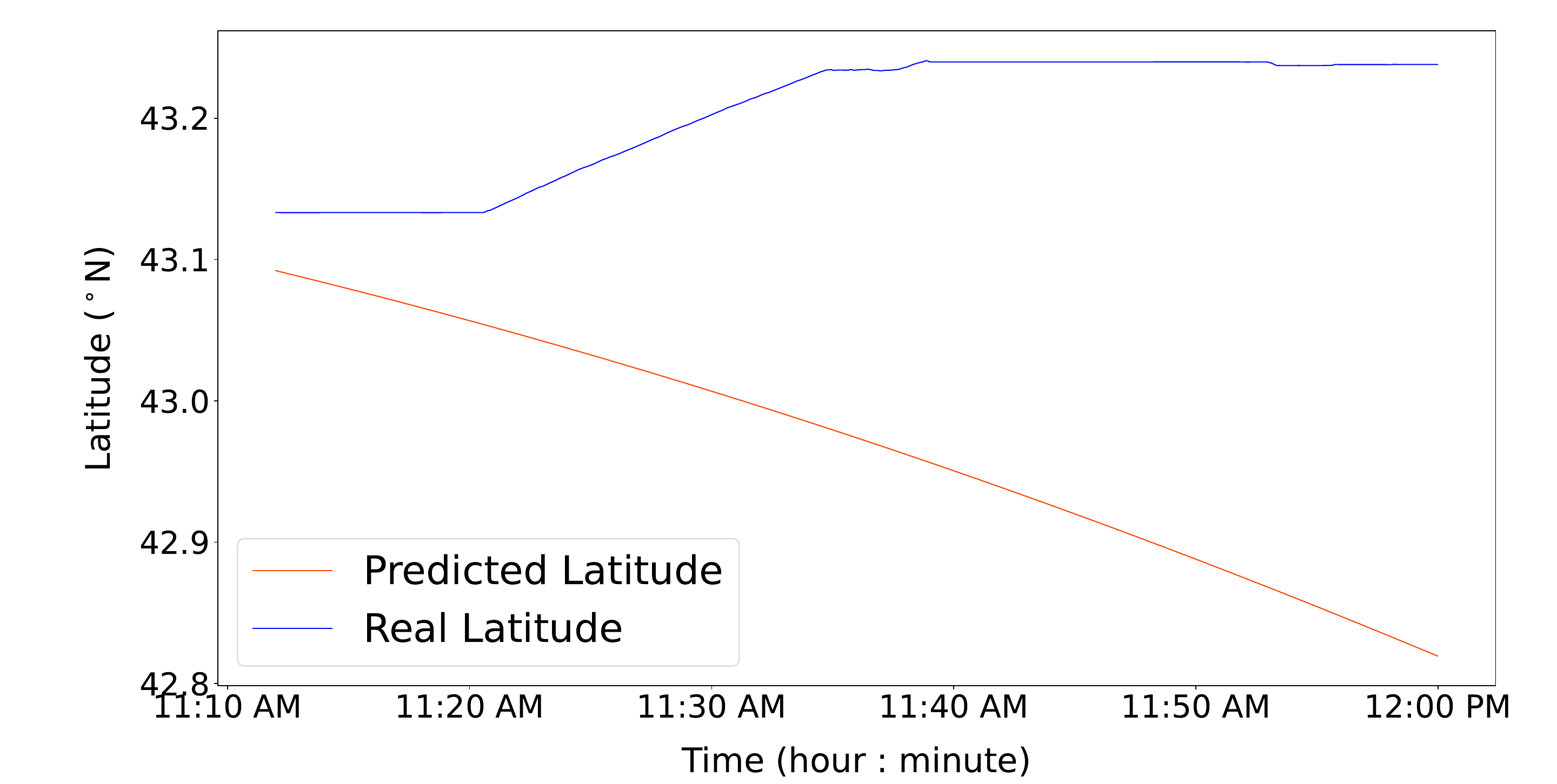}
         \caption{Latitude prediction during the first test data set}
         \label{N-Lat1}
     \end{subfigure}
%###########################
     \centering
     \begin{subfigure}[h]{0.495\textwidth}
         \centering
         \includegraphics[width=\textwidth]{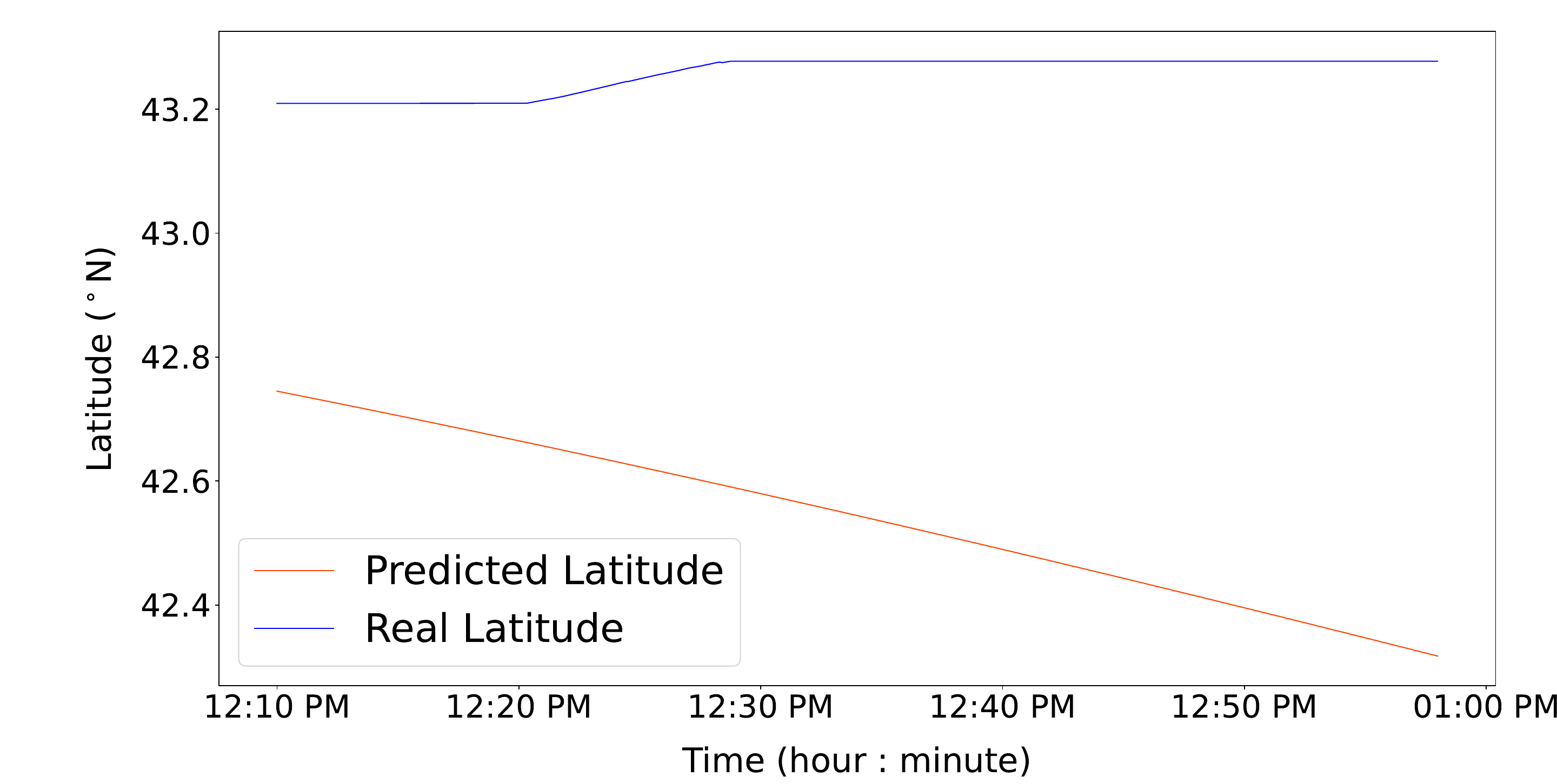}
         \caption{Latitude prediction during the second test data set}
         \label{N-Lat2}
     \end{subfigure}
     \hfill
     \begin{subfigure}[h]{0.495\textwidth}
         \centering
         \includegraphics[width=\textwidth]{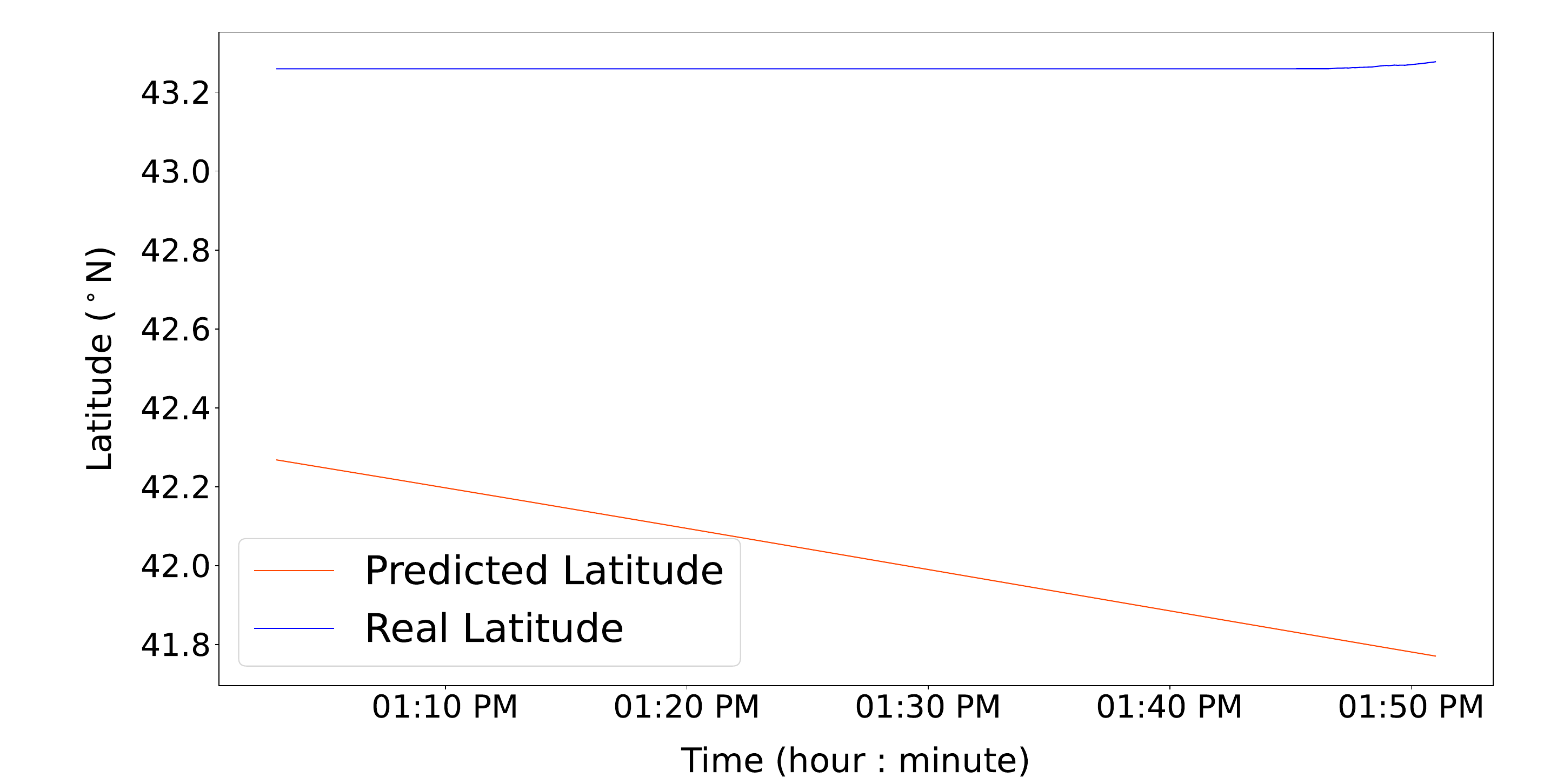}
         \caption{Latitude prediction during the third test data set}
         \label{N-Lat3}
     \end{subfigure}
%###########################
     \centering
     \begin{subfigure}[p]{0.495\textwidth}
         \centering
         \includegraphics[width=\textwidth]{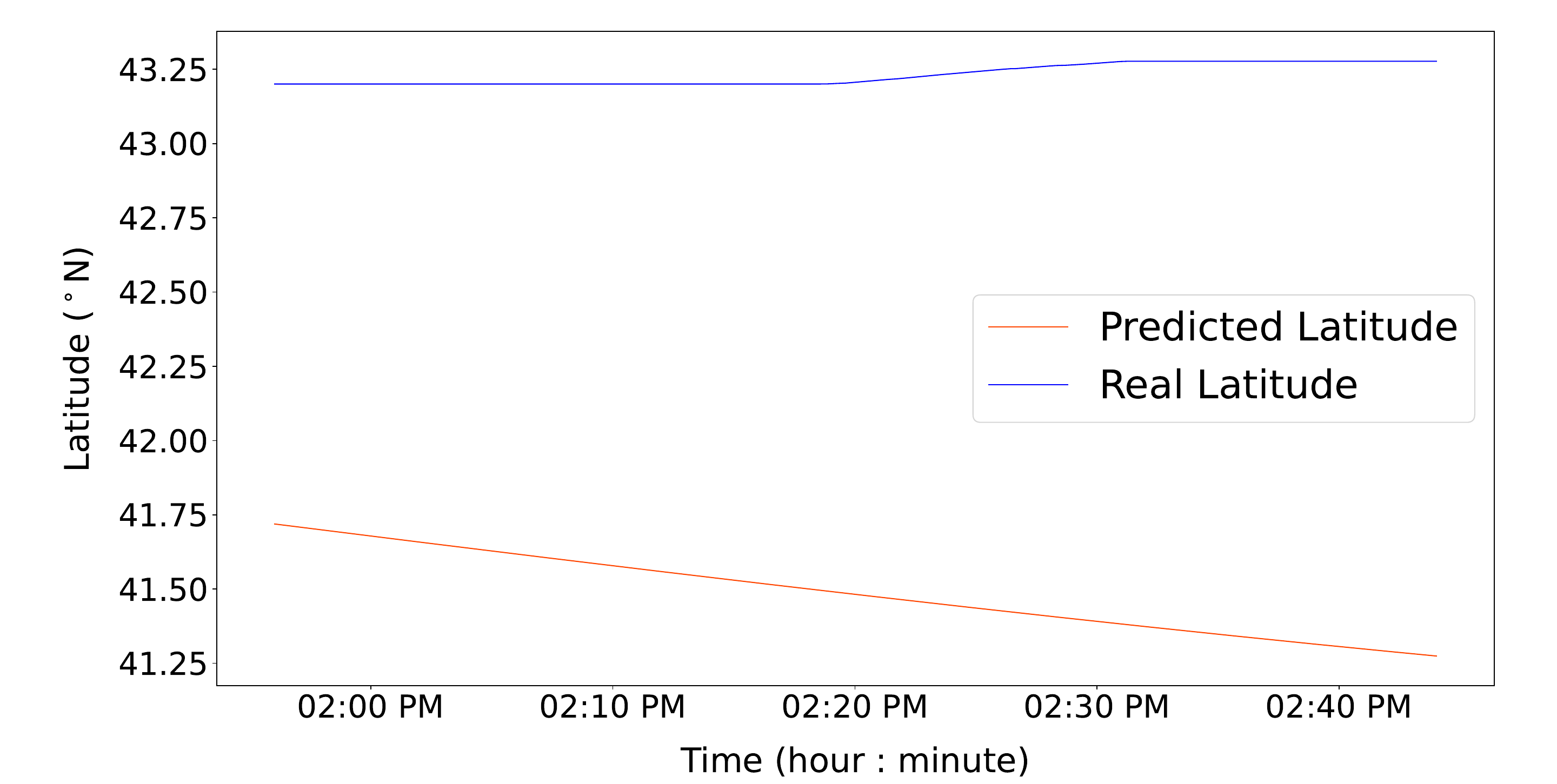}
         \caption{Latitude prediction during the forth test data set}
         \label{N-Lat4}
     \end{subfigure}
     \hfill
     \begin{subfigure}[p]{0.495\textwidth}
         \centering
         \includegraphics[width=\textwidth]{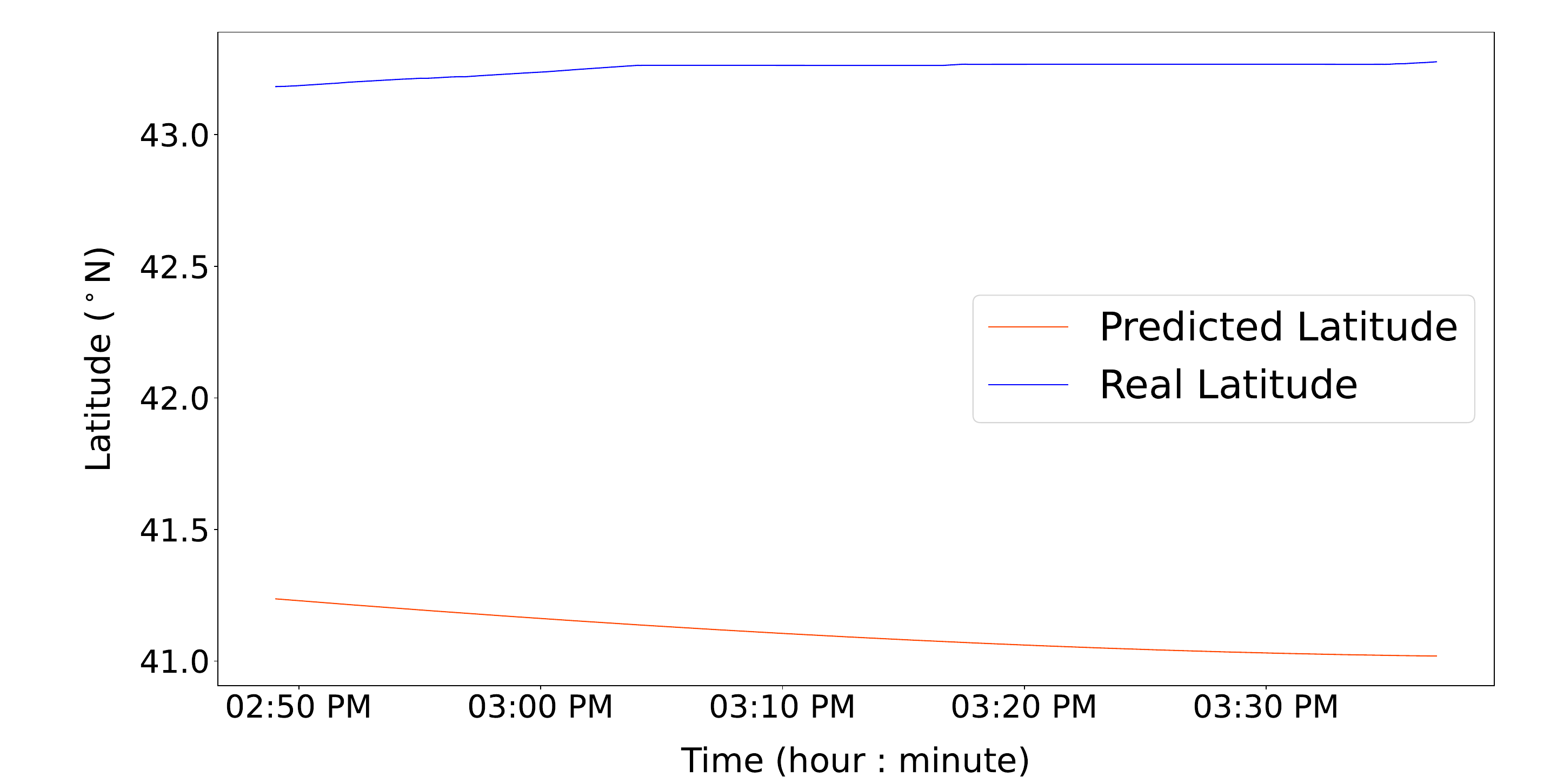}
         \caption{Latitude prediction during the fifth test data set}
         \label{N-Lat5}
     \end{subfigure}
     \caption{Latitude predicted by nonlinear regression model}
     \label{Nonlinear-Latitude}
\end{figure}

\begin{figure}
     \centering
     \begin{subfigure}[p]{0.495\textwidth}
         \centering
         \includegraphics[width=\textwidth]{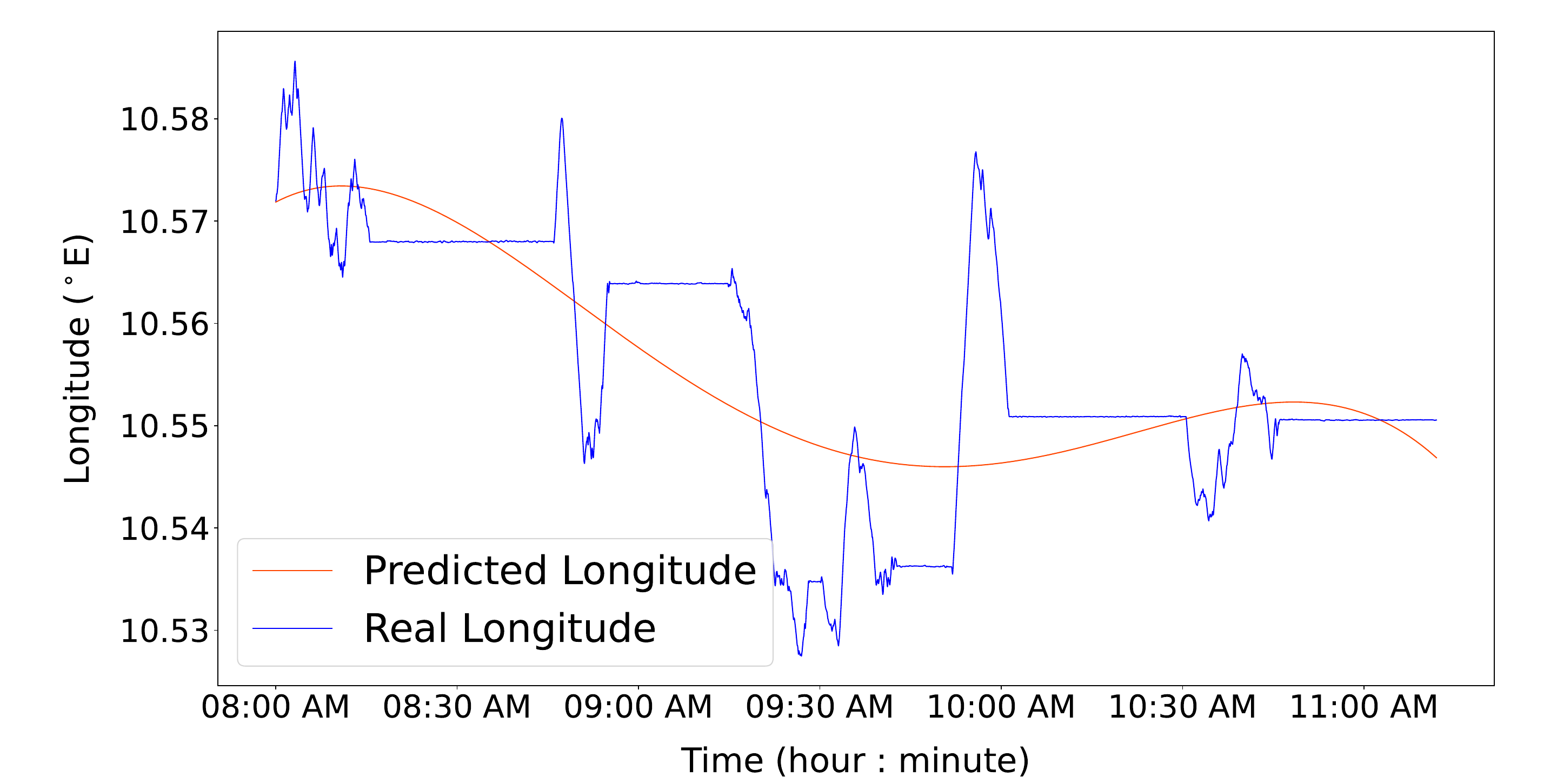}
         \caption{Longitude prediction during training}
         \label{N-Long-T}
     \end{subfigure}
     \hfill
     \begin{subfigure}[p]{0.495\textwidth}
         \centering
         \includegraphics[width=\textwidth]{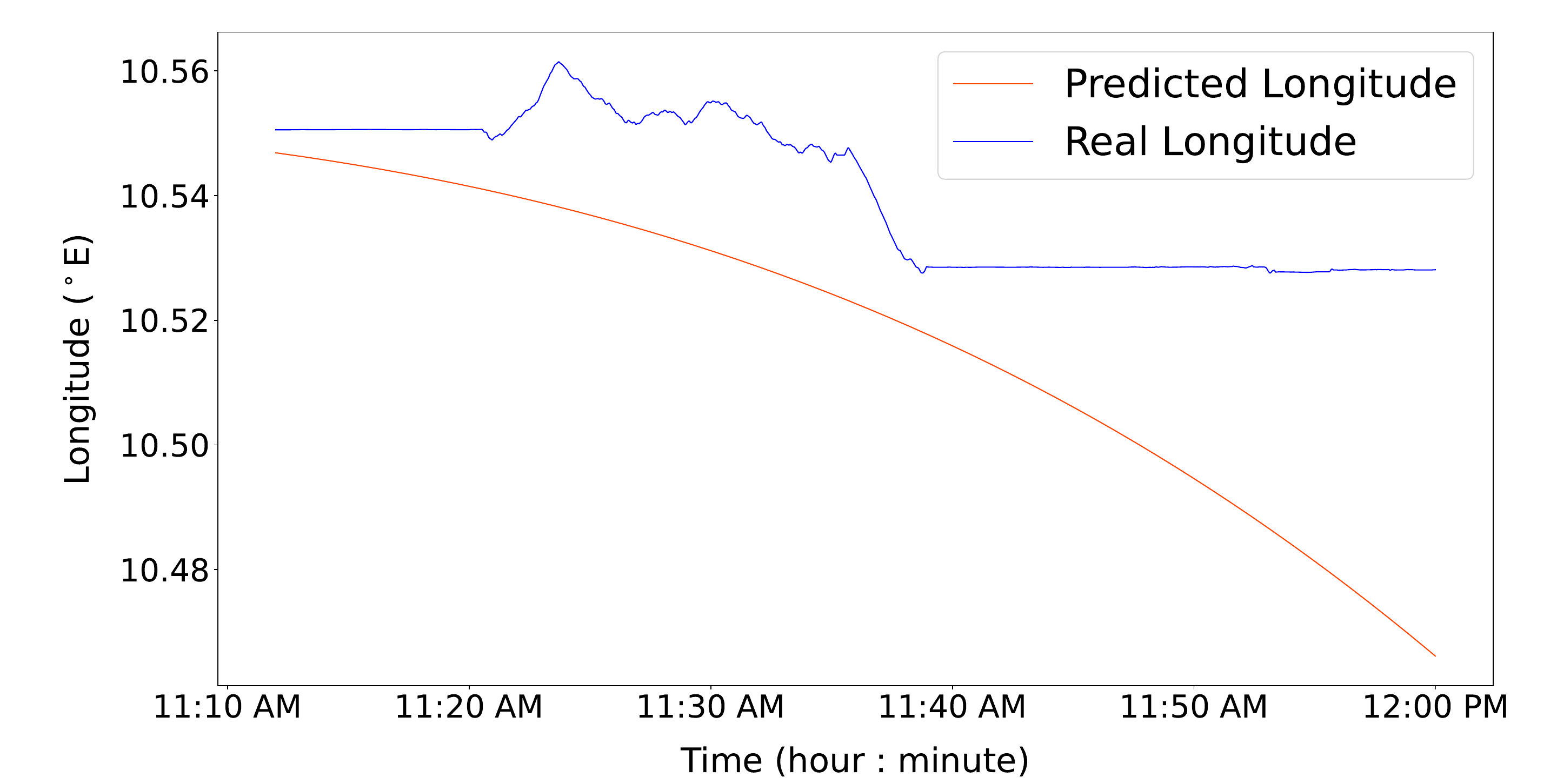}
         \caption{Longitude prediction during the first test data set}
         \label{N-Long1}
     \end{subfigure}
%###########################
     \centering
     \begin{subfigure}[p]{0.495\textwidth}
         \centering
         \includegraphics[width=\textwidth]{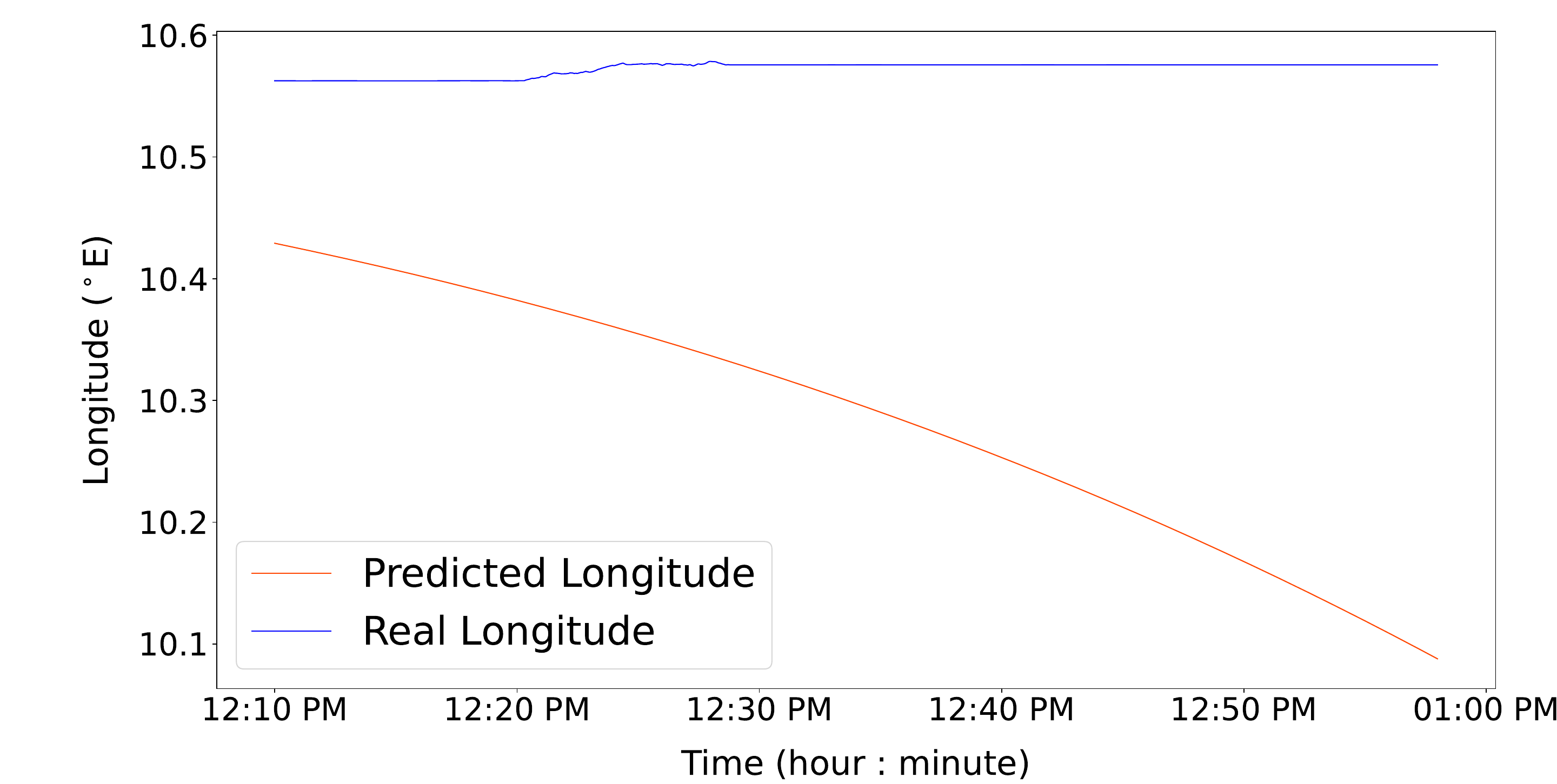}
         \caption{Longitude prediction during the second test data set}
         \label{N-Long2}
     \end{subfigure}
     \hfill
     \begin{subfigure}[p]{0.495\textwidth}
         \centering
         \includegraphics[width=\textwidth]{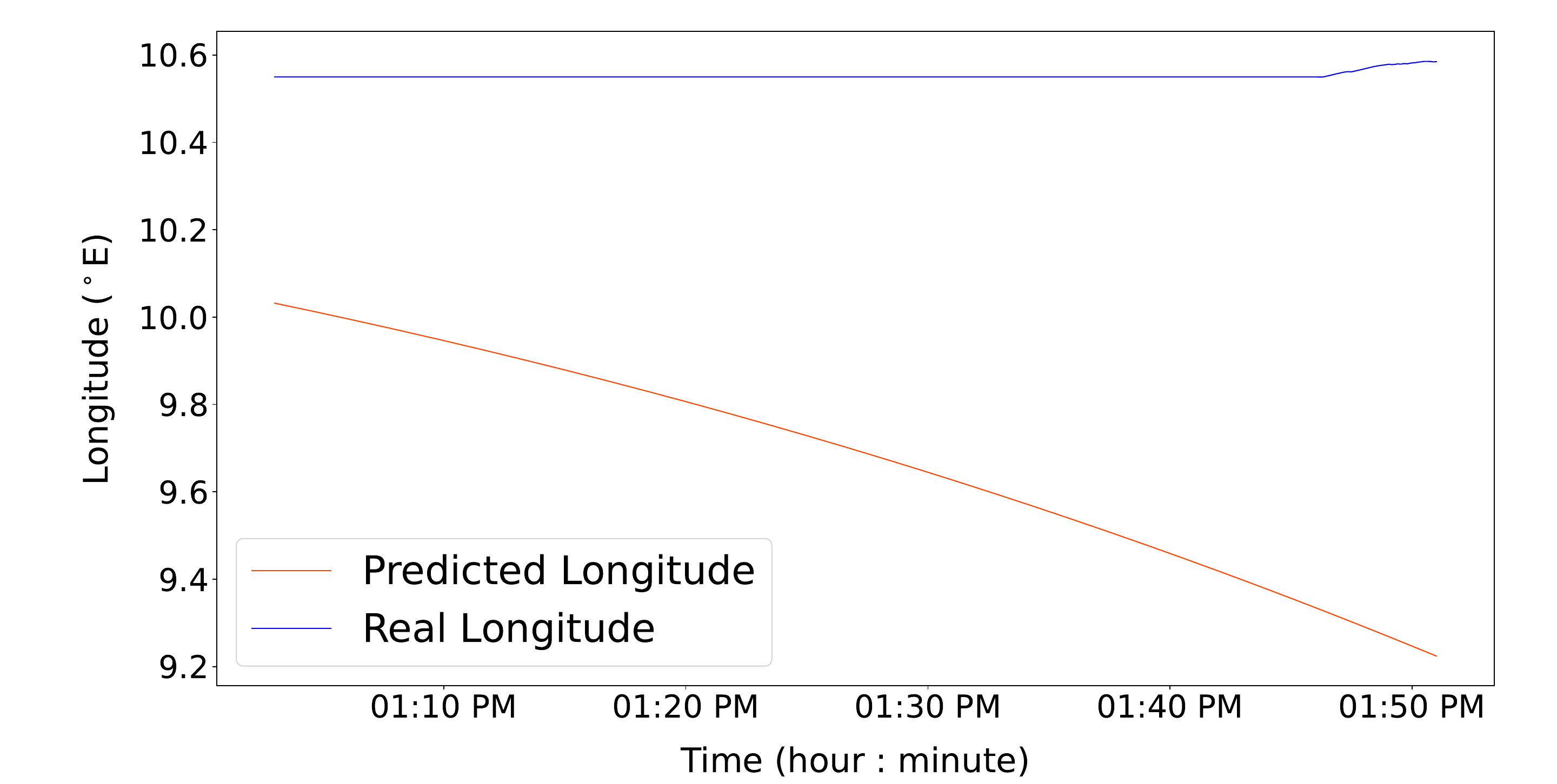}
         \caption{Longitude prediction during the third test data set}
         \label{N-Long3}
     \end{subfigure}
%###########################
     \centering
     \begin{subfigure}[p]{0.495\textwidth}
         \centering
         \includegraphics[width=\textwidth]{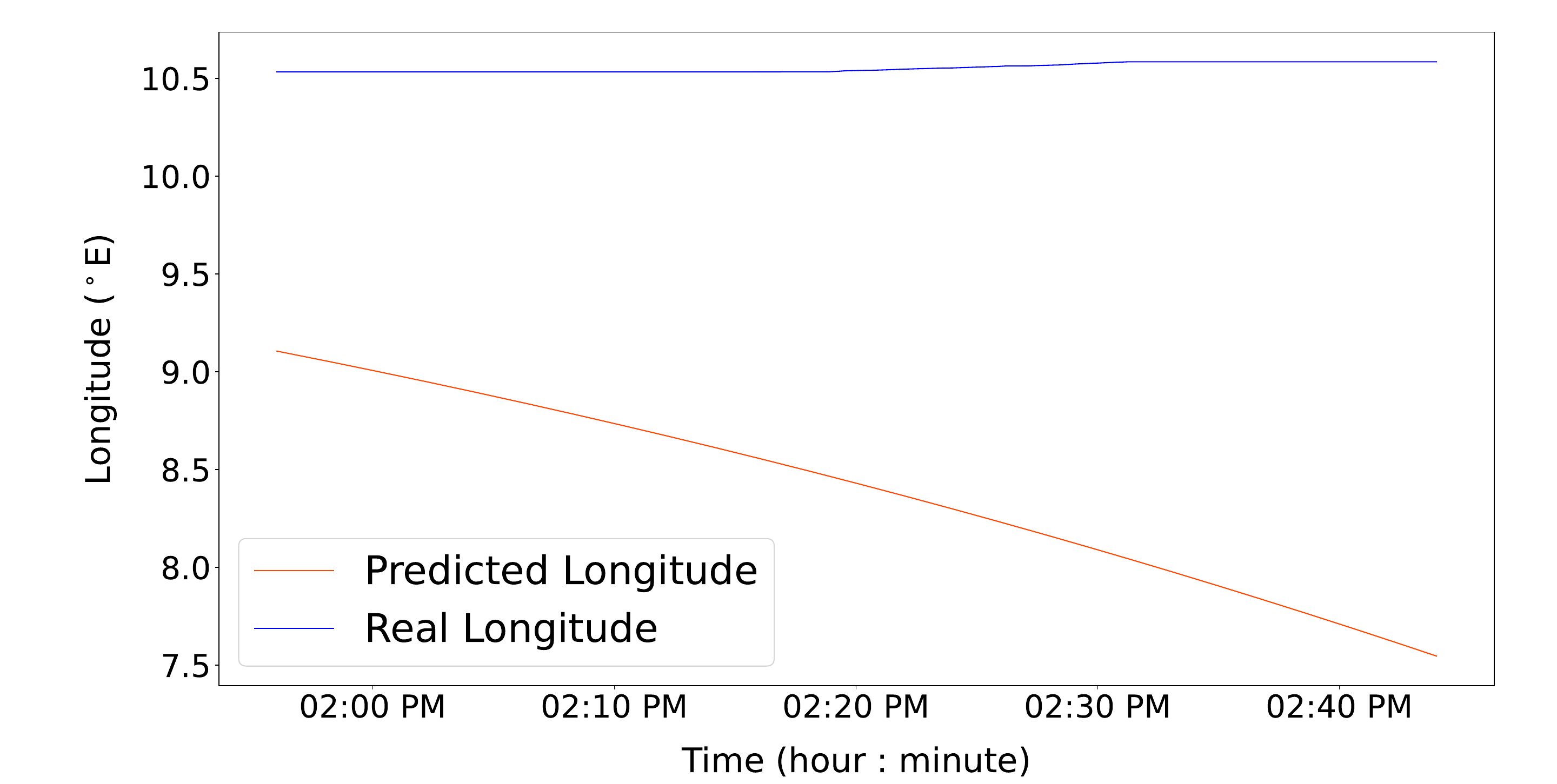}
         \caption{Longitude prediction during the forth test data set}
         \label{N-Long4}
     \end{subfigure}
     \hfill
     \begin{subfigure}[p]{0.495\textwidth}
         \centering
         \includegraphics[width=\textwidth]{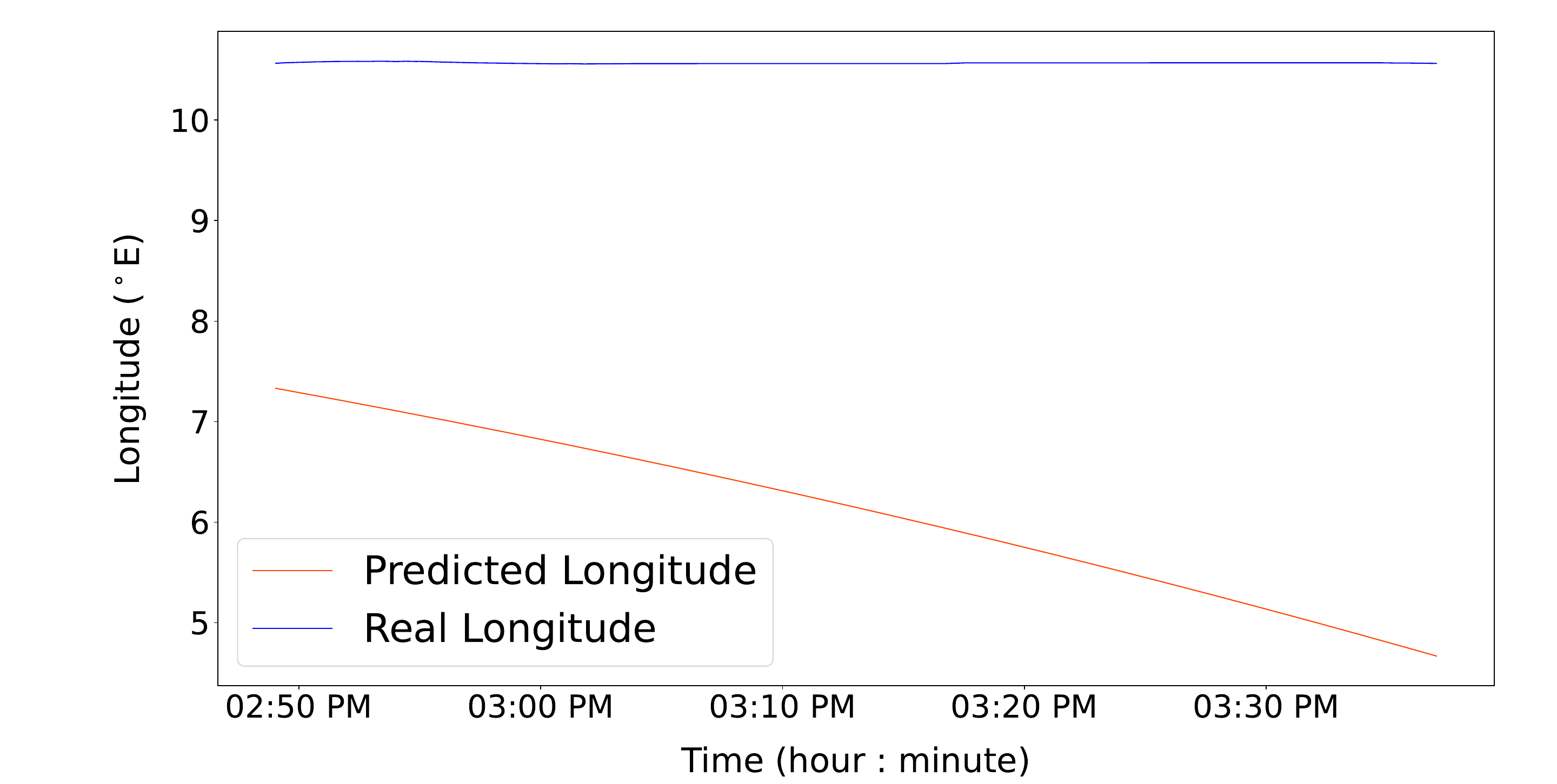}
         \caption{Longitude prediction during the fifth test data set}
         \label{N-Long5}
     \end{subfigure}
     \caption{Longitude predicted by nonlinear regression model}
     \label{Nonlinear-Longitude}
\end{figure}

\newpage
As another benchmark model to the LSTM models implemented, a forth-order polynomial nonlinear regression model has been implemented for latitude and longitude prediction. Four independent parameters in the model have been specified so that the error between the predicted output and the real output in training process gets minimized. Figs. \ref{N-Lat-T} and \ref{N-Long-T} show the performance of the nonlinear regression model during training in latitude and longitude prediction respectively. Fig. \ref{Nonlinear-Latitude} illustrate the performance of the model in latitude prediction for five different test data sets. They have revealed that a nonlinear regression model attempts to apply the fixed pattern it learned during training to different test data sets. This can lead to insufficient performance, as the model fails to adapt to the unique patterns present in the test data sets and instead tries to replicate what it learned during training. The performance of the model in longitude prediction is shown by Fig. \ref{Nonlinear-Longitude}. A divergence can be seen in all different test data set which indicate the insufficient performance of the model in bird movement prediction. Tables \ref{tab:Latitude} and \ref{tab:Longitude} indicate that the performance of both the linear and nonlinear regression models in predicting latitude and longitude is inaccurate for all test data sets. Despite the insufficient performance of both models, the linear regression model achieved a lower MAE in all test data sets for both latitude and longitude compared to the nonlinear regression model.
\floatplacement{figure}{!h}
%################################################Longitude############################################
\floatplacement{figure}{!p}
%###########################################################################################################################################################
%######################################################################################################
\newpage
\subsection{Collision Avoidance using Departure Delays}
\label{R-S-5}
\begin{figure}
     \centering
     \begin{subfigure}[h]{0.495\textwidth}
         \centering
         \includegraphics[width=\textwidth]{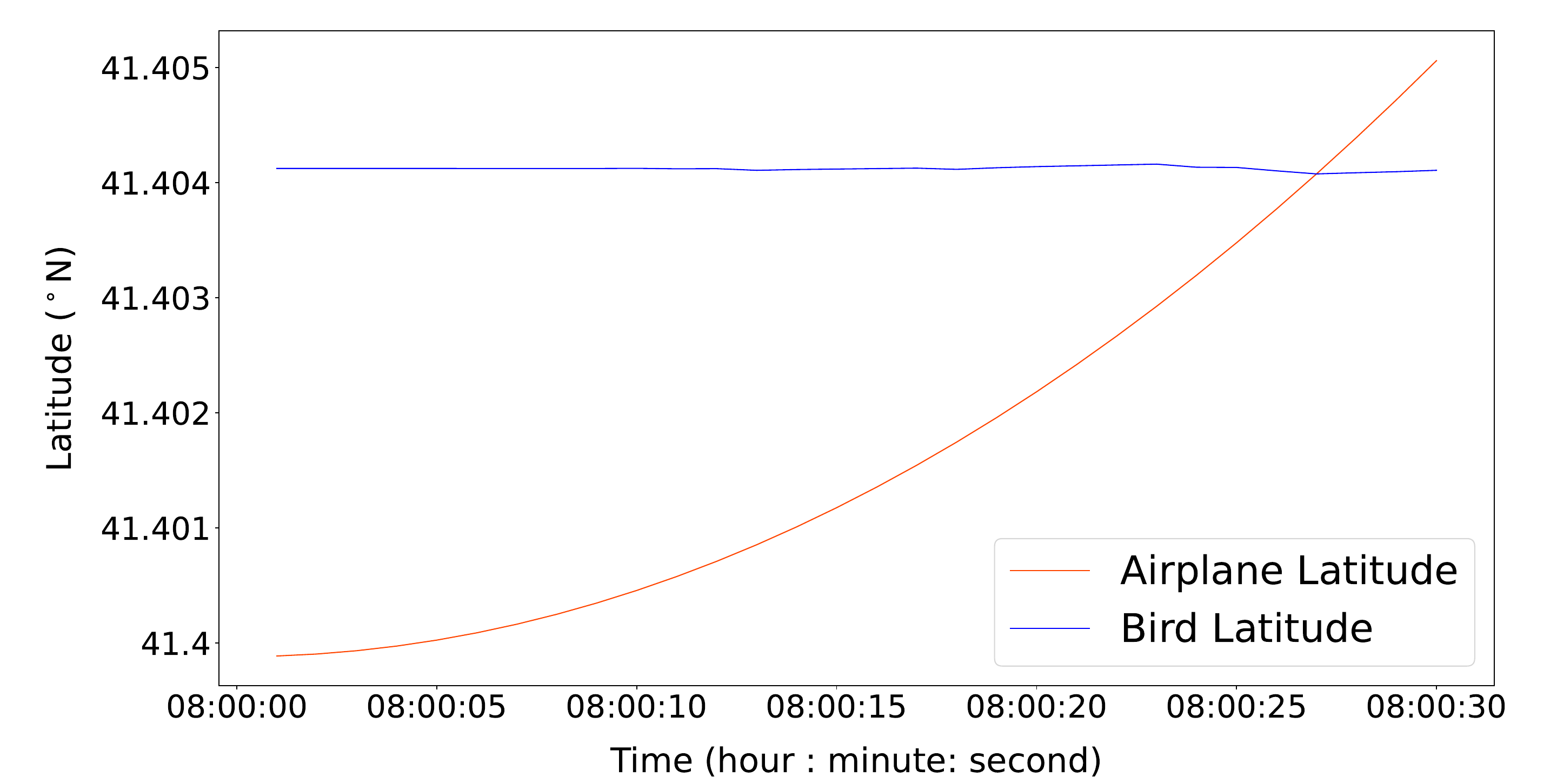}
         \caption{Bird strike in latitude}
         \label{Latitude-B}
     \end{subfigure}
     \hfill
     \begin{subfigure}[h]{0.495\textwidth}
         \centering
         \includegraphics[width=\textwidth]{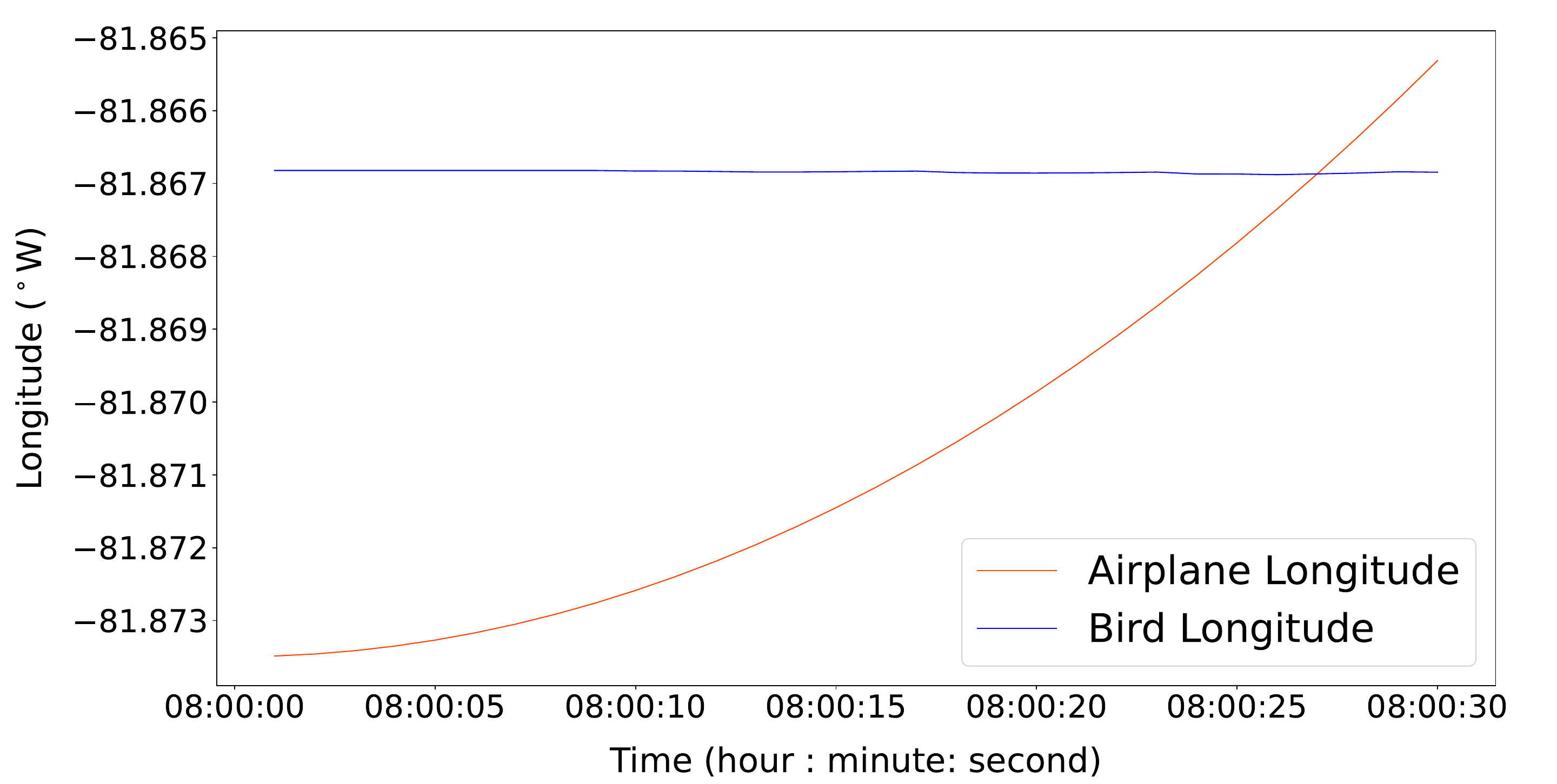}
         \caption{Bird strike in longitude}
         \label{Longitude-B}
     \end{subfigure}
     \caption{Detection of potential bird strike during airplane take off}
%#########################################################################################################################################################################################################
     \centering
     \begin{subfigure}[h]{0.495\textwidth}
         \centering
         \includegraphics[width=\textwidth]{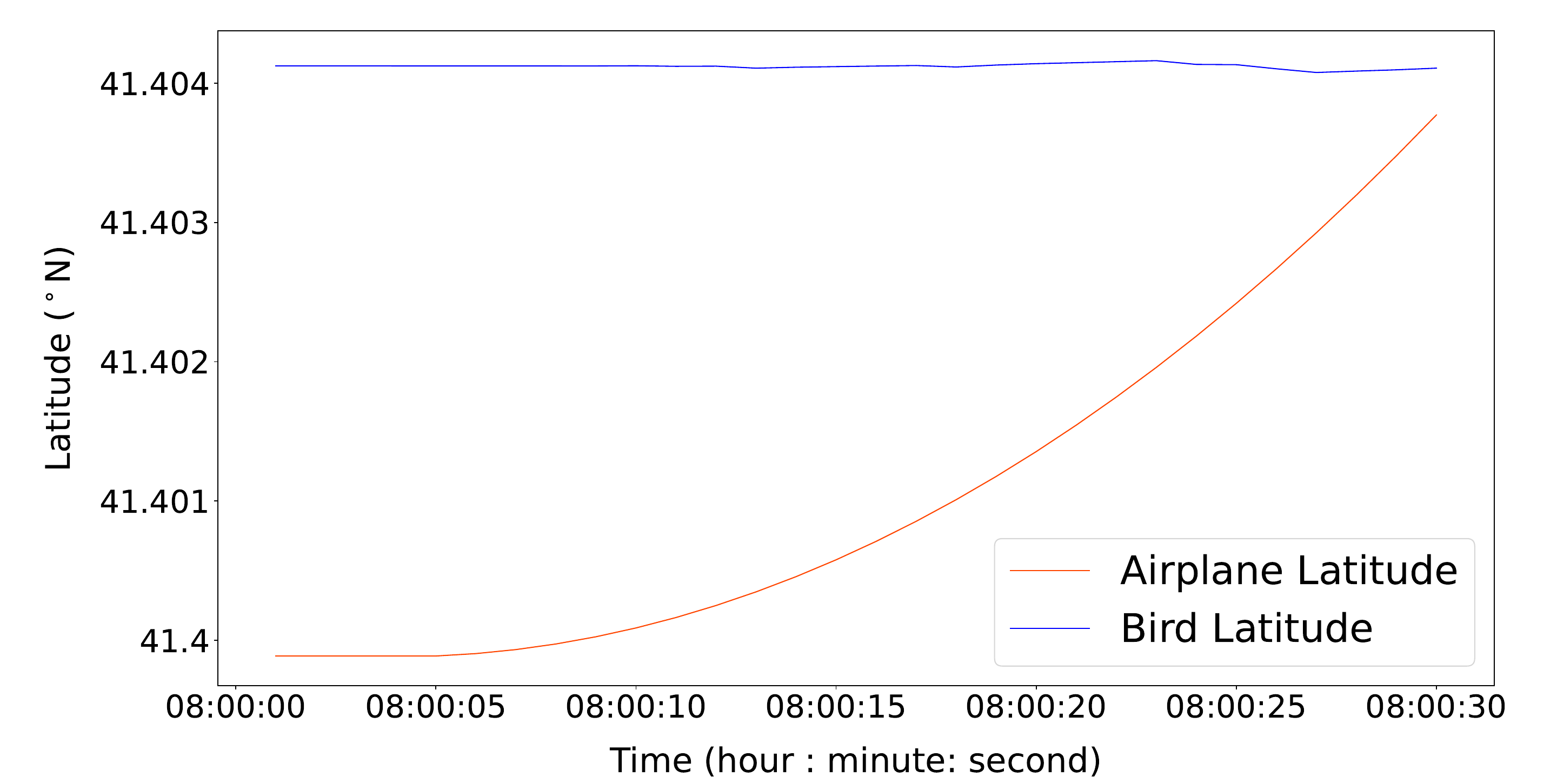}
         \caption{Collision free take off in latitude}
         \label{Latitude-A}
     \end{subfigure}
     \hfill
     \begin{subfigure}[h]{0.495\textwidth}
         \centering
         \includegraphics[width=\textwidth]{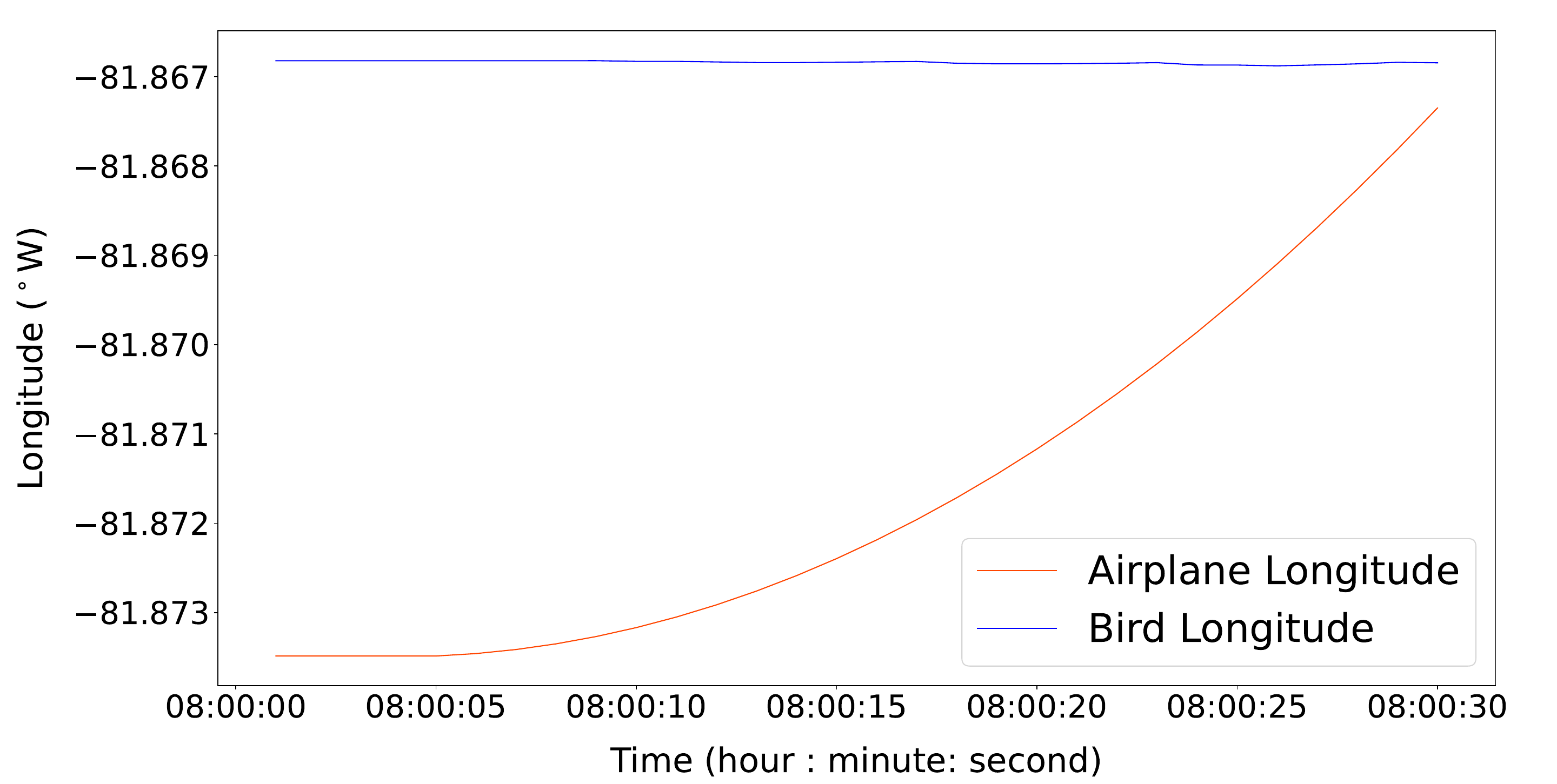}
         \caption{Collision free take off in longitude}
         \label{Longitude-A}
     \end{subfigure}
     \caption{Collision avoidance by delaying departure time}
\end{figure}

This paper presents a method for preventing bird strikes during takeoff by introducing a delay in the departure time. To simulate a real-world scenario, the path of a Boeing 737 on a runway at Cleveland Hopkins Airport was modeled. In addition the results from a vanilla LSTM model to predict bird movements and identify potential collisions in the takeoff path was used. The results revealed that a bird strike was detected simultaneously in both latitude and longitude, as shown in Figs. \ref{Latitude-B} and \ref{Longitude-B}. However, by delaying the departure time by four seconds, the potential bird strike during takeoff can be avoided, as demonstrated in Figs. \ref{Latitude-A} and \ref{Longitude-A}.
\floatplacement{figure}{!h}
\section{Conclusion}
\label{sec: Conclusion}
The accurate prediction of bird movement is critical for flight planning algorithms that aim to minimize the probability of bird strikes. To this end, we developed four different Long Short-Term Memory (LSTM) models, including vanilla, stacked, bidirectional, and encoder-decoder LSTM, to predict bird movement. The results demonstrate that LSTM models can predict bird movement tracks with high accuracy, achieving an MAE of less than 100 meters for all models, which far exceeds the accuracy levels of linear and nonlinear regression models. However, some oscillations were observed in the predictions, and future research can look into ways to reduce these oscillations to further improve the performance of LSTM models. By incorporating bird movement prediction into flight planning, this study shows that a bird strike during takeoff can be prevented by simply adjusting the departure time slightly.

An important area of future work is to take into account the uncertainty of the predictive models. By doing so, we can represent confidence intervals for the bird's movement locations, which would provide valuable insights into the reliability and accuracy of the predictions. Based on the insights we have gained, future work will involve addressing the weak assumption underlying most deconfliction algorithms. Currently, these algorithms assume that perfect and exact positions of other aircraft are known through Automatic Dependent Surveillance–Broadcast (ADS-B), which is not always applicable. For instance, noncooperative aircraft may intentionally refrain from turning on their ADS-B, making it difficult to track their movements. In such cases, predictive models can be used to estimate the aircraft's movements with confidence intervals, thereby providing a more robust and reliable means of deconfliction. 

\section*{Acknowledgment}
The authors acknowledge and express their gratitude for the technical guidance and support received from Dr. Isabel C. Metz  and Dr. Sophie F. Armanini during the performance of this research project. Additionally, the authors would like to thank the reviewers for reviewing this work.

\bibliography{main}

\begin{thebibliography}{45}
\newcommand{\enquote}[1]{``#1''}
\providecommand{\natexlab}[1]{#1}
\providecommand{\url}[1]{\texttt{#1}}
\providecommand{\urlprefix}{URL }
\expandafter\ifx\csname urlstyle\endcsname\relax
  \providecommand{\doi}[1]{\discretionary{}{}{}https://doi.org/#1}\else
  \providecommand{\doi}[1]{\discretionary{}{}{}\urlstyle{rm}\url{https://doi.org/#1}}\fi

\bibitem[{Skybrary(2019)}]{BS4}
Skybrary, \enquote{Bird Strike,} , 2019.
\newblock \urlprefix\url{https://www.skybrary.aero/articles/bird-strike},
  accessed 10 April 2023.

\bibitem[{Dolbeer(2013)}]{dolbeer2013history}
Dolbeer, R.~A., \enquote{The history of wildlife strikes and management at
  airports,} \emph{USDA National Wildlife Research Center-Staff Publications},
  2013.

\bibitem[{Cleary and Dolbeer(2005)}]{cleary2005wildlife}
Cleary, E.~C., and Dolbeer, R.~A., \enquote{Wildlife hazard management at
  airports: a manual for airport personnel,} \emph{USDA National Wildlife
  Research Center-Staff Publications}, 2005, p. 133.

\bibitem[{Simpleflying(2022)}]{BS2-2}
Simpleflying, \enquote{What Happened To The Airbus A320 That Landed On The
  Hudson?} , 2022.
\newblock
  \urlprefix\url{https://simpleflying.com/miracle-on-the-hudson-aicraft-fate/},
  accessed 10 April 2023.

\bibitem[{News(2012)}]{dennisbird}
News, W.~A., \enquote{Sita Air Dornier 228 Crashes in Nepal, 19 killed,} ,
  2012.
\newblock
  \urlprefix\url{https://worldairlinenews.com/2012/09/29/sita-air-dornier-228-crashes-in-nepal-19-killed/},
  accessed 10 April 2023.

\bibitem[{Herald(2019)}]{BS3}
Herald, A., \enquote{Accident: Ural A321 at Moscow on Aug 15th 2019, Bird
  Strike into Both Engines Forces Landing in Cornfield,} , 2019.
\newblock \urlprefix\url{http://avherald.com/h?article=4cb94927&opt=0},
  accessed 10 April 2023.

\bibitem[{Allan(2000)}]{costs}
Allan, J.~R., \enquote{The costs of bird strikes and bird strike prevention,}
  \emph{Human conflicts with wildlife: economic considerations}, 2000, p.~18.

\bibitem[{Dolbeer et~al.(2021)Dolbeer, Begier, Miller, Weller, Anderson
  et~al.}]{dolbeer2021wildlife}
Dolbeer, R.~A., Begier, M.~J., Miller, P.~R., Weller, J.~R., Anderson, A.~L.,
  et~al., \enquote{Wildlife Strikes to Civil Aircraft in the United States,
  1990--2021,} Tech. rep., United States. Department of Transportation. Federal
  Aviation Administration, 2021.

\bibitem[{Systems(2021)}]{Robin}
Systems, R.~R., \enquote{Three Reasons Why Bird Strike on Aircraft Are on the
  Rise,} , 2021.
\newblock
  \urlprefix\url{https://www.robinradar.com/press/blog/3-reasons-why-bird-strikes-on-aircraft-are-on-the-rise},
  accessed 24 April 2023.

\bibitem[{McKee et~al.(2016)McKee, Shaw, Dekker, and
  Patrick}]{mckee2016approaches}
McKee, J., Shaw, P., Dekker, A., and Patrick, K., \enquote{Approaches to
  wildlife management in aviation,} \emph{Problematic wildlife}, Springer,
  2016, pp. 465--488.

\bibitem[{Dolbeer(2011)}]{dolbeer2011increasing}
Dolbeer, R.~A., \enquote{Increasing trend of damaging bird strikes with
  aircraft outside the airport boundary: implications for mitigation measures,}
  \emph{Human-Wildlife Interactions}, Vol.~5, No.~2, 2011, pp. 235--248.

\bibitem[{Avrenli and Dempsey(2014)}]{avrenli2014statistical}
Avrenli, K.~A., and Dempsey, B.~J., \enquote{Statistical analysis of
  aircraft--bird strikes resulting in engine failure,} \emph{Transportation
  Research Record}, Vol. 2449, No.~1, 2014, pp. 14--23.

\bibitem[{Ning and Chen(2014)}]{ning2014bird}
Ning, H., and Chen, W., \enquote{Bird strike risk evaluation at airports,}
  \emph{Aircraft Engineering and Aerospace Technology: An International
  Journal}, Vol.~86, No.~2, 2014, pp. 129--137.

\bibitem[{Metz et~al.(2020)Metz, Ellerbroek, M{\"u}hlhausen, K{\"u}gler, and
  Hoekstra}]{metz2020bird}
Metz, I.~C., Ellerbroek, J., M{\"u}hlhausen, T., K{\"u}gler, D., and Hoekstra,
  J.~M., \enquote{The bird strike challenge,} \emph{Aerospace}, Vol.~7, No.~3,
  2020, p.~26.

\bibitem[{Metz(2021)}]{metz2021air}
Metz, I.~C., \enquote{Air Traffic Control Advisory System for the Prevention of
  Bird Strikes,} 2021.

\bibitem[{Metz et~al.(2021)Metz, Ellerbroek, M{\"u}hlhausen, K{\"u}gler, and
  Hoekstra}]{metz2021analysis}
Metz, I.~C., Ellerbroek, J., M{\"u}hlhausen, T., K{\"u}gler, D., and Hoekstra,
  J.~M., \enquote{Analysis of risk-based operational bird strike prevention,}
  \emph{Aerospace}, Vol.~8, No.~2, 2021, p.~32.

\bibitem[{Li et~al.(2020)Li, Lu, Zhang, and Chen}]{Moving-objects}
Li, M., Lu, F., Zhang, H., and Chen, J., \enquote{Predicting future locations
  of moving objects with deep fuzzy-LSTM networks,} \emph{Transportmetrica A:
  Transport Science}, Vol.~16, No.~1, 2020, pp. 119--136.

\bibitem[{Sundermeyer et~al.(2012)Sundermeyer, Schl{\"u}ter, and
  Ney}]{Language-modeling}
Sundermeyer, M., Schl{\"u}ter, R., and Ney, H., \enquote{LSTM neural networks
  for language modeling,} \emph{Thirteenth annual conference of the
  international speech communication association}, 2012.

\bibitem[{Nammous and Saeed(2019)}]{natural-l-processing}
Nammous, M.~K., and Saeed, K., \enquote{Natural language processing: speaker,
  language, and gender identification with LSTM,} \emph{Advanced Computing and
  Systems for Security: Volume Eight}, 2019, pp. 143--156.

\bibitem[{Zhuge et~al.(2017)Zhuge, Xu, and Zhang}]{stock-market}
Zhuge, Q., Xu, L., and Zhang, G., \enquote{LSTM Neural Network with Emotional
  Analysis for prediction of stock price.} \emph{Engineering letters}, Vol.~25,
  No.~2, 2017.

\bibitem[{Karevan and Suykens(2020)}]{weather-pattern}
Karevan, Z., and Suykens, J.~A., \enquote{Transductive LSTM for time-series
  prediction: An application to weather forecasting,} \emph{Neural Networks},
  Vol. 125, 2020, pp. 1--9.

\bibitem[{Liu et~al.(2019)Liu, Shao, Wong, Li, Su, and
  Kankanhalli}]{video-captioning}
Liu, A.-A., Shao, Z., Wong, Y., Li, J., Su, Y.-T., and Kankanhalli, M.,
  \enquote{LSTM-based multi-label video event detection,} \emph{Multimedia
  Tools and Applications}, Vol.~78, 2019, pp. 677--695.

\bibitem[{Wang et~al.(2021)Wang, Lu, Li, Liu, Xing, Lv, Cao, Li, Zhang, and
  Hashemi}]{autonomous}
Wang, H., Lu, B., Li, J., Liu, T., Xing, Y., Lv, C., Cao, D., Li, J., Zhang,
  J., and Hashemi, E., \enquote{Risk assessment and mitigation in local path
  planning for autonomous vehicles with LSTM based predictive model,}
  \emph{IEEE Transactions on Automation Science and Engineering}, Vol.~19,
  No.~4, 2021, pp. 2738--2749.

\bibitem[{Mathaiyan et~al.(2021)Mathaiyan, Vijayanandh, and
  Jung}]{mathaiyan2021determination}
Mathaiyan, V., Vijayanandh, R., and Jung, D.~W., \enquote{Determination of
  Strong Factor in Bird Strike Analysis using Taguchis method for Aircraft
  Manufacturing guide,} \emph{Journal of Physics: Conference Series}, Vol.
  1733, IOP Publishing, 2021, p. 012002.

\bibitem[{Shao et~al.(2020)Shao, Zhou, Zhu, Ma, and Shao}]{shao2020key}
Shao, Q., Zhou, Y., Zhu, P., Ma, Y., and Shao, M., \enquote{Key factors
  assessment on bird strike density distribution in airport habitats: Spatial
  heterogeneity and geographically weighted regression model,}
  \emph{Sustainability}, Vol.~12, No.~18, 2020, p. 7235.

\bibitem[{Smojver and Ivan{\v{c}}evi{\'c}(2011)}]{smojver2011bird}
Smojver, I., and Ivan{\v{c}}evi{\'c}, D., \enquote{Bird strike damage analysis
  in aircraft structures using Abaqus/Explicit and coupled Eulerian Lagrangian
  approach,} \emph{Composites science and technology}, Vol.~71, No.~4, 2011,
  pp. 489--498.

\bibitem[{Riccio et~al.(2016)Riccio, Cristiano, and Saputo}]{riccio2016brief}
Riccio, A., Cristiano, R., and Saputo, S., \enquote{A brief introduction to the
  bird strike numerical simulation,} \emph{Am. J. Eng. Applied Sci}, Vol.~9,
  2016, pp. 946--950.

\bibitem[{Company(2020)}]{bird-habitat}
Company, T. D.~B., \enquote{Airport Bird Control Drone,} , 2020.
\newblock
  \urlprefix\url{https://www.thedronebird.com/safe-humane-and-effective-bird-control/applications/airports/},
  accessed 10 April 2023.

\bibitem[{Bishop et~al.(2003)Bishop, McKay, Parrott, and Allan}]{B-H1}
Bishop, J., McKay, H., Parrott, D., and Allan, J., \enquote{Review of
  international research literature regarding the effectiveness of auditory
  bird scaring techniques and potential alternatives,} \emph{Food and rural
  affairs, London}, 2003, pp. 1--53.

\bibitem[{Seamans and Gosser(2016)}]{B-H2}
Seamans, T.~W., and Gosser, A.~L., \enquote{Bird dispersal techniques,} 2016.

\bibitem[{Paranjape et~al.(2018)Paranjape, Chung, Kim, and
  Shim}]{paranjape2018robotic}
Paranjape, A.~A., Chung, S.-J., Kim, K., and Shim, D.~H., \enquote{Robotic
  herding of a flock of birds using an unmanned aerial vehicle,} \emph{IEEE
  Transactions on Robotics}, Vol.~34, No.~4, 2018, pp. 901--915.

\bibitem[{van Gasteren et~al.(2019)van Gasteren, Krijgsveld, Klauke, Leshem,
  Metz, Skakuj, Sorbi, Schekler, and Shamoun-Baranes}]{van2019aeroecology}
van Gasteren, H., Krijgsveld, K.~L., Klauke, N., Leshem, Y., Metz, I.~C.,
  Skakuj, M., Sorbi, S., Schekler, I., and Shamoun-Baranes, J.,
  \enquote{Aeroecology meets aviation safety: Early warning systems in Europe
  and the Middle East prevent collisions between birds and aircraft,}
  \emph{Ecography}, Vol.~42, No.~5, 2019, pp. 899--911.

\bibitem[{Zhao et~al.(2021)Zhao, Erzberger, and Liu}]{zhao2021multiple}
Zhao, P., Erzberger, H., and Liu, Y., \enquote{Multiple-aircraft-conflict
  resolution under uncertainties,} \emph{Journal of Guidance, Control, and
  Dynamics}, Vol.~44, No.~11, 2021, pp. 2031--2049.

\bibitem[{Sislak et~al.(2007)Sislak, Volf, Komenda, Samek, and
  Pechoucek}]{4227576}
Sislak, D., Volf, P., Komenda, A., Samek, J., and Pechoucek, M.,
  \enquote{Agent-Based Multi-Layer Collision Avoidance to Unmanned Aerial
  Vehicles,} \emph{2007 International Conference on Integration of Knowledge
  Intensive Multi-Agent Systems}, 2007, pp. 365--370.
\newblock \doi{10.1109/KIMAS.2007.369837}.

\bibitem[{Sislak et~al.(2006)Sislak, Rehak, Pechoucek, Pavlicek, and
  Uller}]{1633455}
Sislak, D., Rehak, M., Pechoucek, M., Pavlicek, D., and Uller, M.,
  \enquote{Negotiation-Based Approach to Unmanned Aerial Vehicles,} \emph{IEEE
  Workshop on Distributed Intelligent Systems: Collective Intelligence and Its
  Applications (DIS'06)}, 2006, pp. 279--284.
\newblock \doi{10.1109/DIS.2006.55}.

\bibitem[{Fasano et~al.(2008)Fasano, Accardo, Moccia, Carbone, Ciniglio,
  Corraro, and Luongo}]{fasano2008multi}
Fasano, G., Accardo, D., Moccia, A., Carbone, C., Ciniglio, U., Corraro, F.,
  and Luongo, S., \enquote{Multi-sensor-based fully autonomous non-cooperative
  collision avoidance system for unmanned air vehicles,} \emph{Journal of
  aerospace computing, information, and communication}, Vol.~5, No.~10, 2008,
  pp. 338--360.

\bibitem[{Administration(2022)}]{BS1}
Administration, F.~A., \enquote{Transport Airlines,} , 2022.
\newblock \urlprefix\url{https://www.faa.gov}, accessed 10 April 2023.

\bibitem[{MoveBank(2022)}]{BS6}
MoveBank, \enquote{Data,} , 2022.
\newblock \urlprefix\url{https://www.movebank.org/cms/movebank-main}, accessed
  10 April 2023.

\bibitem[{Mastery(2016)}]{BS7}
Mastery, M.~L., \enquote{Time Series Prediction with LSTM Recurrent Neural
  Networks in Python with Keras,} , 2016.
\newblock
  \urlprefix\url{https://machinelearningmastery.com/time-series-prediction-lstm-recurrent-neural-networks-python-keras/},
  accessed 10 April 2023.

\bibitem[{Vidhya(2021)}]{BS5}
Vidhya, A., \enquote{Introduction to Long Short Term Memory (LSTM),} , 2021.
\newblock
  \urlprefix\url{https://www.analyticsvidhya.com/blog/2021/03/introduction-to-long-short-term-memory-lstm/},
  accessed 10 April 2023.

\bibitem[{PluralSight(2020)}]{BS9}
PluralSight, \enquote{LSTM,} , 2020.
\newblock
  \urlprefix\url{https://www.pluralsight.com/guides/introduction-to-lstm-units-in-rnn},
  accessed 10 April 2023.

\bibitem[{Sahar and Han(2018)}]{sahar2018lstm}
Sahar, A., and Han, D., \enquote{An LSTM-based indoor positioning method using
  Wi-Fi signals,} \emph{Proceedings of the 2nd International Conference on
  Vision, Image and Signal Processing}, 2018, pp. 1--5.

\bibitem[{Tavakoli(2019)}]{tavakoli2019modeling}
Tavakoli, N., \enquote{Modeling genome data using bidirectional LSTM,}
  \emph{2019 IEEE 43rd Annual Computer Software and Applications Conference
  (COMPSAC)}, Vol.~2, IEEE, 2019, pp. 183--188.

\bibitem[{Jacoby et~al.(2021)Jacoby, Ostrometzky, and Messer}]{jacoby2021short}
Jacoby, D., Ostrometzky, J., and Messer, H., \enquote{Short-term prediction of
  the attenuation in a commercial microwave link using LSTM-based RNN,}
  \emph{2020 28th European Signal Processing Conference (EUSIPCO)}, IEEE, 2021,
  pp. 1628--1632.

\bibitem[{Colah(2015)}]{BS8}
Colah, \enquote{Understanding LSTM Networks,} , 2015.
\newblock
  \urlprefix\url{https://colah.github.io/posts/2015-08-Understanding-LSTMs/},
  accessed 10 April 2023.

\end{thebibliography}

\newpage
\section{Appendix}
\label{appendix}

In this section, we present the training and test results of stacked, bidirectional, and encoder-decoder LSTM models used to predict bird movements. We also compare the performance of these models with respect to the vanilla LSTM model, which was previously presented in Section \ref{sec:Results}.

\subsection{Stacked LSTM}

A stacked LSTM consists of more than one hidden layer to make the model deeper. Among the conducted experiments, an LSTM model with two hidden layers was found to have the lowest error for latitude prediction. The model had 16 neurons in the first hidden layer and 8 neurons in the second hidden layer. Its performance during training and testing is shown in Fig. \ref{Fig: Latitude-Vanilla}. For longitude prediction, an LSTM model with two hidden layers was used, consisting of 32 and 8 neurons in the first and second layer respectively. The performance of LSTM for longitude prediction in training and five test data sets are illustrated in Fig. \ref{Fig: Longitude-Vanilla}. The performance of the stacked LSTM is relatively similar to, or slightly worse than, the vanilla LSTM, indicating that increasing the number of hidden layers does not necessarily improve accuracy and heavily relies on the data. In the performance of the stacked LSTM, more oscillations can be seen during the periods when the bird's movement is in a straight line. 
% \floatplacement{figure}{!h}
\begin{figure}
     \centering
     \begin{subfigure}[!h]{0.495\textwidth}
         \centering
         \includegraphics[width=\textwidth]{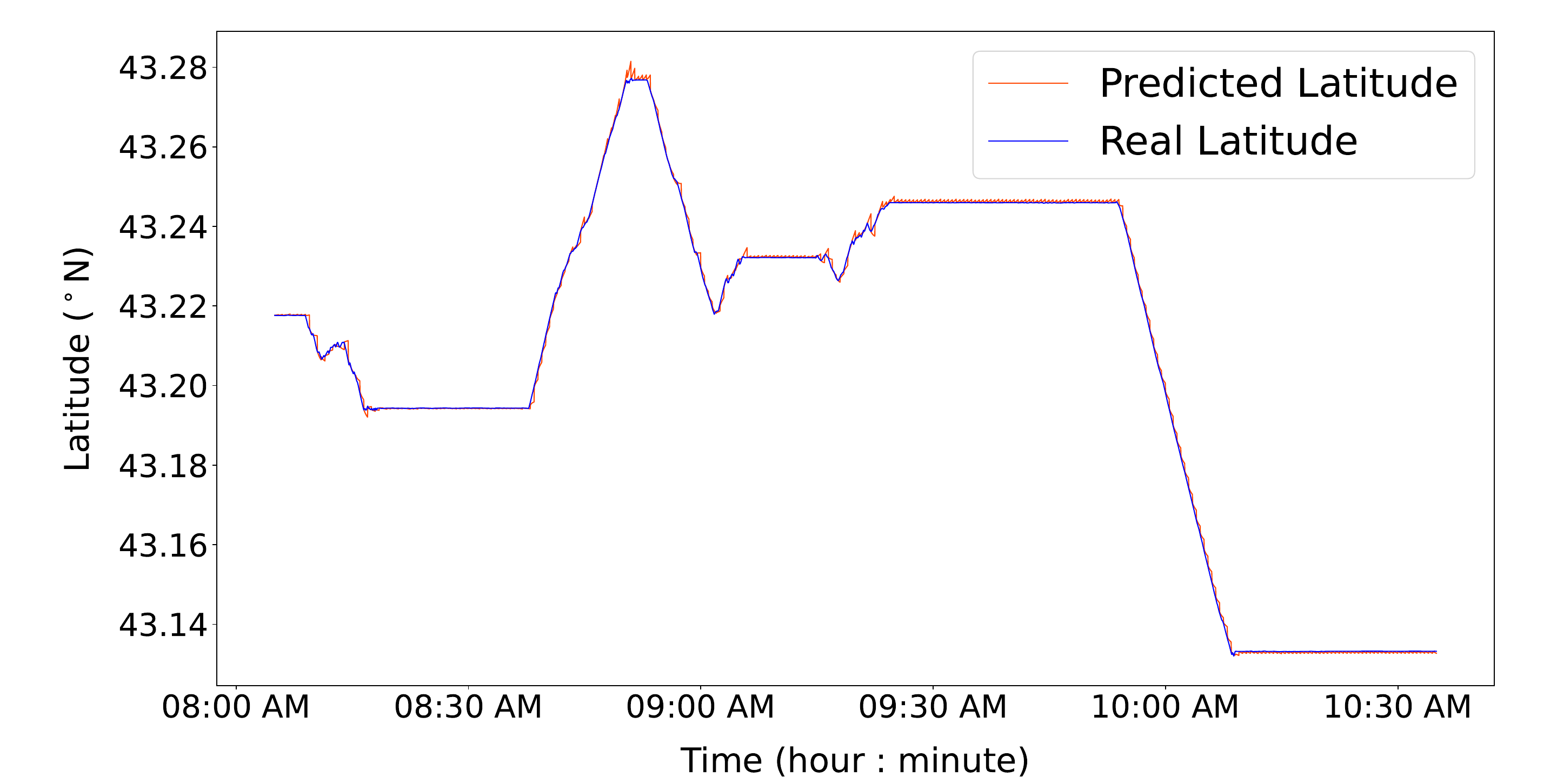}
         \caption{Latitude prediction during training}
         \label{S-Lat-T}
     \end{subfigure}
     \hfill
     \begin{subfigure}[h]{0.495\textwidth}
         \centering
         \includegraphics[width=\textwidth]{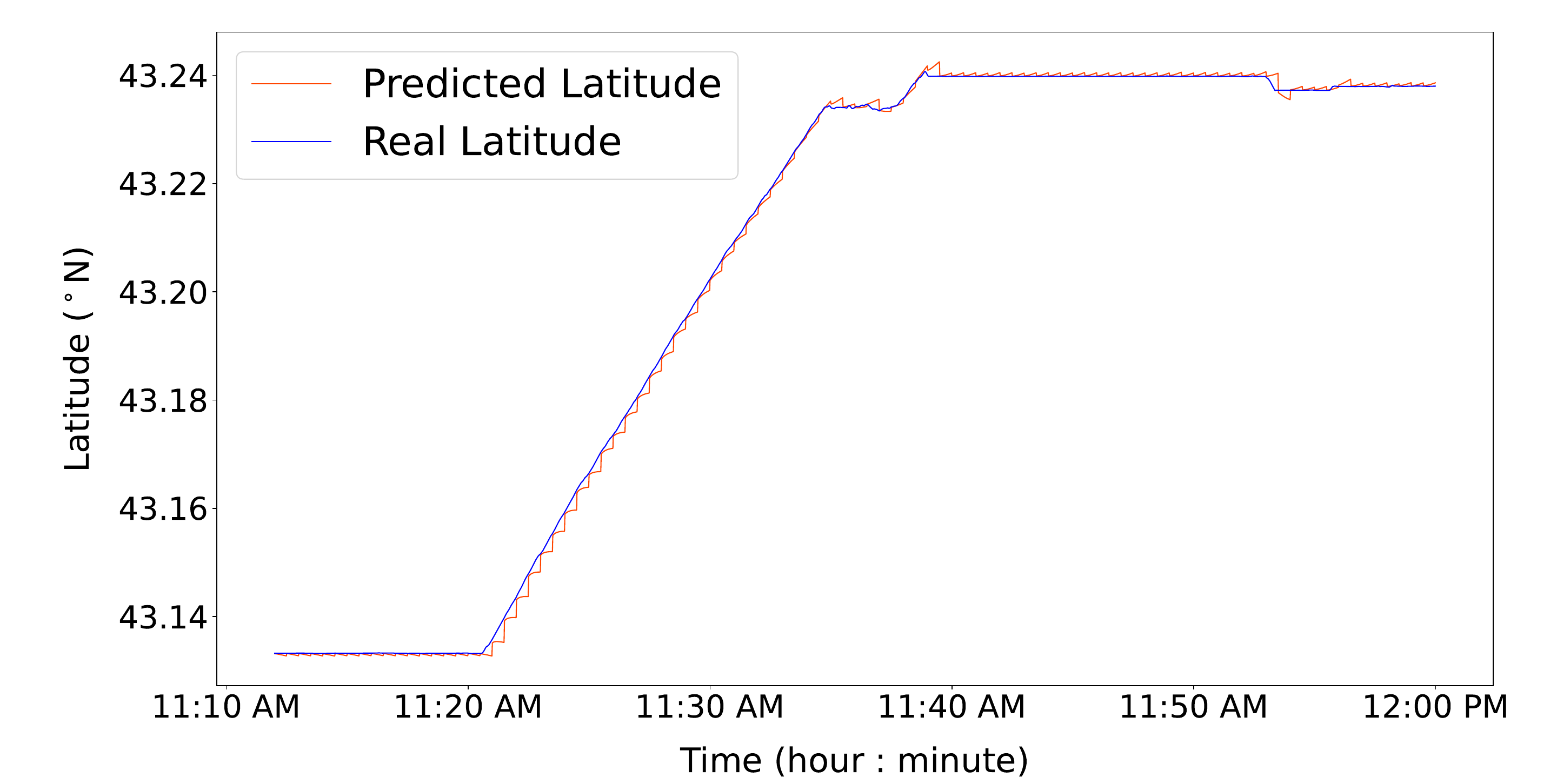}
         \caption{Latitude prediction during the first test data set}
         \label{S-Lat1}
     \end{subfigure}
%###########################
     \centering
     \begin{subfigure}[h]{0.495\textwidth}
         \centering
         \includegraphics[width=\textwidth]{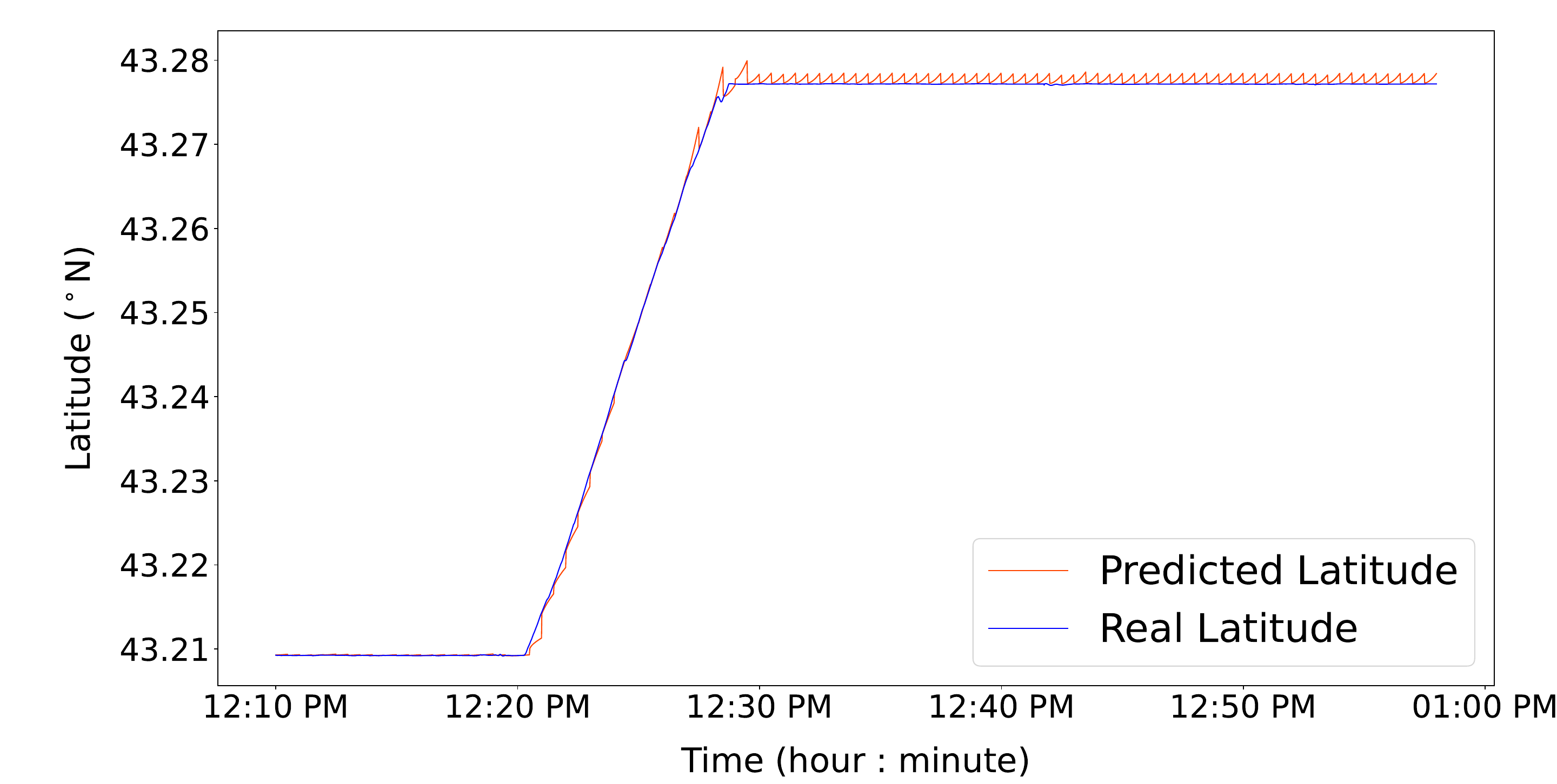}
         \caption{Latitude prediction during the second test data set}
         \label{S-Lat2}
     \end{subfigure}
     \hfill
     \begin{subfigure}[h]{0.495\textwidth}
         \centering
         \includegraphics[width=\textwidth]{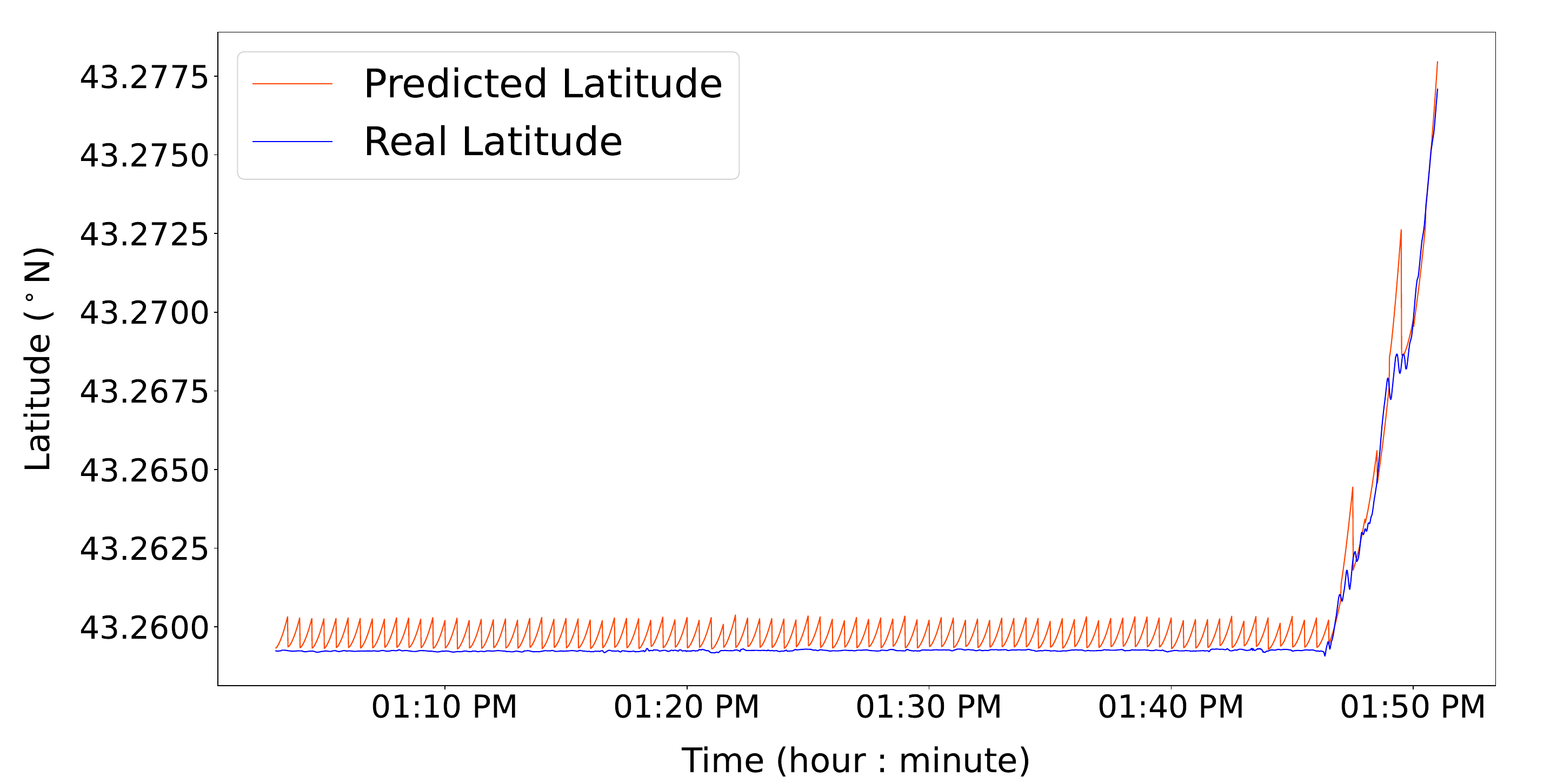}
         \caption{Latitude prediction during the third test data set}
         \label{S-Lat3}
     \end{subfigure}
%###########################
     \centering
     \begin{subfigure}[h]{0.495\textwidth}
         \centering
         \includegraphics[width=\textwidth]{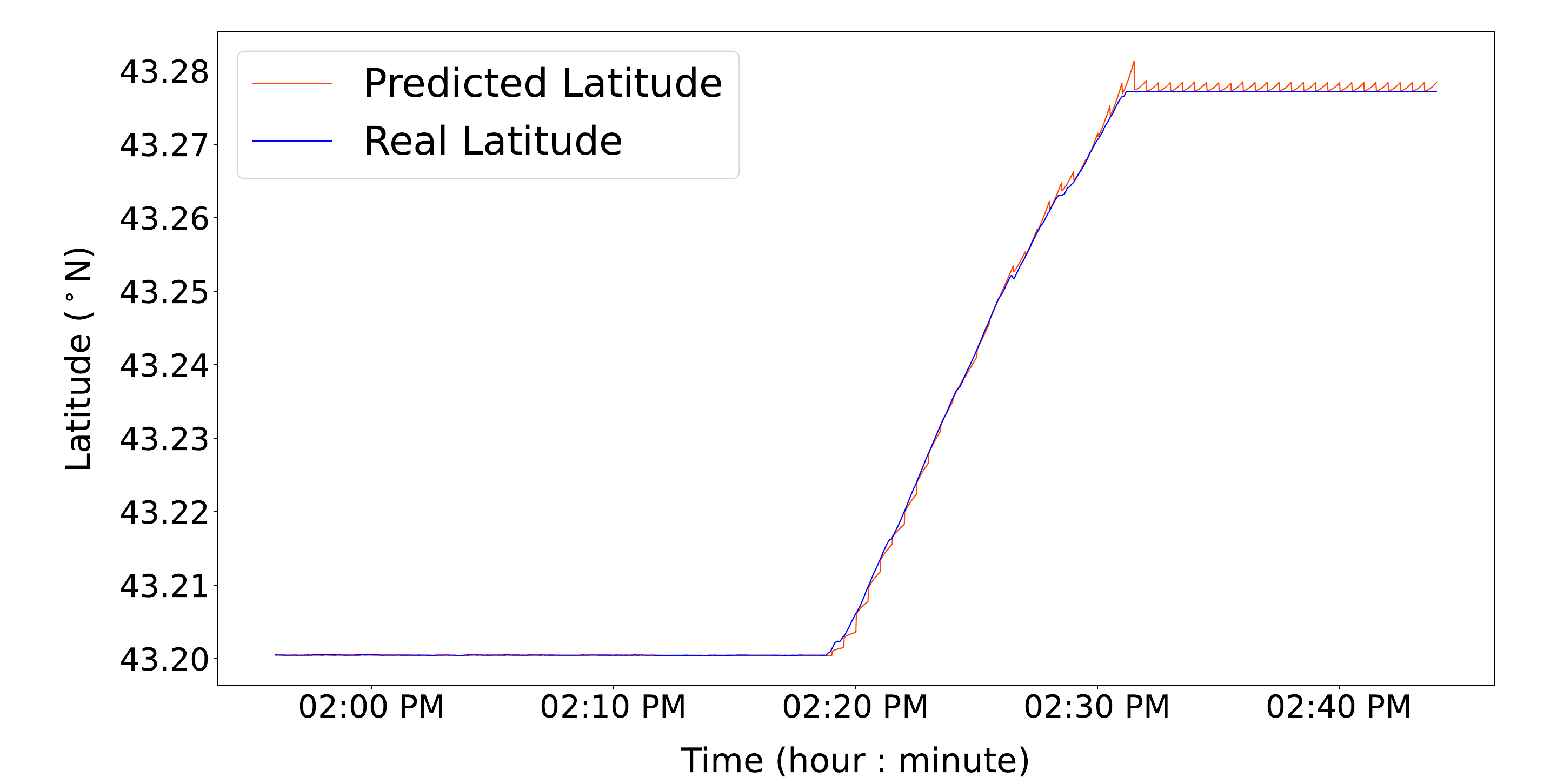}
         \caption{Latitude prediction during the forth test data set}
         \label{S-Lat4}
     \end{subfigure}
     \hfill
     \begin{subfigure}[h]{0.495\textwidth}
         \centering
         \includegraphics[width=\textwidth]{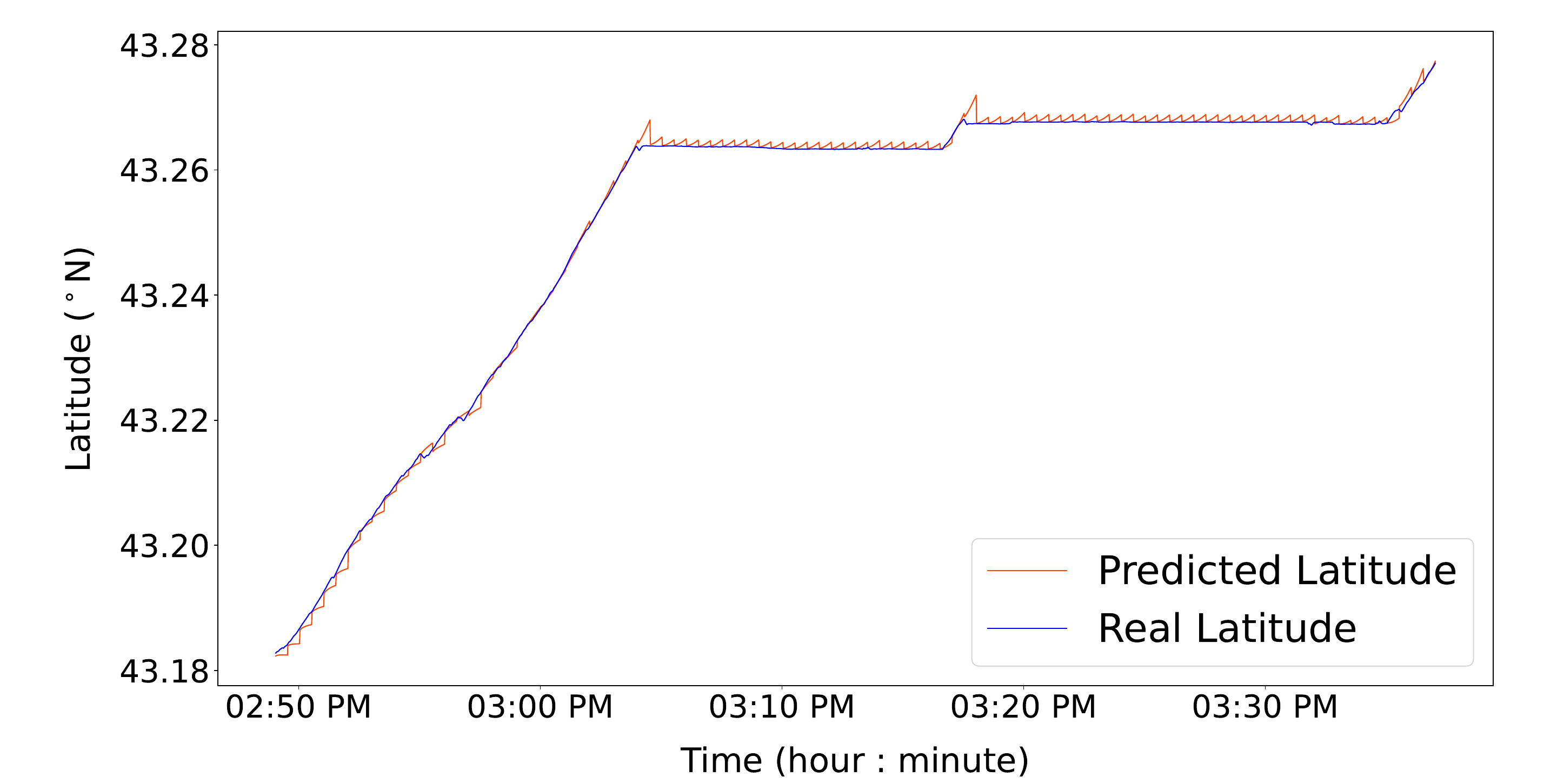}
         \caption{Latitude prediction during the fifth test data set}
         \label{S-Lat5}
     \end{subfigure}
     \caption{Latitude predicted by stacked LSTM}
     \label{stacked-Latitude}
\end{figure}
%################################################Longitude############################################
\floatplacement{figure}{!p}
\begin{figure}
     \centering
     \begin{subfigure}[p]{0.495\textwidth}
         \centering
         \includegraphics[width=\textwidth]{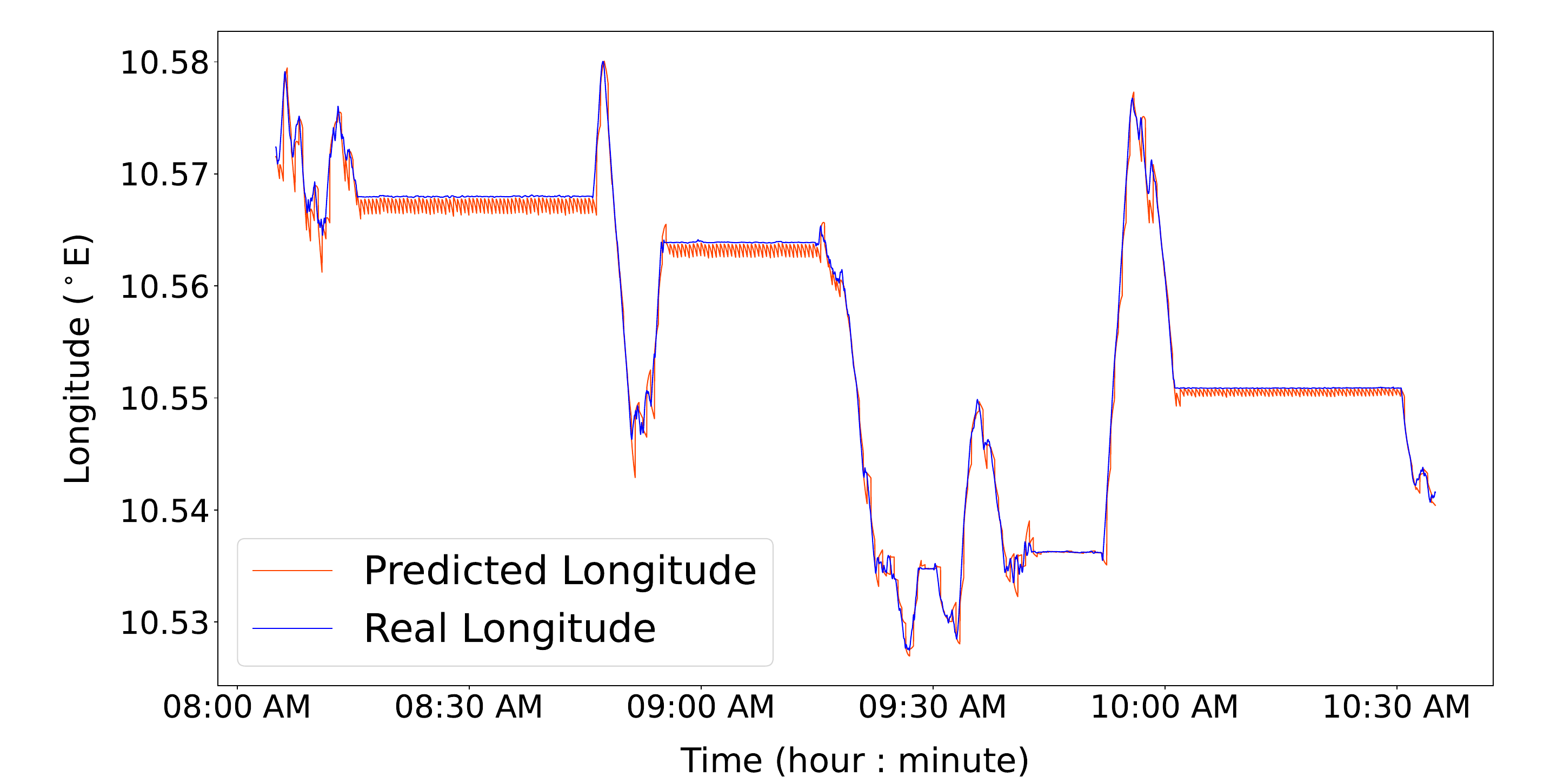}
         \caption{Longitude prediction during training}
         \label{S-Long-T}
     \end{subfigure}
     \hfill
     \begin{subfigure}[p]{0.495\textwidth}
         \centering
         \includegraphics[width=\textwidth]{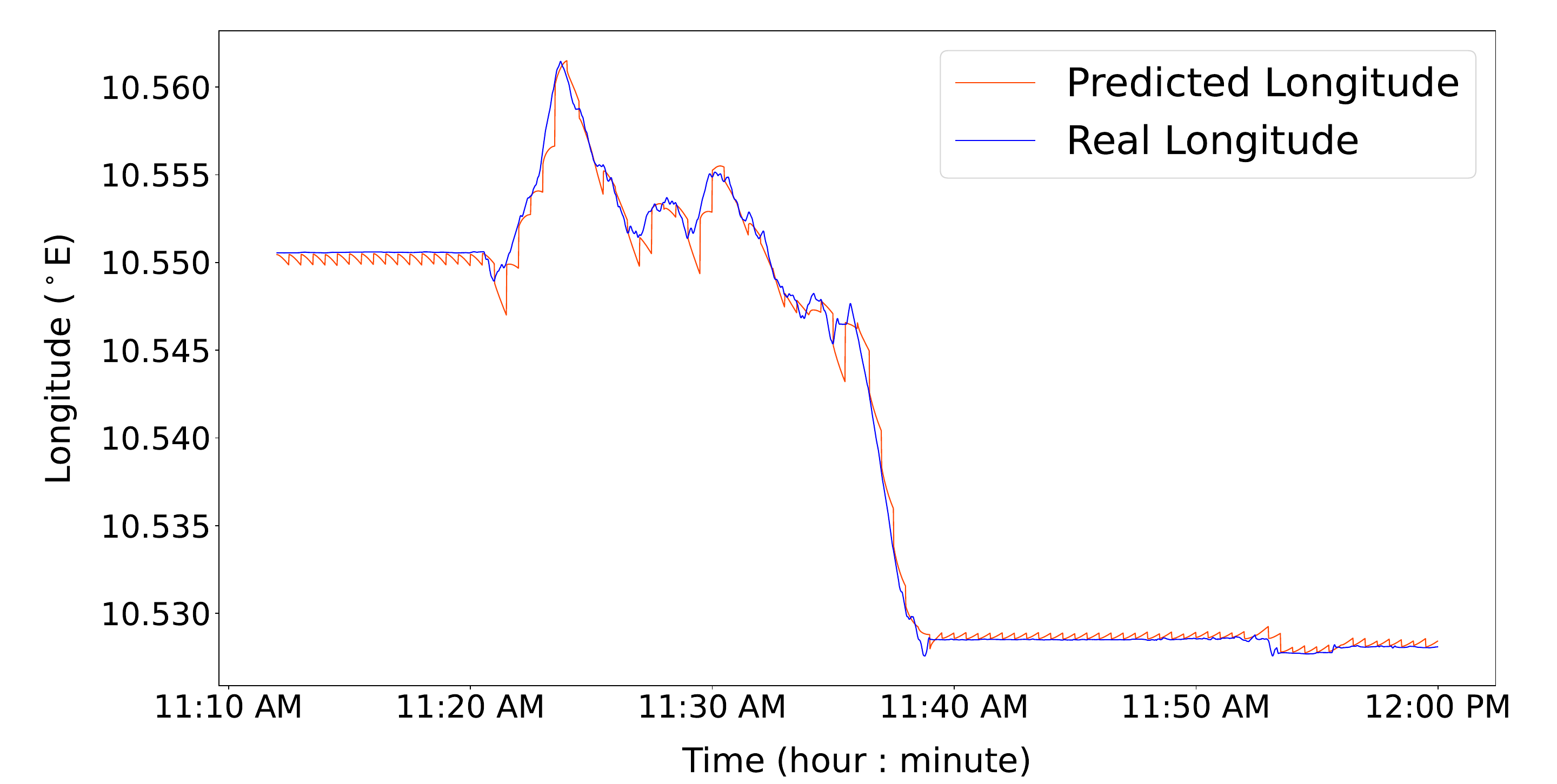}
         \caption{Longitude prediction during the first test data set}
         \label{S-Long1}
     \end{subfigure}
%###########################
     \centering
     \begin{subfigure}[p]{0.495\textwidth}
         \centering
         \includegraphics[width=\textwidth]{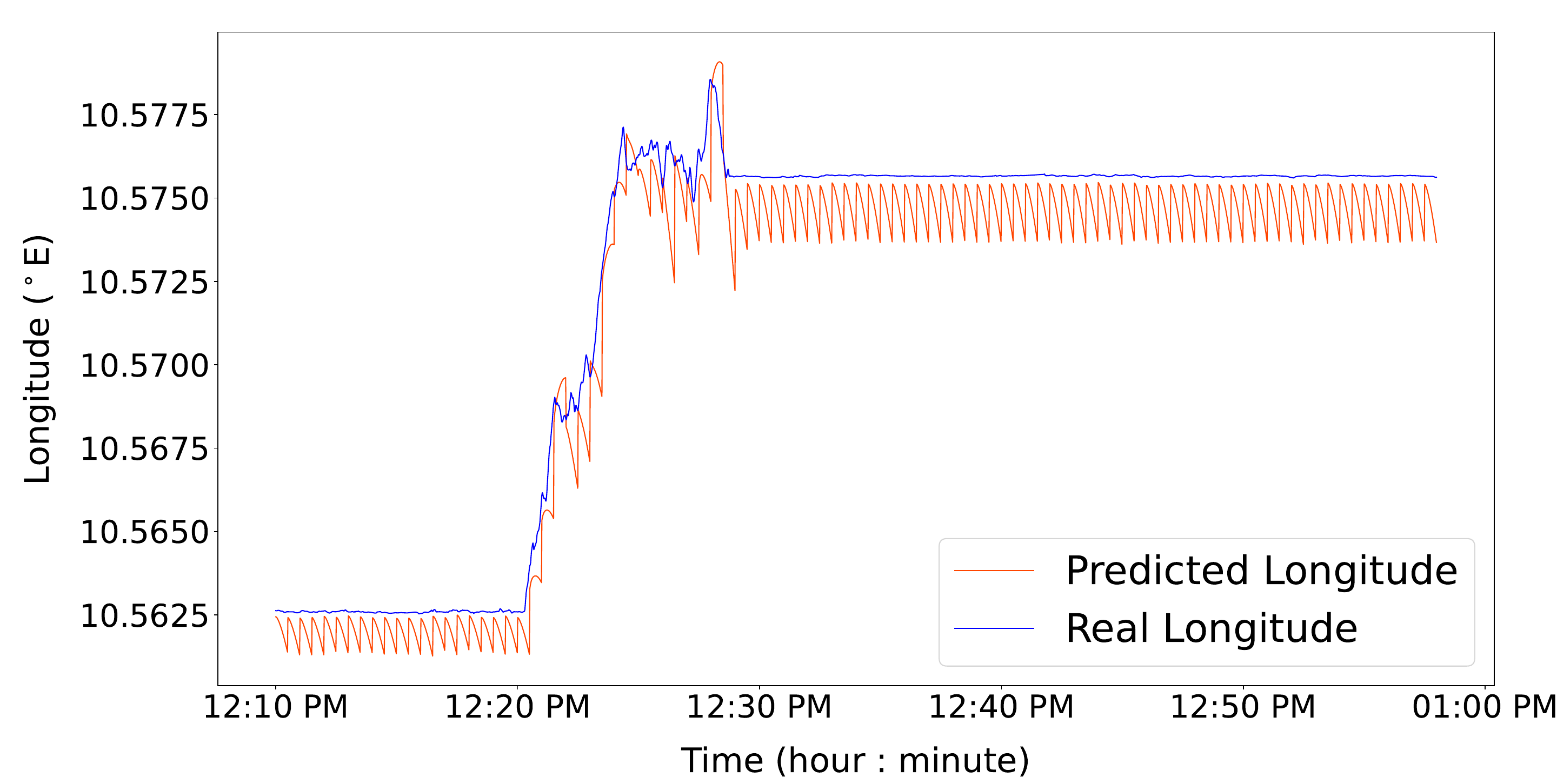}
         \caption{Longitude prediction during the second test data set}
         \label{S-Long2}
     \end{subfigure}
     \hfill
     \begin{subfigure}[p]{0.495\textwidth}
         \centering
         \includegraphics[width=\textwidth]{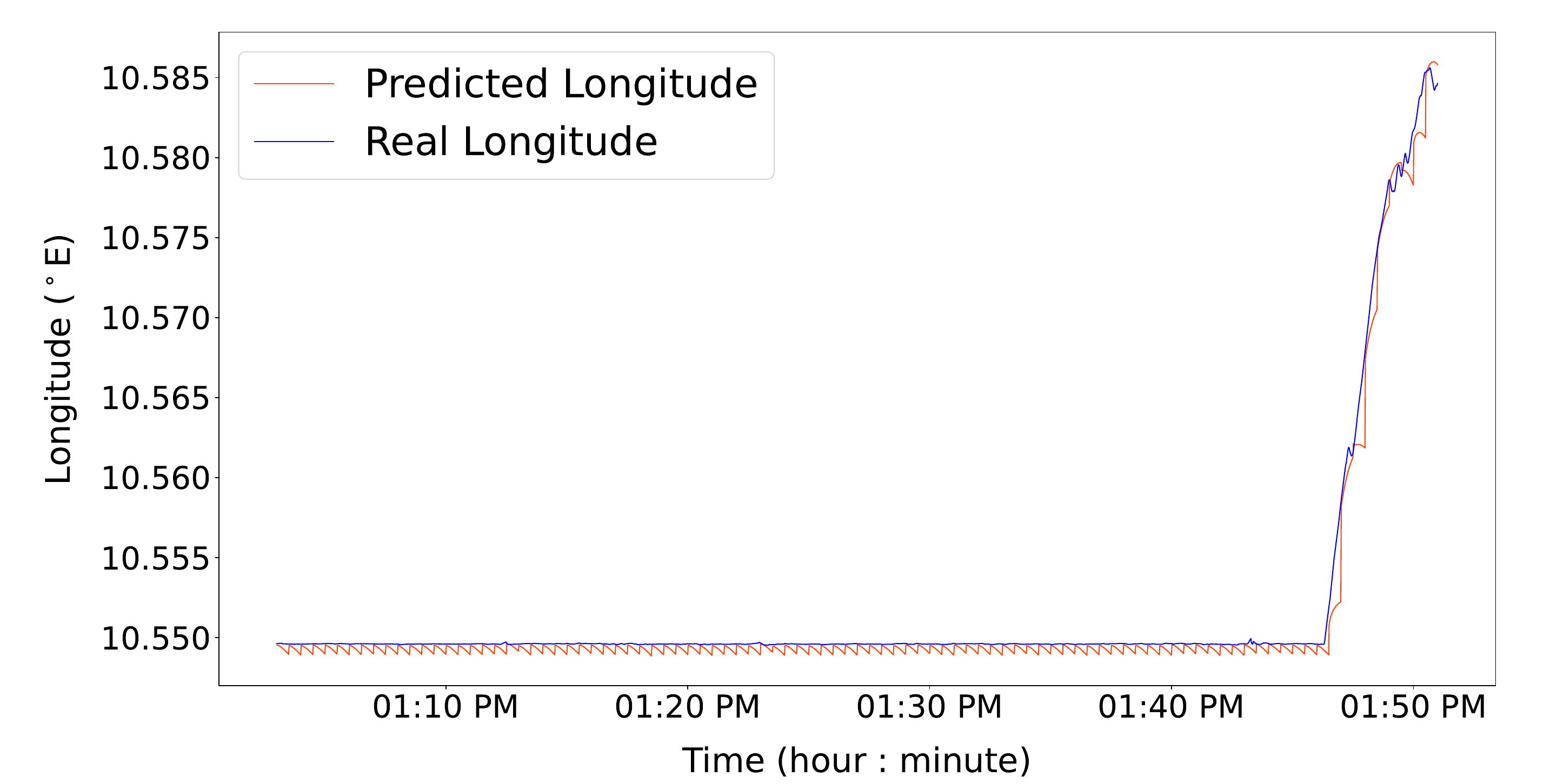}
         \caption{Longitude prediction during the third test data set}
         \label{S-Long3}
     \end{subfigure}
%###########################
     \centering
     \begin{subfigure}[p]{0.495\textwidth}
         \centering
         \includegraphics[width=\textwidth]{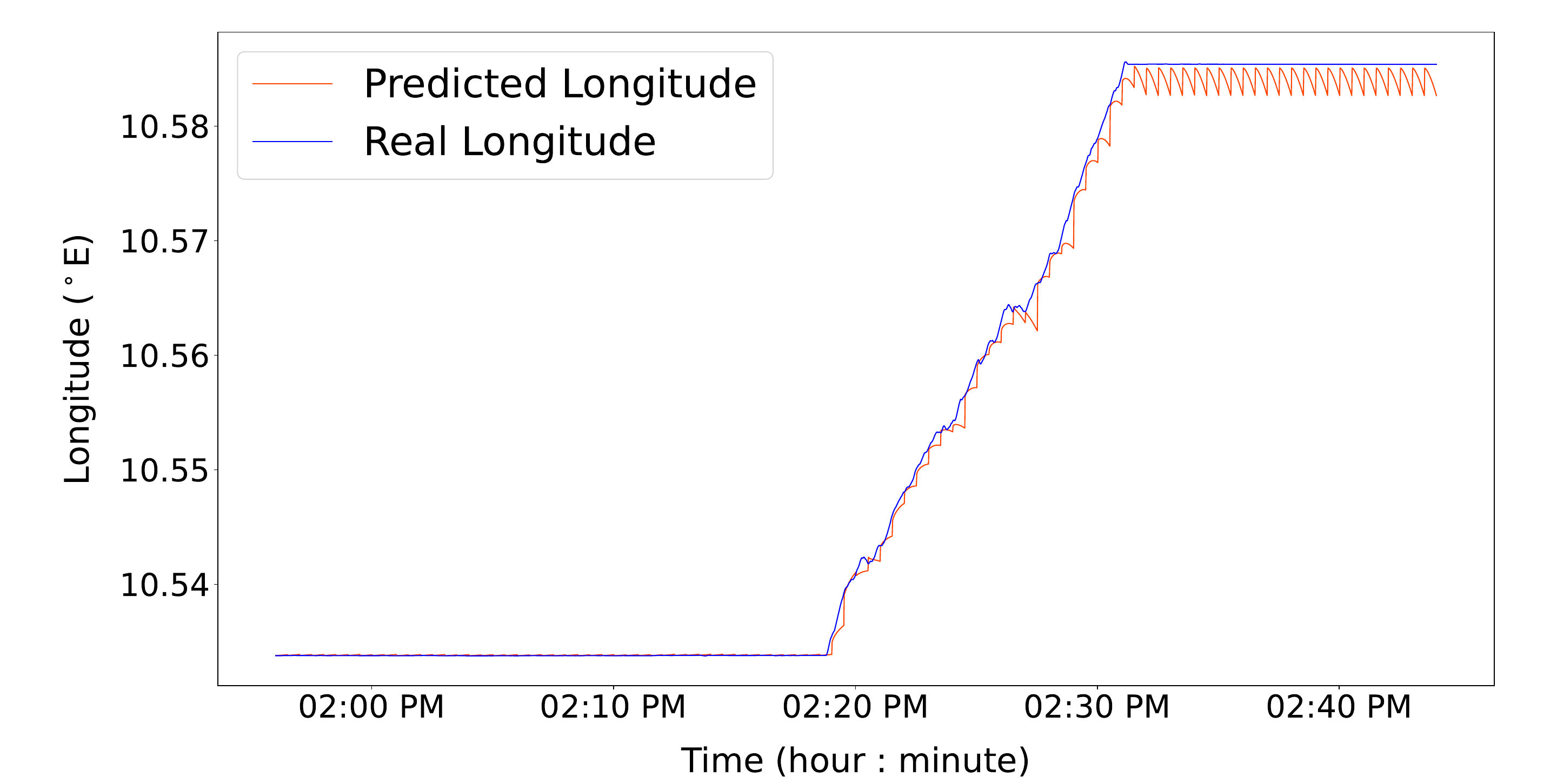}
         \caption{Longitude prediction during the forth test data set}
         \label{S-Long4}
     \end{subfigure}
     \hfill
     \begin{subfigure}[p]{0.495\textwidth}
         \centering
         \includegraphics[width=\textwidth]{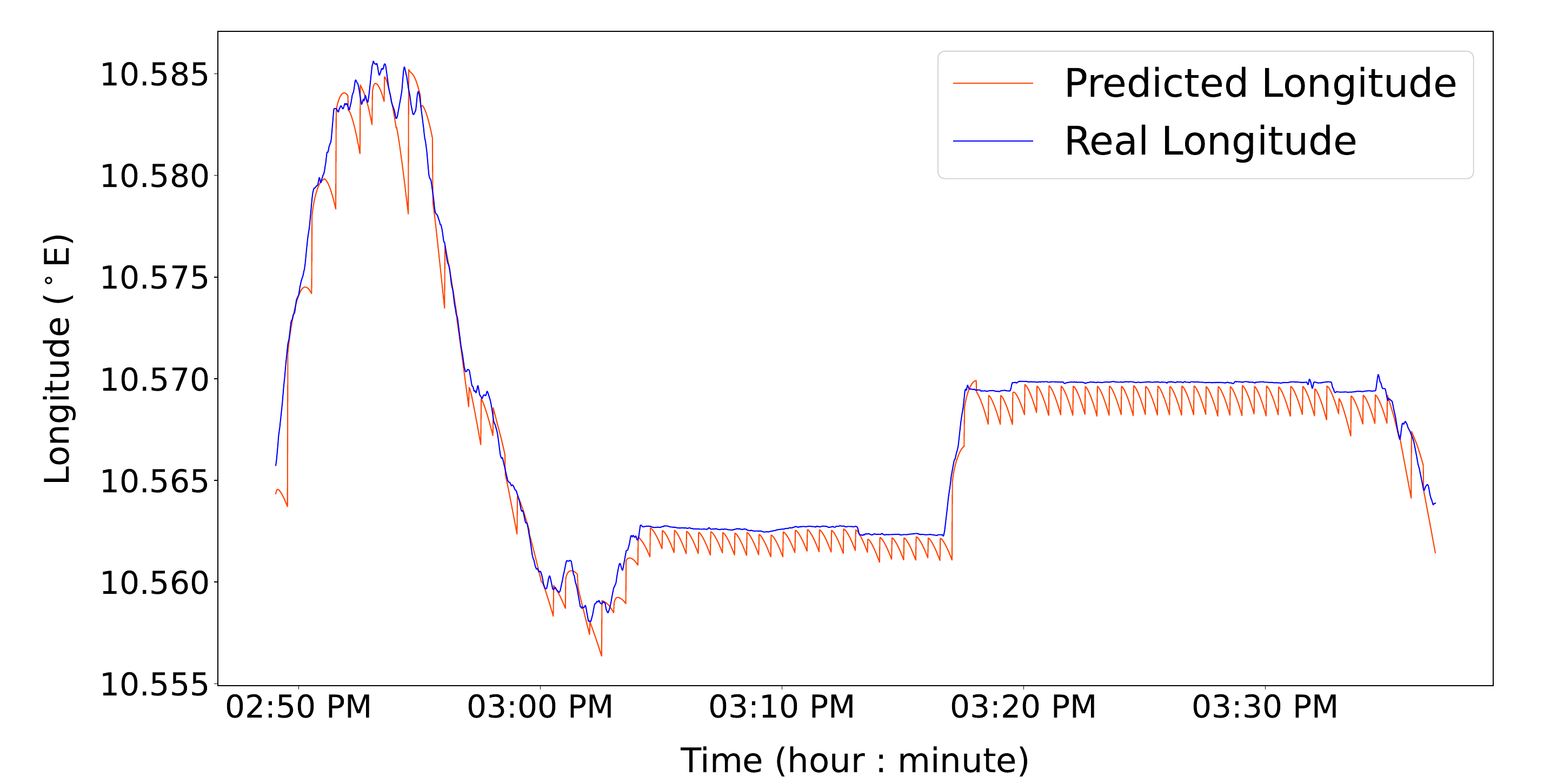}
         \caption{Longitude prediction during the fifth test data set}
         \label{S-Long5}
     \end{subfigure}
     \caption{Longitude predicted by stacked LSTM}
     \label{stacked-Longitude}
\end{figure}
%######################################################################################################
\newpage
\subsection{Bidirectional LSTM}
\begin{figure}
     \centering
     \begin{subfigure}[h]{0.495\textwidth}
         \centering
         \includegraphics[width=\textwidth]{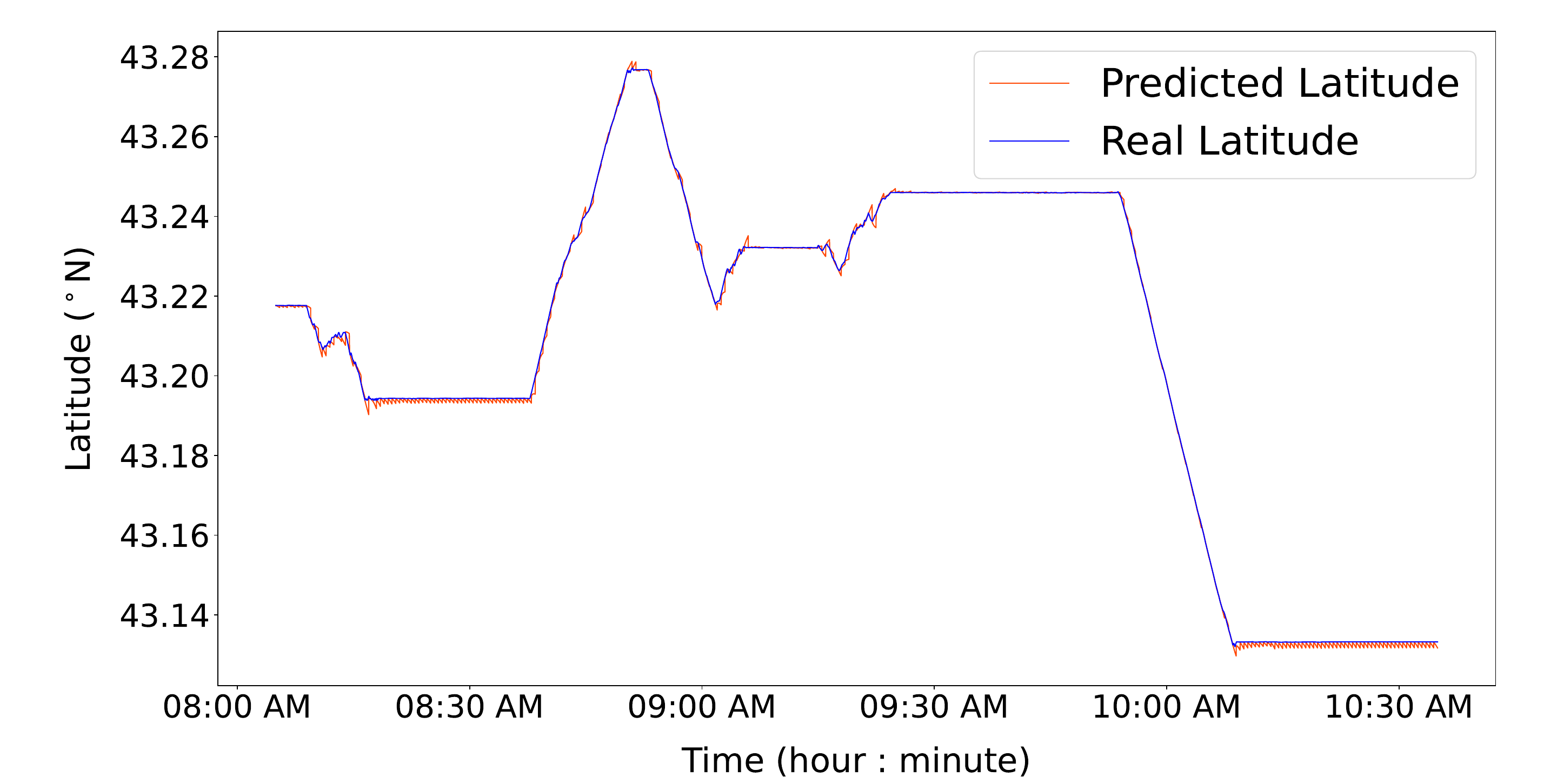}
         \caption{Latitude prediction during training}
         \label{B-Lat-T}
     \end{subfigure}
     \hfill
     \begin{subfigure}[h]{0.495\textwidth}
         \centering
         \includegraphics[width=\textwidth]{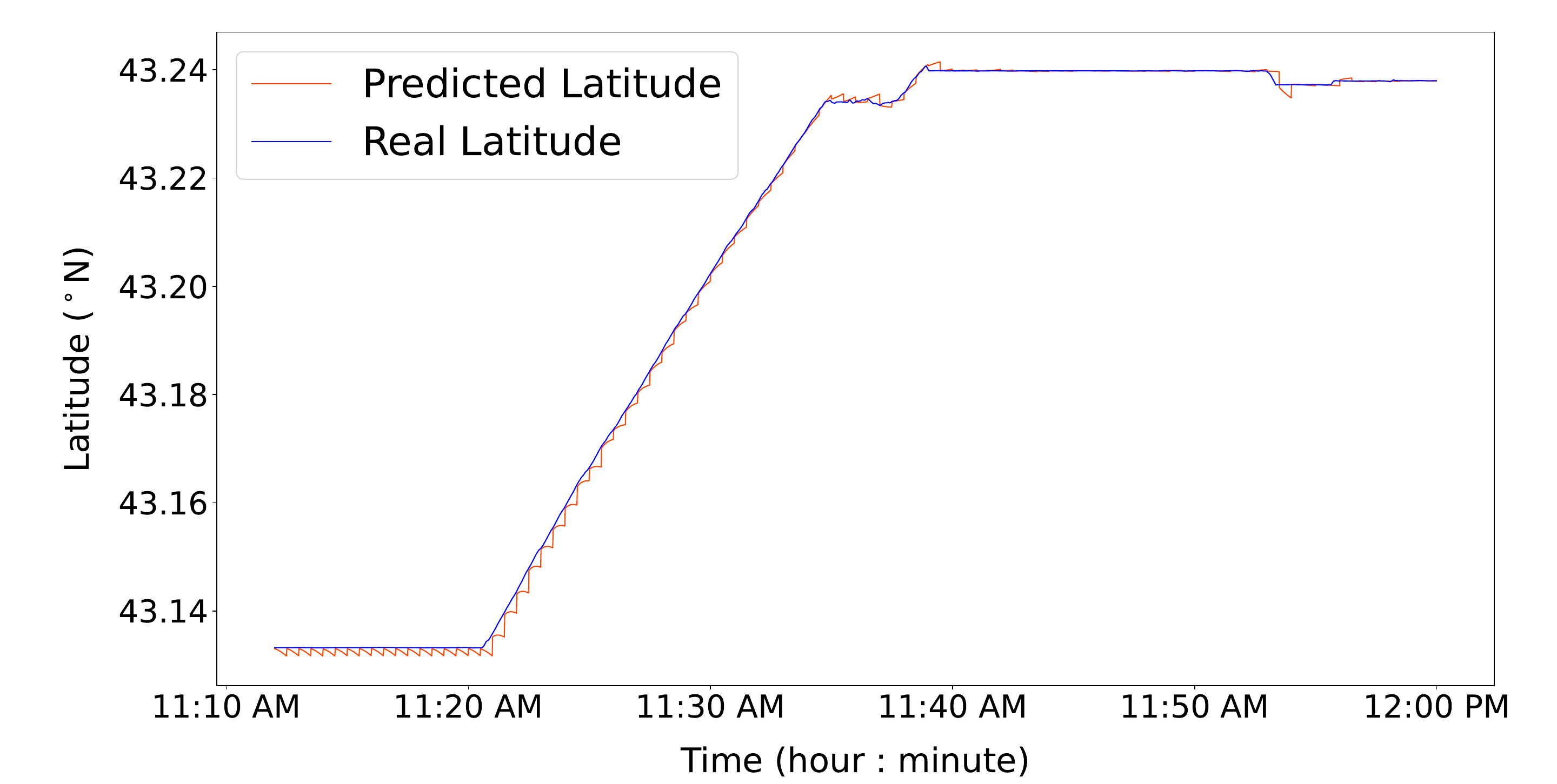}
         \caption{Latitude prediction during the first test data set}
         \label{B-Lat1}
     \end{subfigure}
%###########################
     \centering
     \begin{subfigure}[h]{0.495\textwidth}
         \centering
         \includegraphics[width=\textwidth]{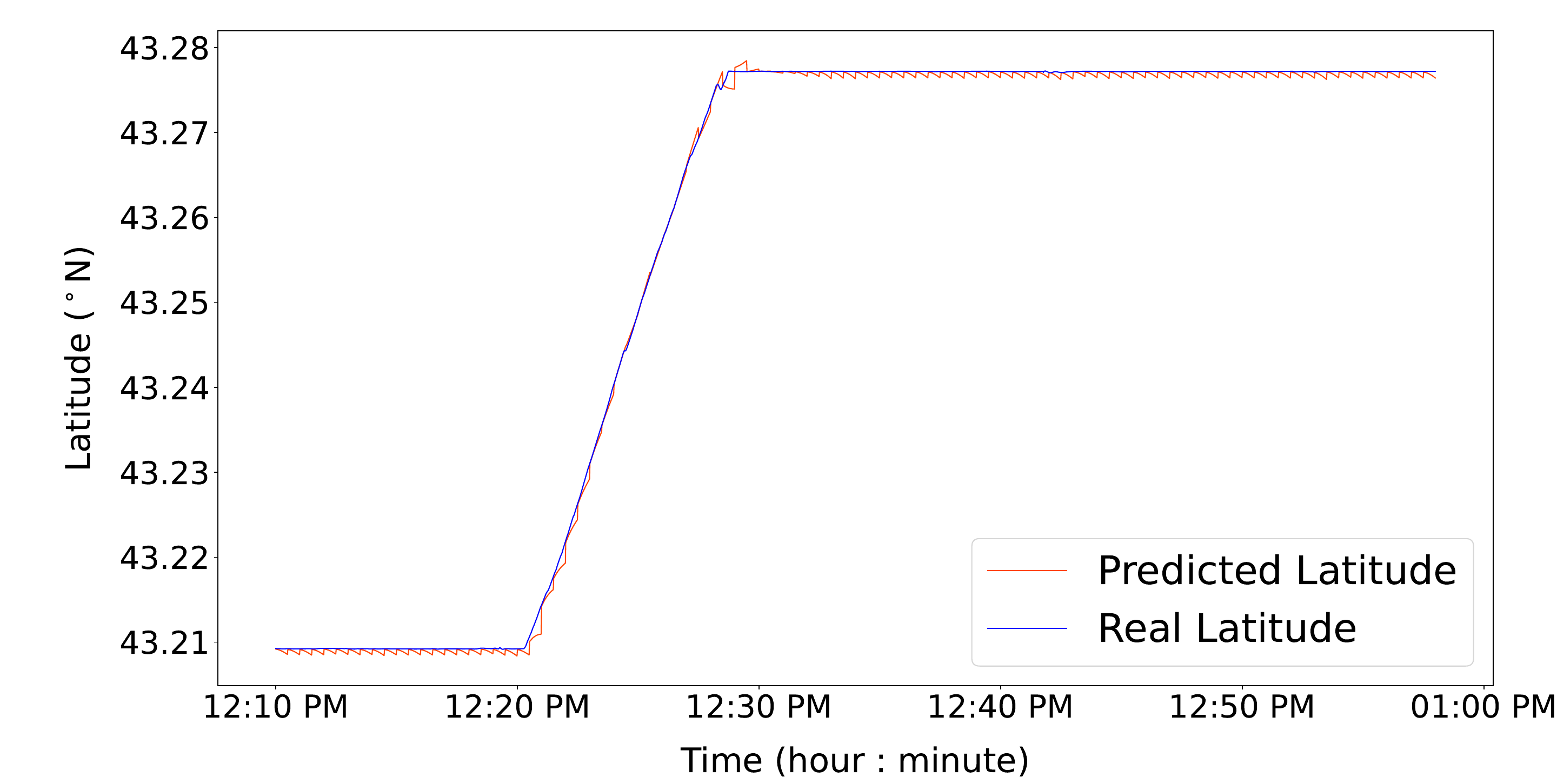}
         \caption{Latitude prediction during the second test data set}
         \label{B-Lat2}
     \end{subfigure}
     \hfill
     \begin{subfigure}[h]{0.495\textwidth}
         \centering
         \includegraphics[width=\textwidth]{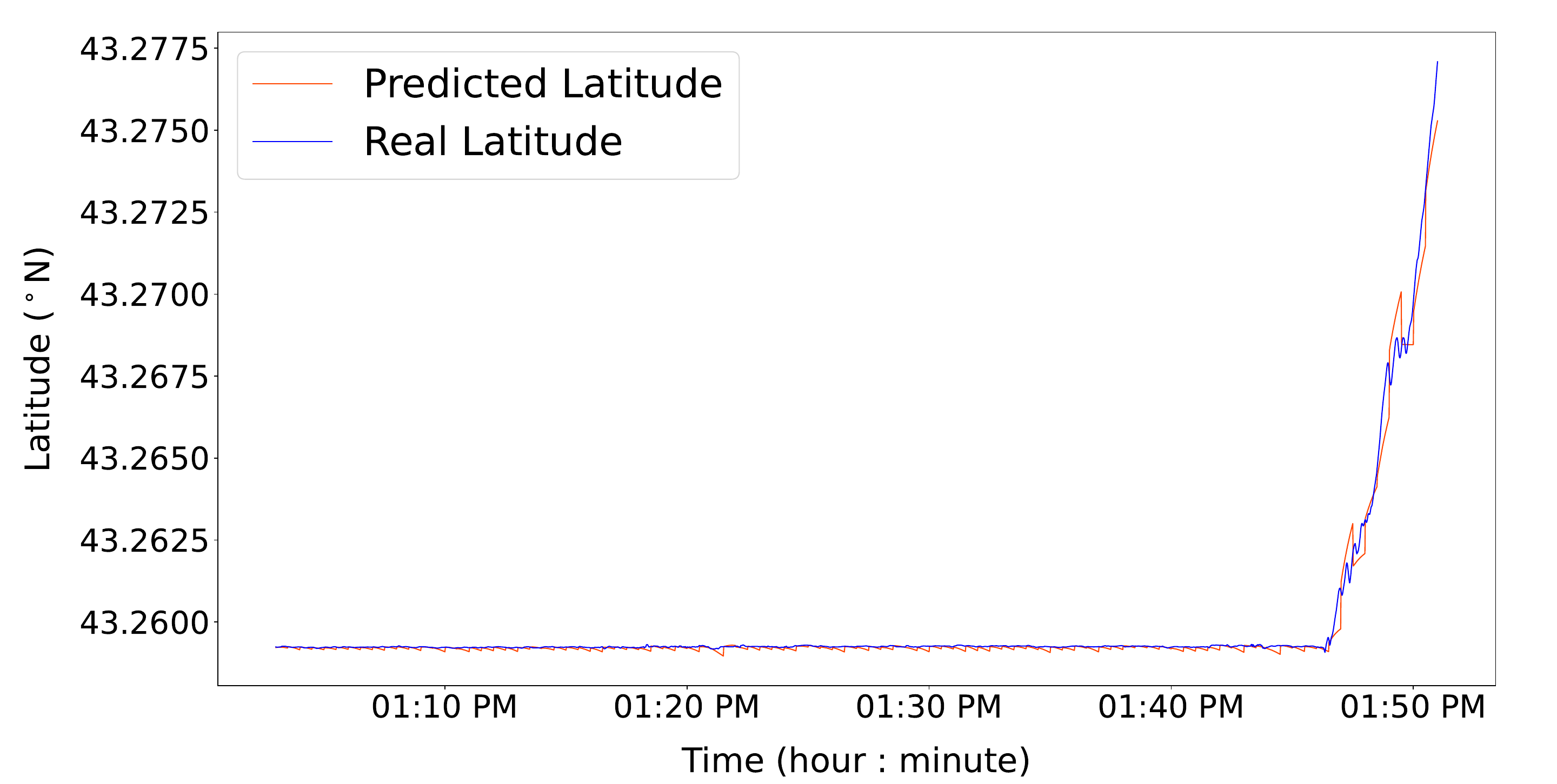}
         \caption{Latitude prediction during the third test data set}
         \label{B-Lat3}
     \end{subfigure}
%###########################
     \centering
     \begin{subfigure}[h]{0.495\textwidth}
         \centering
         \includegraphics[width=\textwidth]{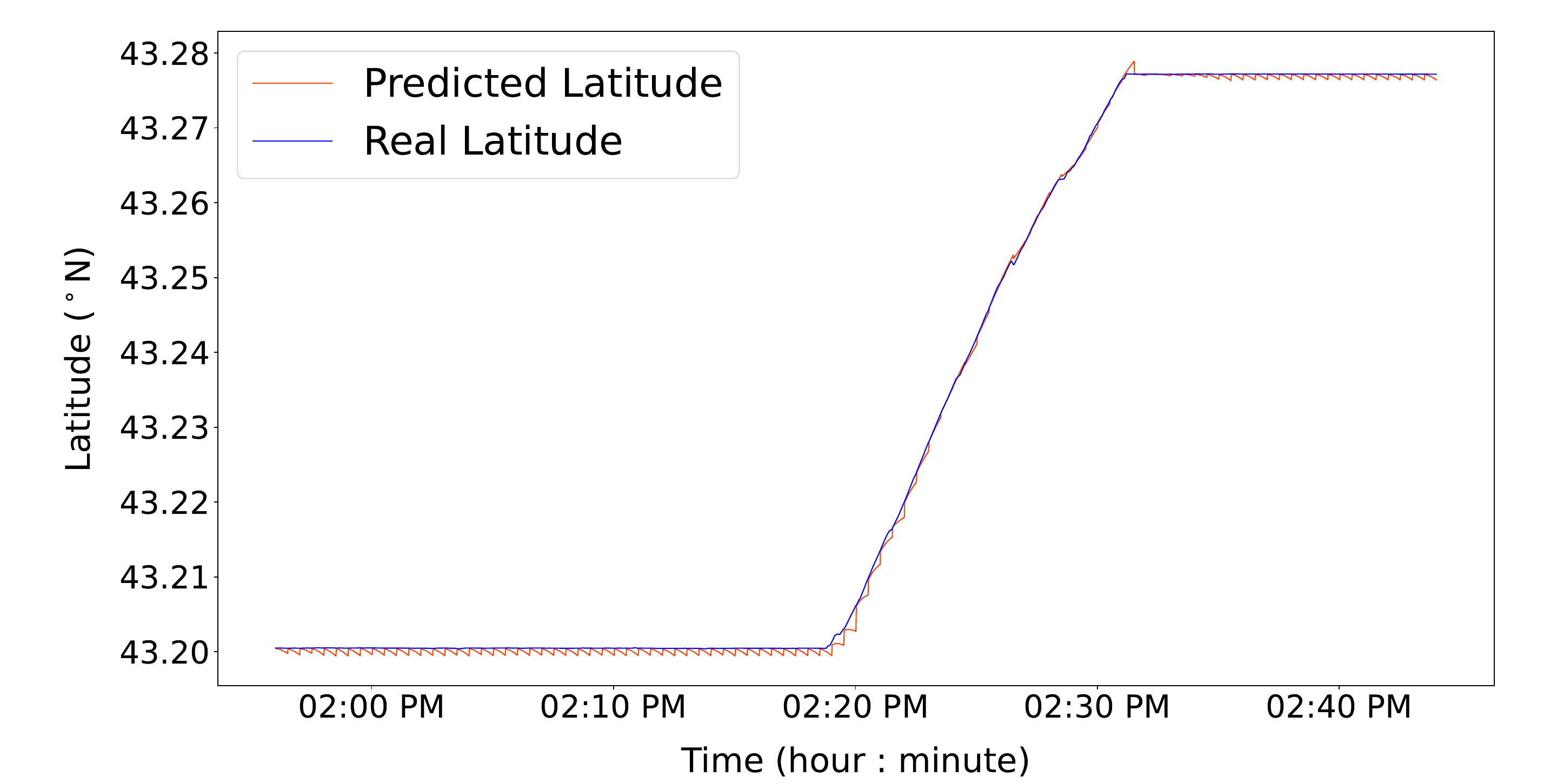}
         \caption{Latitude prediction during the forth test data set}
         \label{B-Lat4}
     \end{subfigure}
     \hfill
     \begin{subfigure}[h]{0.495\textwidth}
         \centering
         \includegraphics[width=\textwidth]{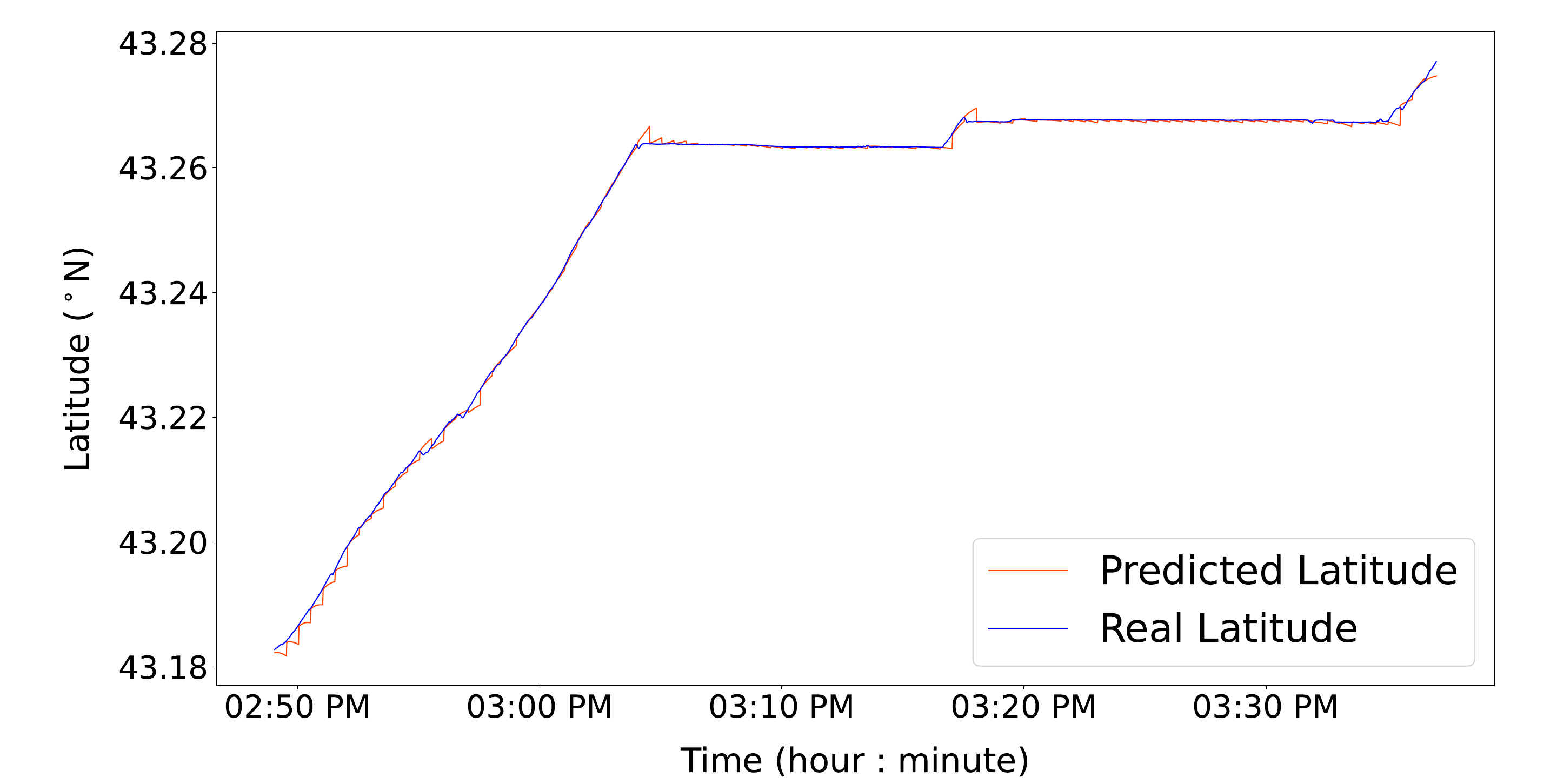}
         \caption{Latitude prediction during the fifth test data set}
         \label{B-Lat5}
     \end{subfigure}
     \caption{Latitude predicted by bidirectional LSTM}
     \label{bidirectional-Latitude}
\end{figure}

\begin{figure}
     \centering
     \begin{subfigure}[p]{0.495\textwidth}
         \centering
         \includegraphics[width=\textwidth]{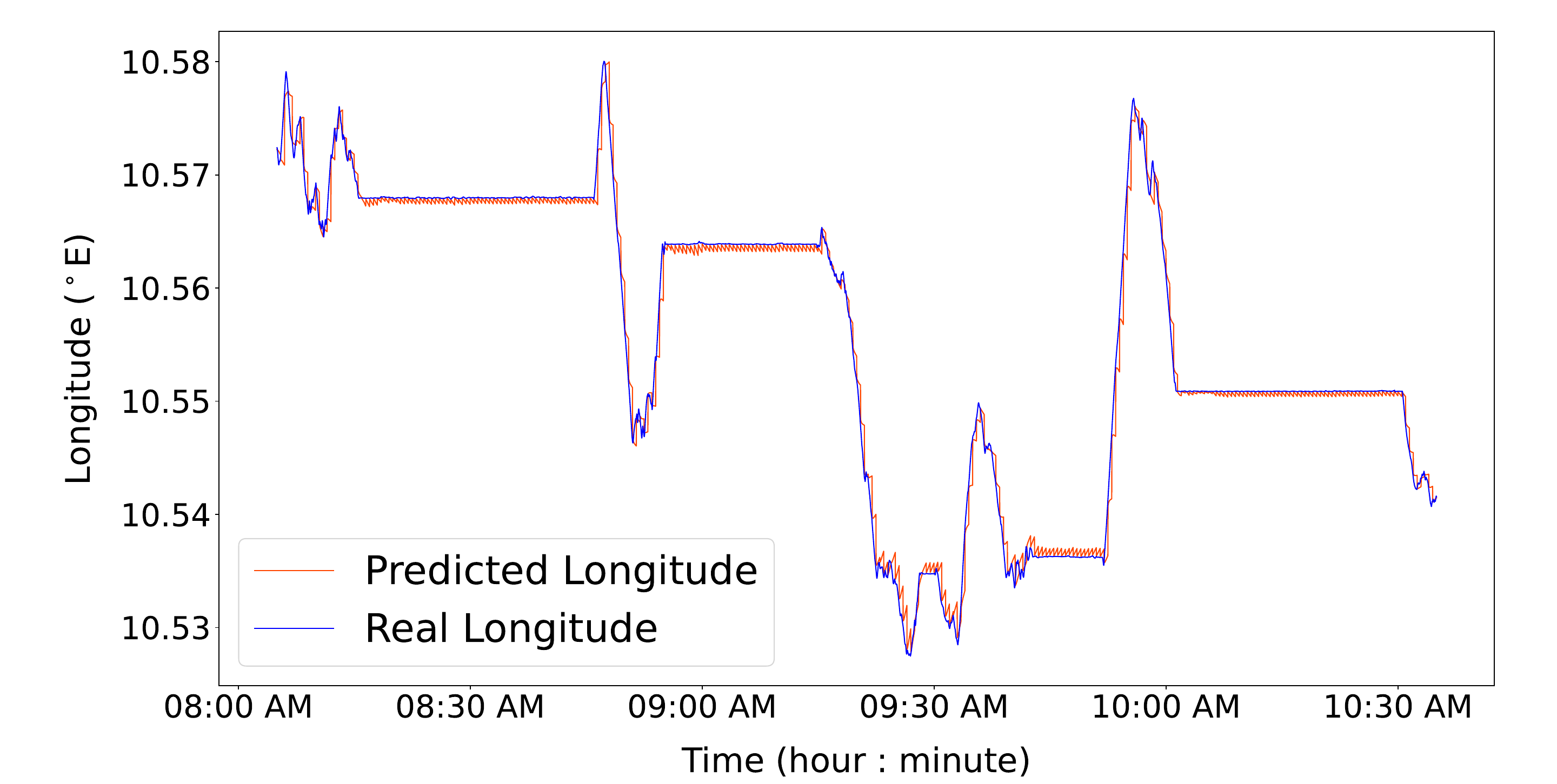}
         \caption{Longitude prediction during training}
         \label{B-Long-T}
     \end{subfigure}
     \hfill
     \begin{subfigure}[p]{0.495\textwidth}
         \centering
         \includegraphics[width=\textwidth]{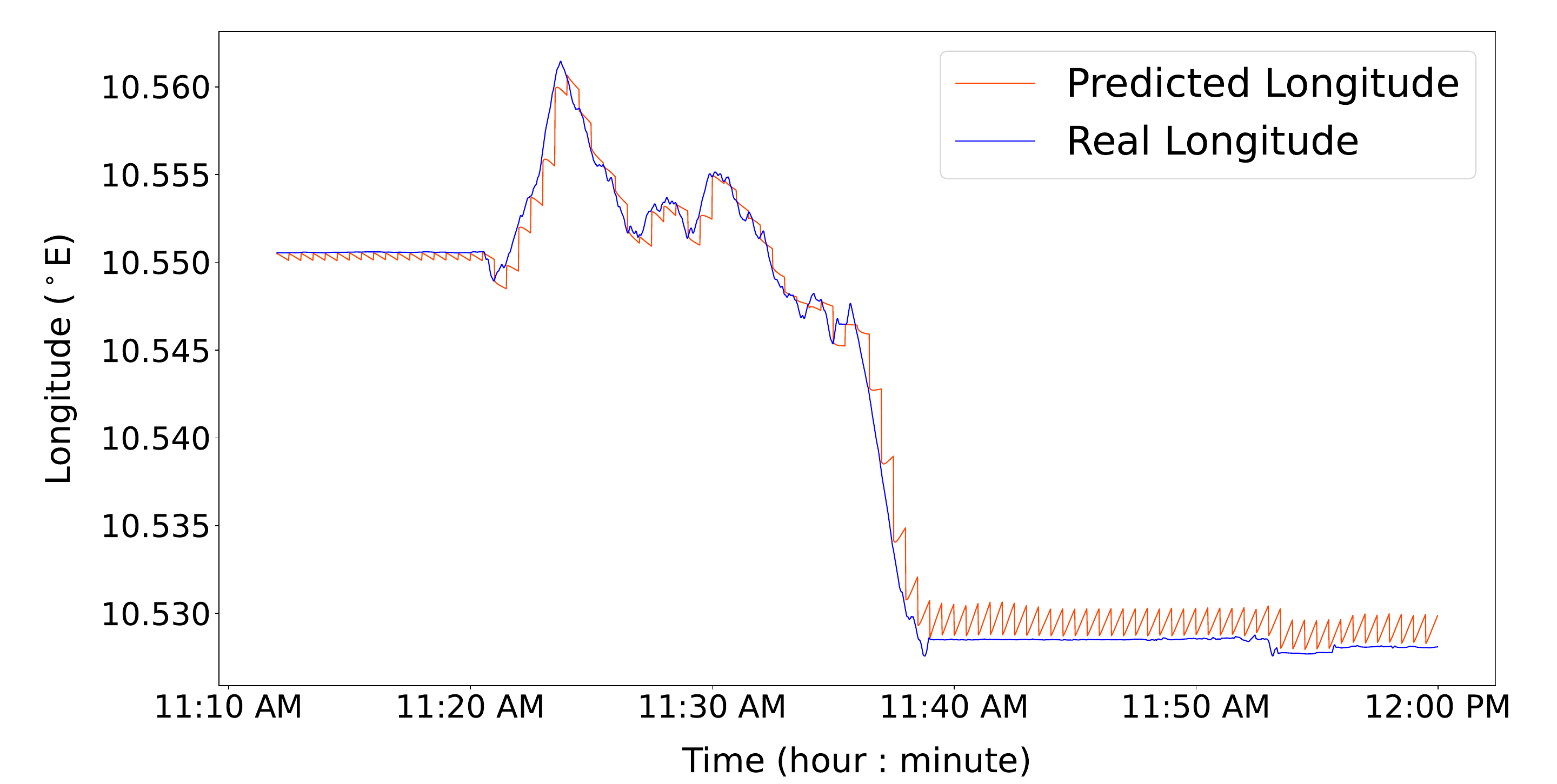}
         \caption{Longitude prediction during the first test data set}
         \label{B-Long1}
     \end{subfigure}
%###########################
     \centering
     \begin{subfigure}[p]{0.495\textwidth}
         \centering
         \includegraphics[width=\textwidth]{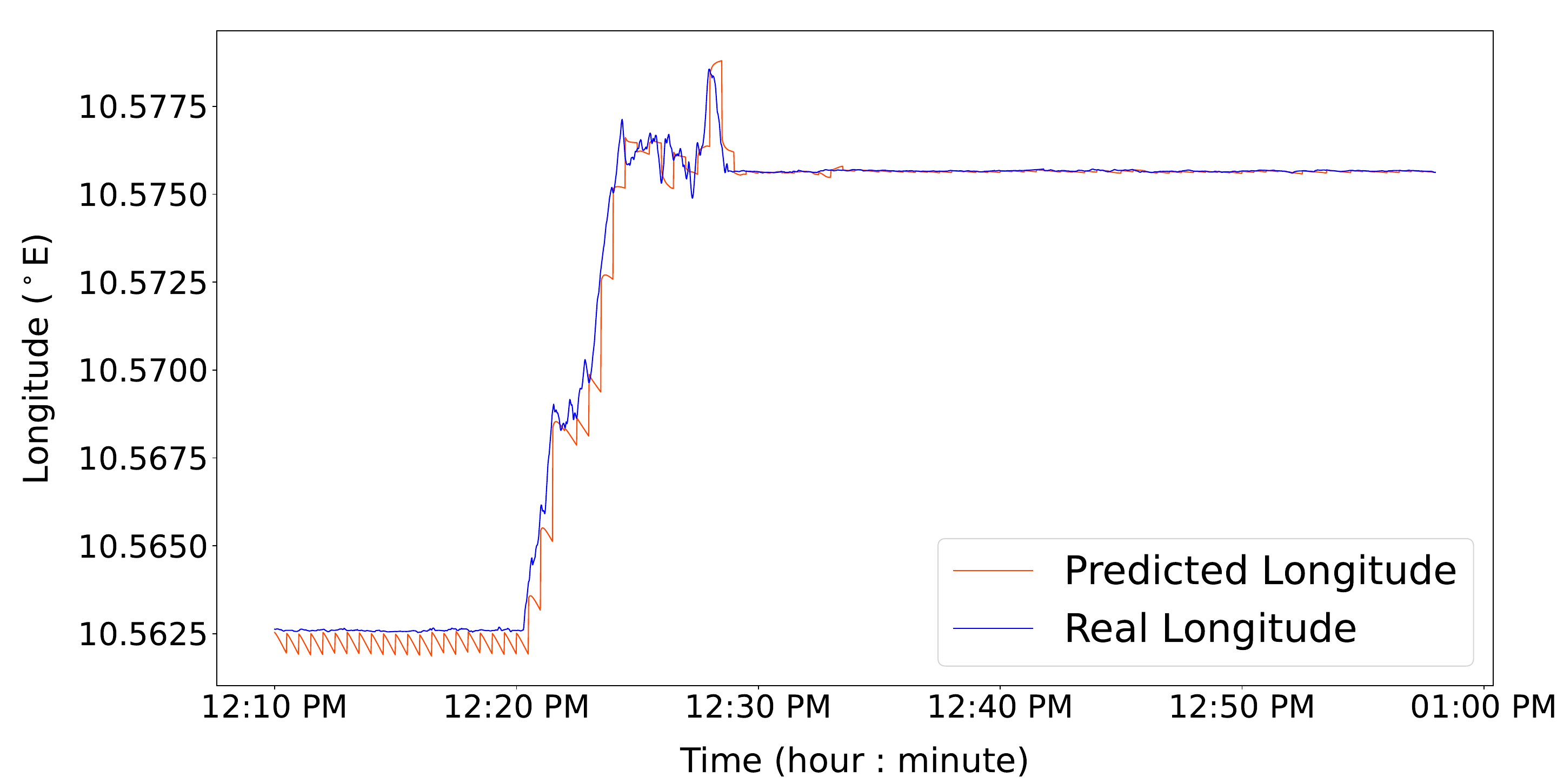}
         \caption{Longitude prediction during the second test data set}
         \label{B-Long2}
     \end{subfigure}
     \hfill
     \begin{subfigure}[p]{0.495\textwidth}
         \centering
         \includegraphics[width=\textwidth]{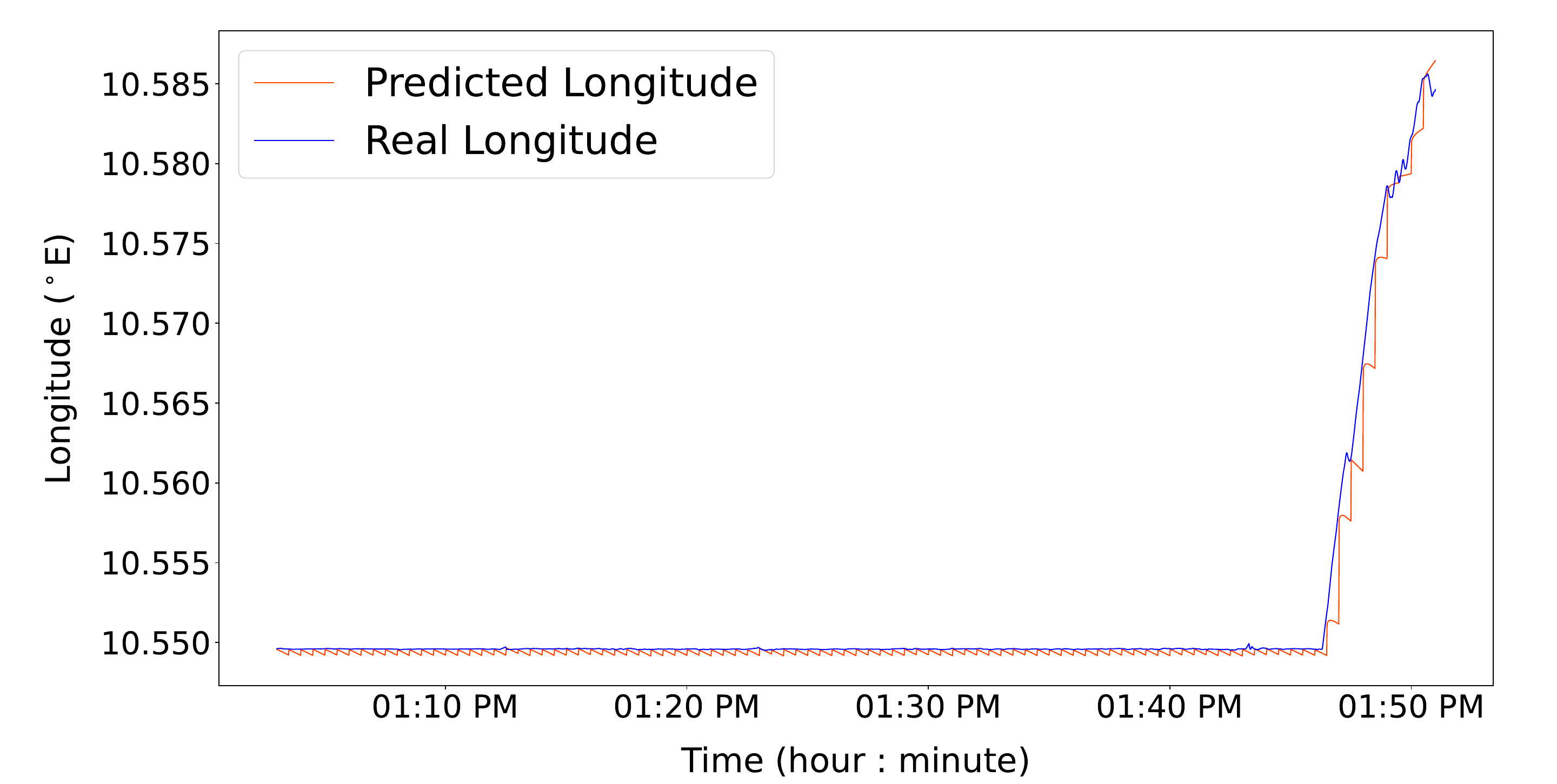}
         \caption{Longitude prediction during the third test data set}
         \label{B-Long3}
     \end{subfigure}
%###########################
     \centering
     \begin{subfigure}[p]{0.495\textwidth}
         \centering
         \includegraphics[width=\textwidth]{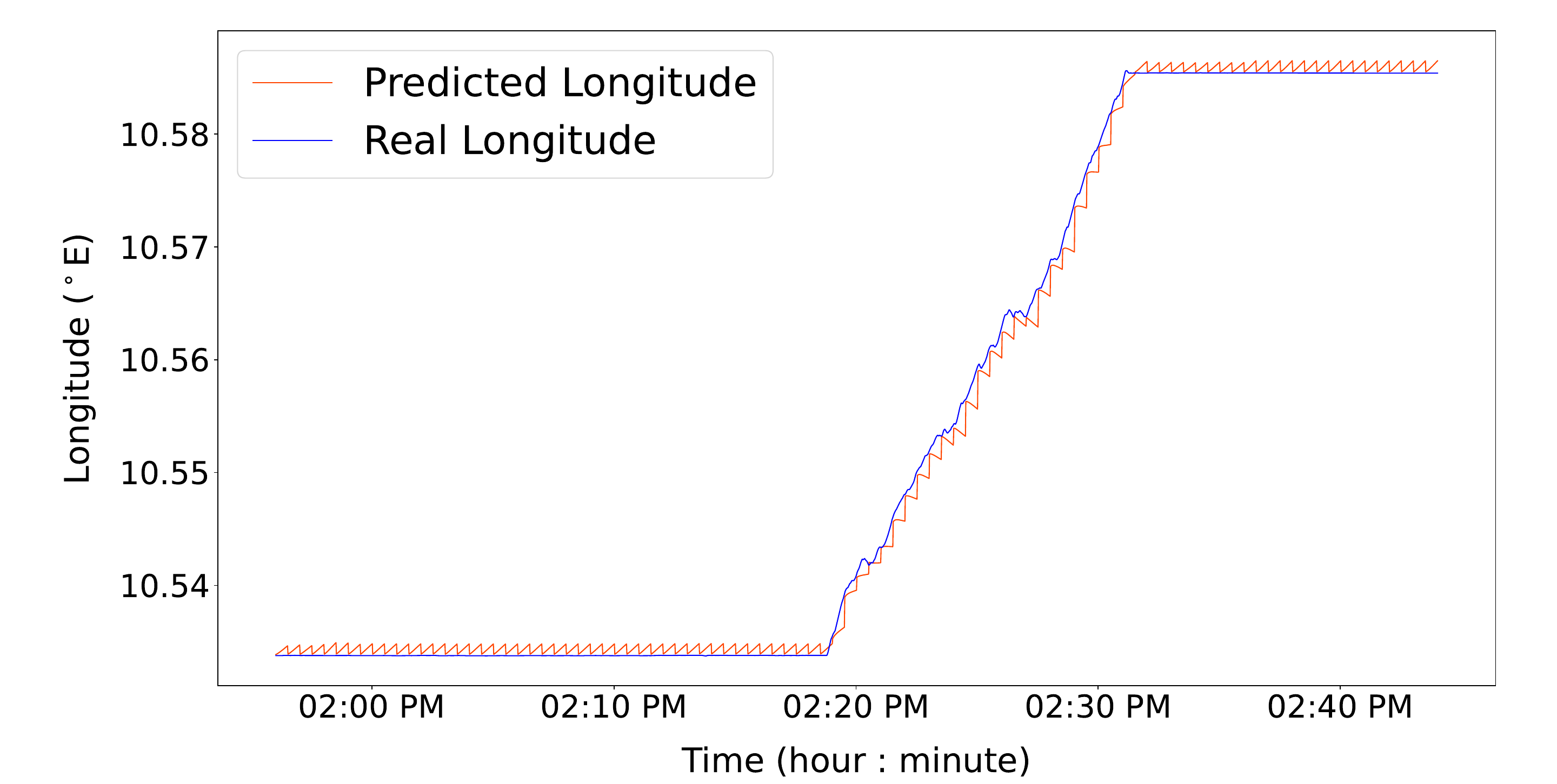}
         \caption{Longitude prediction during the forth test data set}
         \label{B-Long4}
     \end{subfigure}
     \hfill
     \begin{subfigure}[p]{0.495\textwidth}
         \centering
         \includegraphics[width=\textwidth]{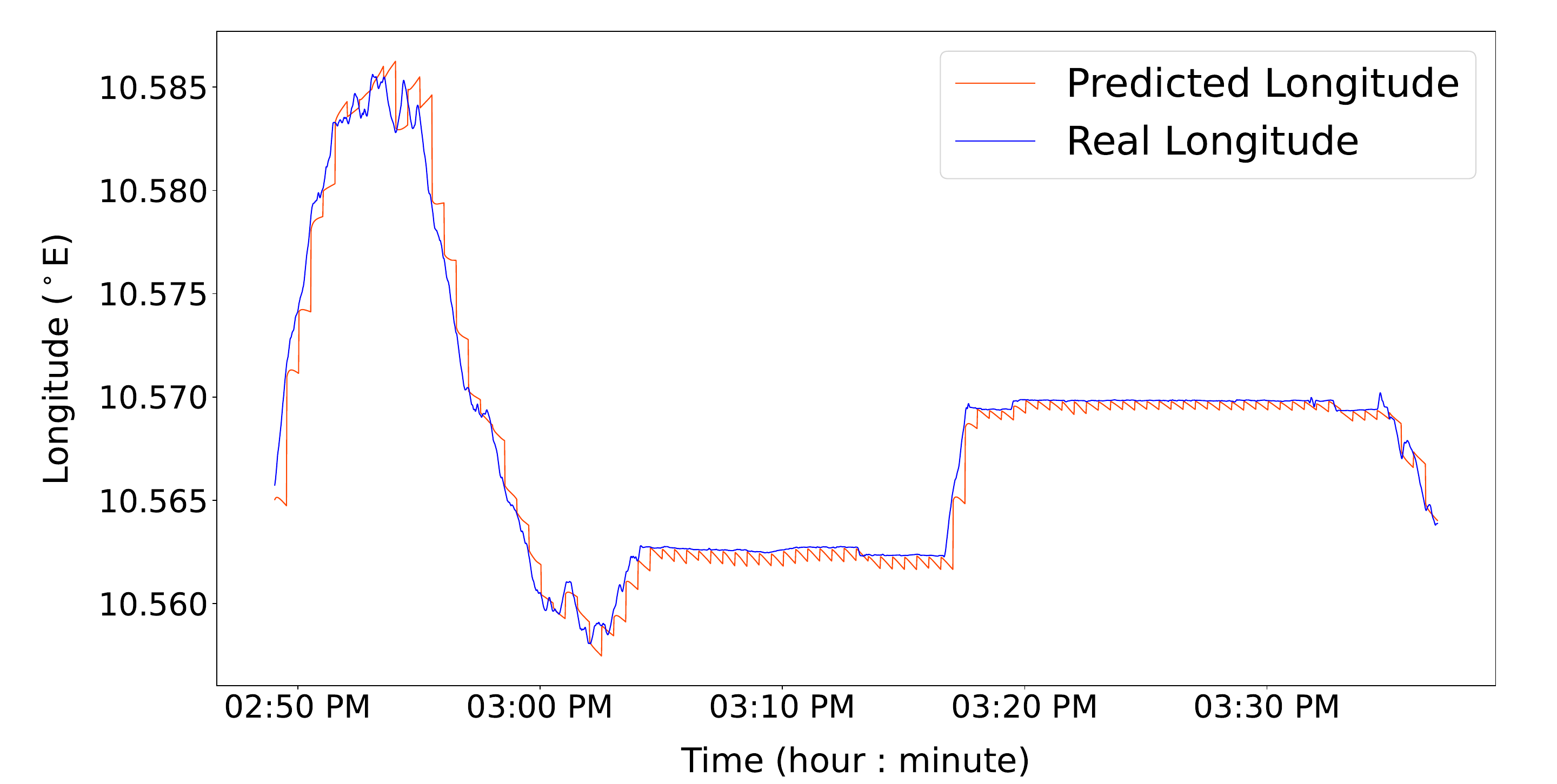}
         \caption{Longitude prediction during the fifth test data set}
         \label{B-Long5}
     \end{subfigure}
     \label{fig:linearRegressionModelPredictions}
     \caption{Longitude predicted by bidirectional LSTM}
     \label{bidirectional-Longitude}
\end{figure}

The main difference between vanilla LSTM and bidirectional LSTM models is that the latter can interpret data from both forward and backward directions, potentially leading to improved accuracy in certain cases. However, this is not a universal rule that applies to all data sets. In our study, we implemented two separate bidirectional LSTM models for latitude and longitude prediction, both of which had 32 neurons in their hidden layer. We evaluated the performance of the LSTM model for latitude prediction during training and testing, as shown in Fig. \ref{bidirectional-Latitude}. For longitude prediction, we also evaluated the performance of the model during training and testing, which are illustrated in Fig. \ref{bidirectional-Longitude}. The performance of the bidirectional LSTM in predicting both latitude and longitude is approximately better than that of the vanilla LSTM and less oscillations can be observed in the performance of the bidirectional LSTM. 
\floatplacement{figure}{!h}
%#########################################################Longitude#############################################
\floatplacement{figure}{!p}
%######################################################################################################
\newpage
\subsection{Encoder-decoder LSTM} 
\begin{figure}
     \centering
     \begin{subfigure}[h]{0.495\textwidth}
         \centering
         \includegraphics[width=\textwidth]{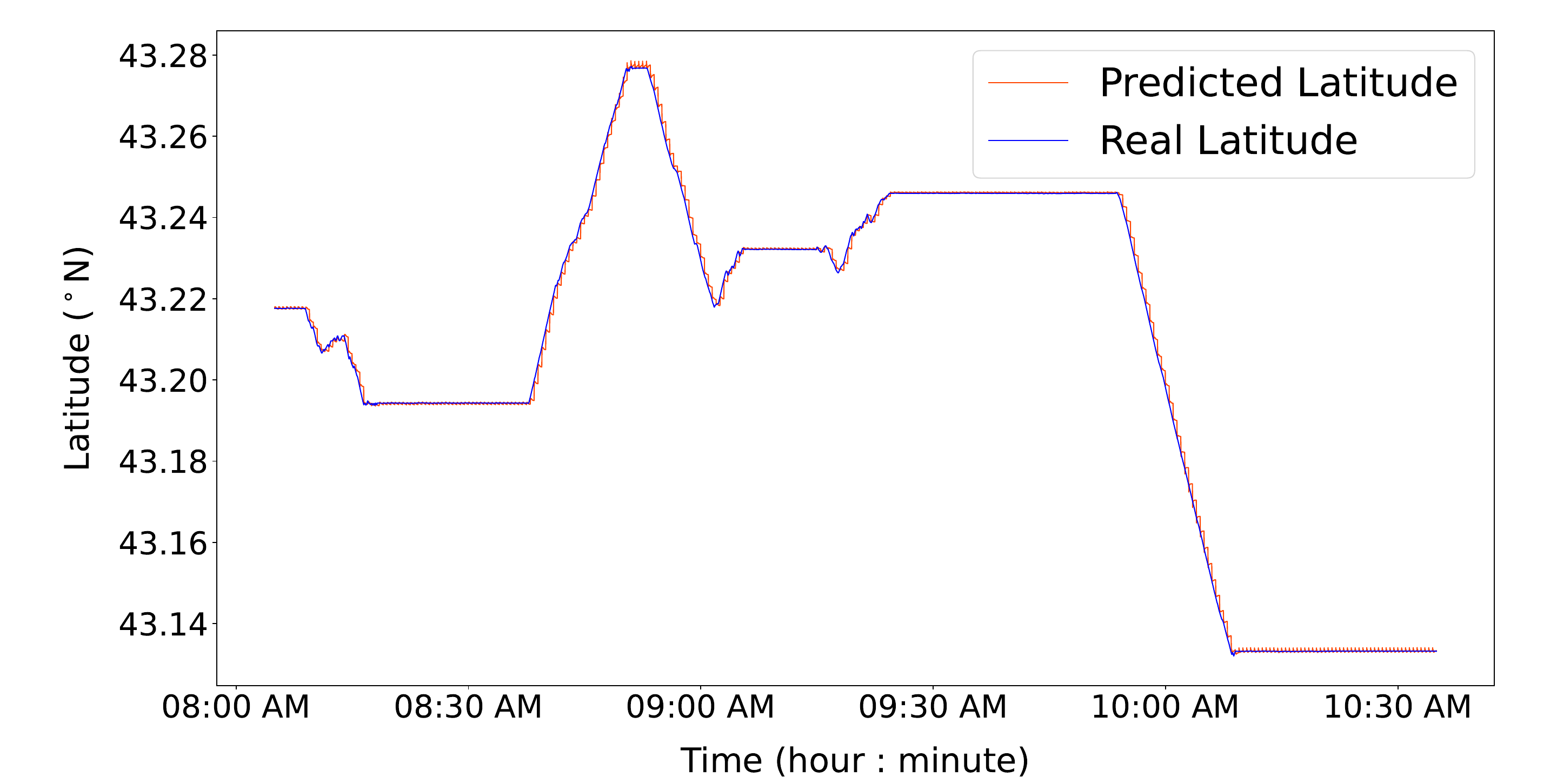}
         \caption{Latitude prediction during training}
         \label{ED-Lat-T}
     \end{subfigure}
     \hfill
     \begin{subfigure}[h]{0.495\textwidth}
         \centering
         \includegraphics[width=\textwidth]{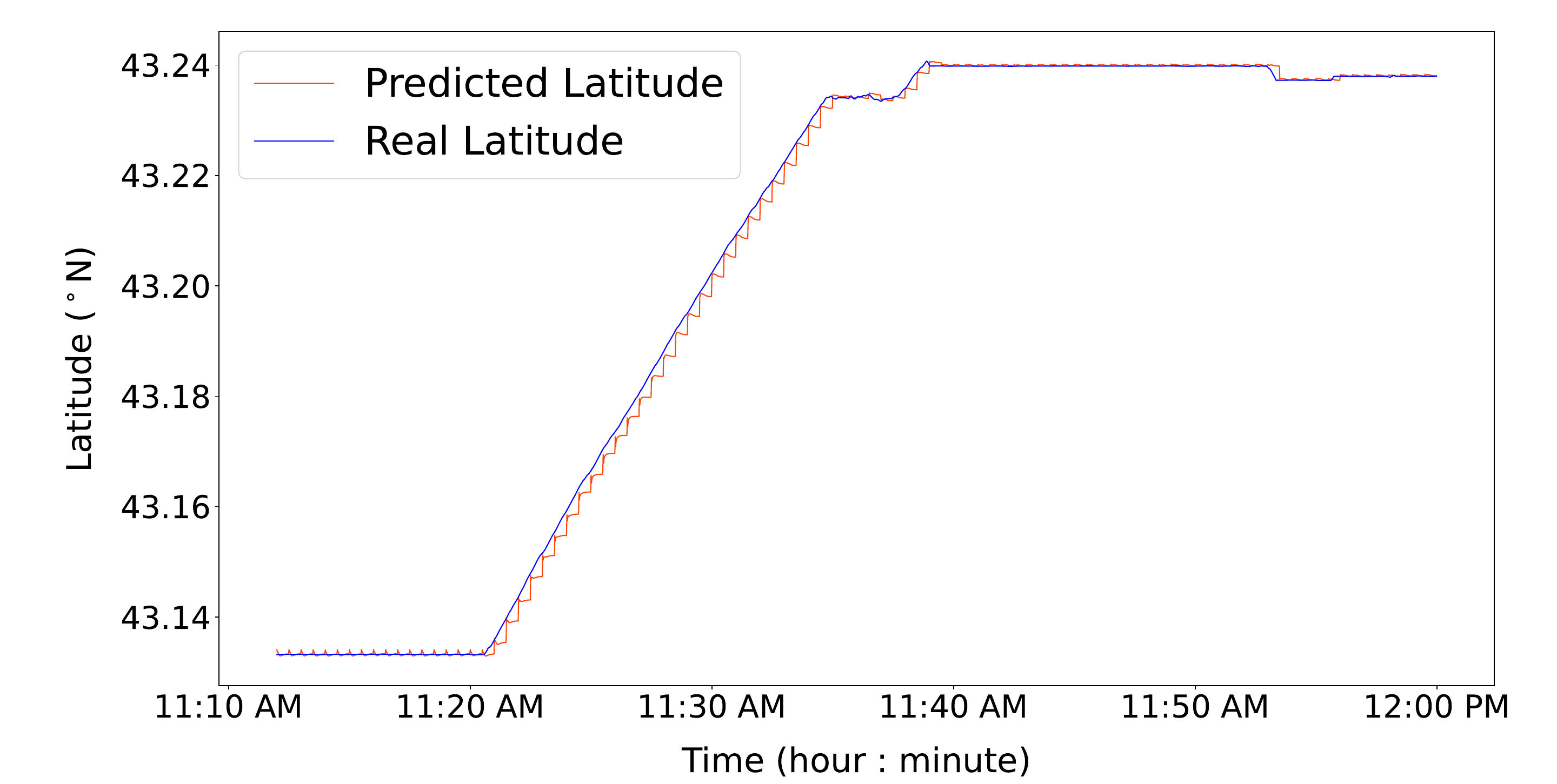}
         \caption{Latitude prediction during the first test data set}
         \label{ED-Lat1}
     \end{subfigure}
     \label{fig:linearRegressionModelPredictions}
%###########################
     \centering
     \begin{subfigure}[h]{0.495\textwidth}
         \centering
         \includegraphics[width=\textwidth]{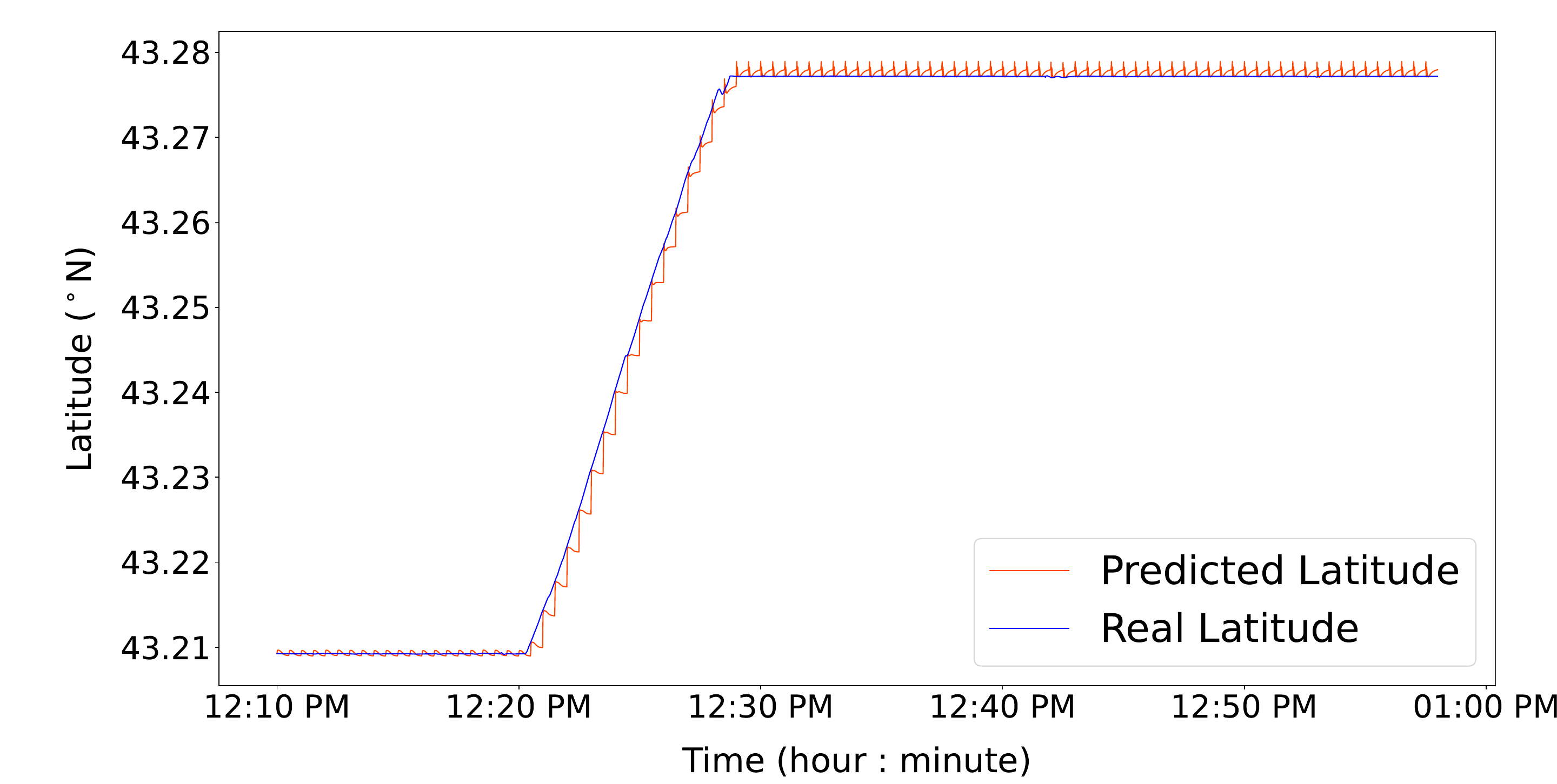}
         \caption{Latitude prediction during the second test data set}
         \label{ED-Lat2}
     \end{subfigure}
     \hfill
     \begin{subfigure}[h]{0.495\textwidth}
         \centering
         \includegraphics[width=\textwidth]{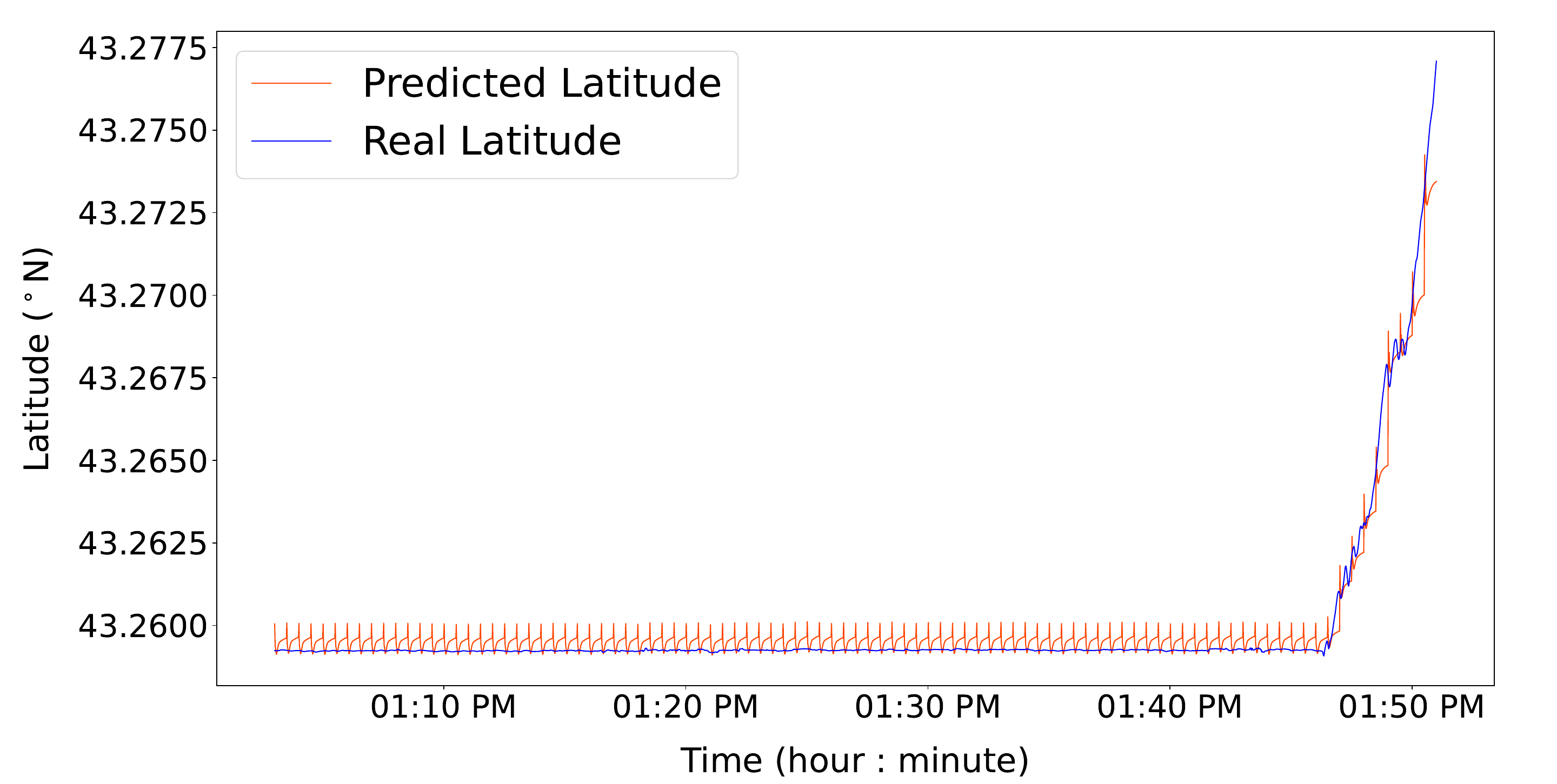}
         \caption{Latitude prediction during the third test data set}
         \label{ED-Lat3}
     \end{subfigure}
%###########################
     \centering
     \begin{subfigure}[h]{0.495\textwidth}
         \centering
         \includegraphics[width=\textwidth]{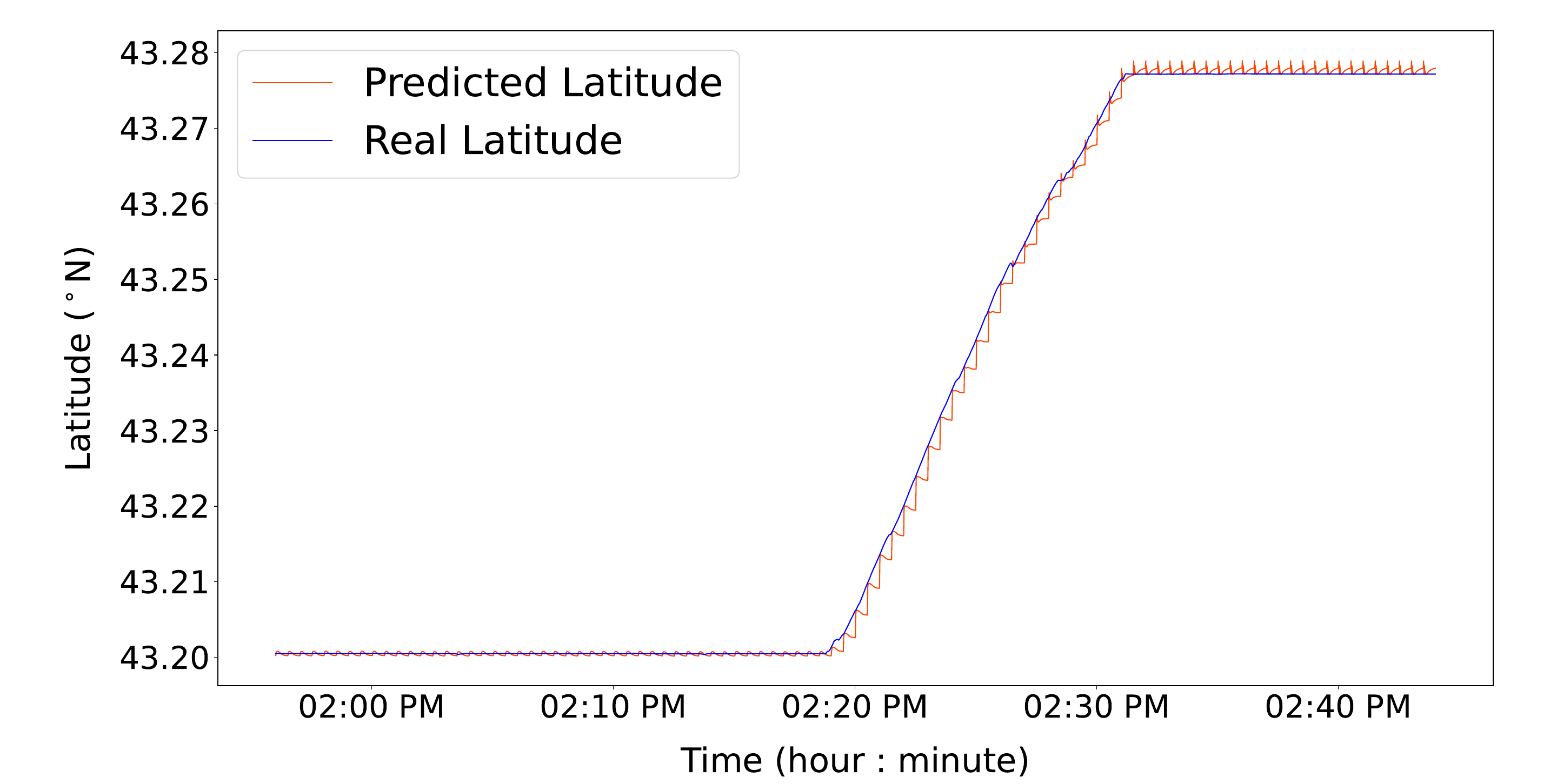}
         \caption{Latitude prediction during the forth test data set}
         \label{ED-Lat4}
     \end{subfigure}
     \hfill
     \begin{subfigure}[h]{0.495\textwidth}
         \centering
         \includegraphics[width=\textwidth]{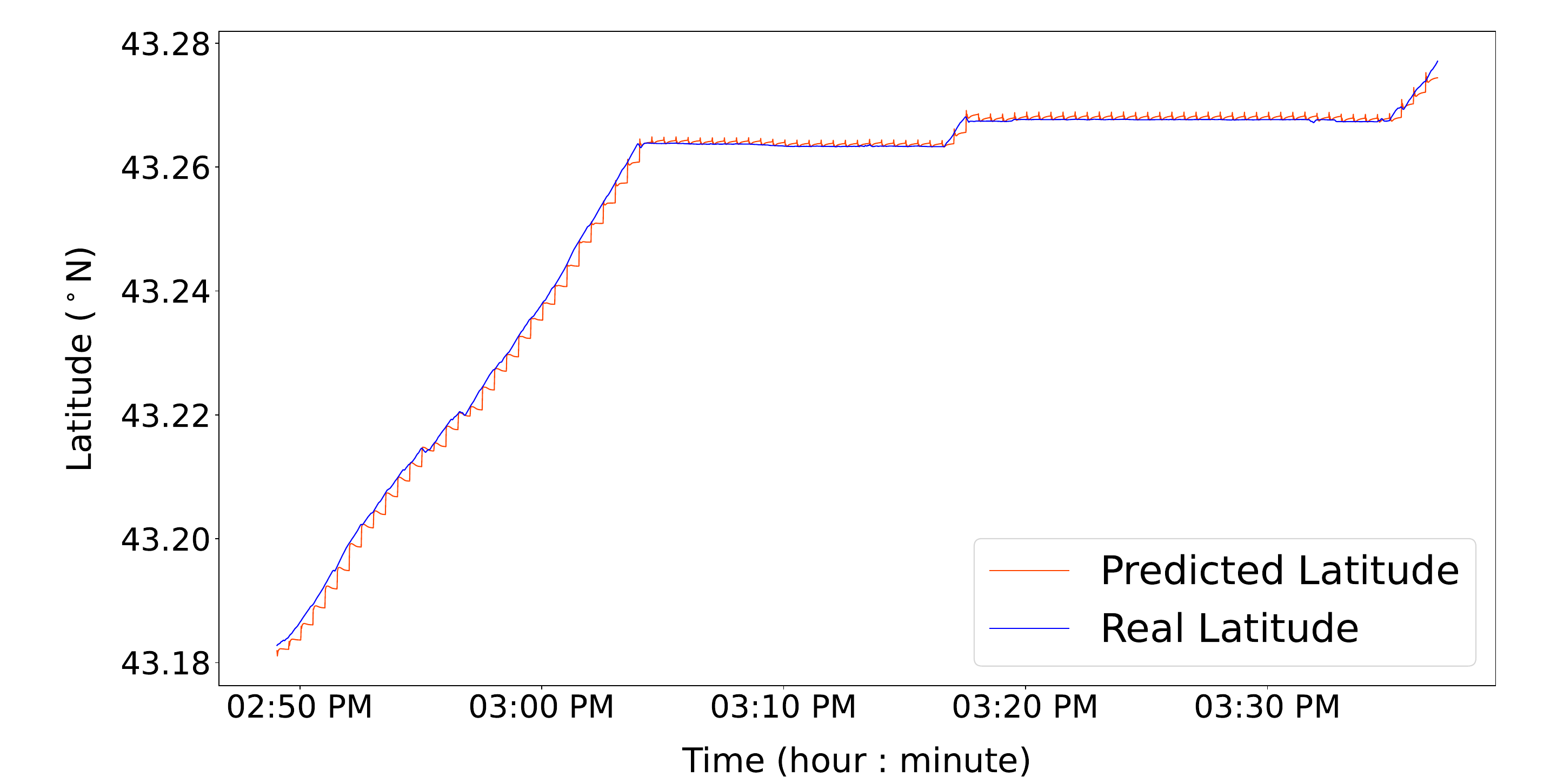}
         \caption{Latitude prediction during the fifth test data set}
         \label{ED-Lat5}
     \end{subfigure}
     \label{fig:linearRegressionModelPredictions}
     \caption{Latitude predicted by encoder-decoder LSTM}
     \label{encoder-Latitude}
\end{figure}

\begin{figure}
     \centering
     \begin{subfigure}[p]{0.495\textwidth}
         \centering
         \includegraphics[width=\textwidth]{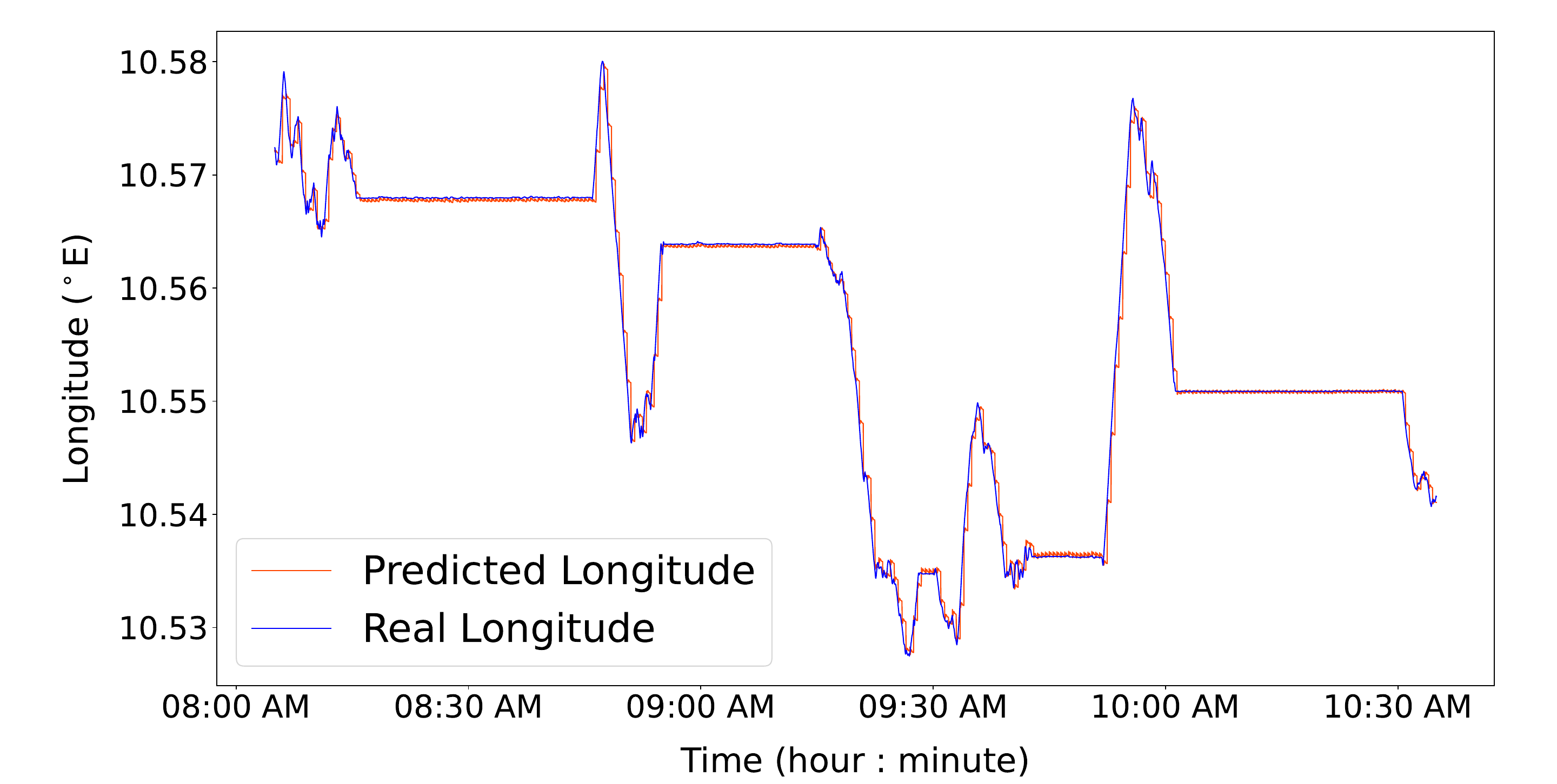}
         \caption{Longitude prediction during training}
         \label{ED-Long-T}
     \end{subfigure}
     \hfill
     \begin{subfigure}[p]{0.495\textwidth}
         \centering
         \includegraphics[width=\textwidth]{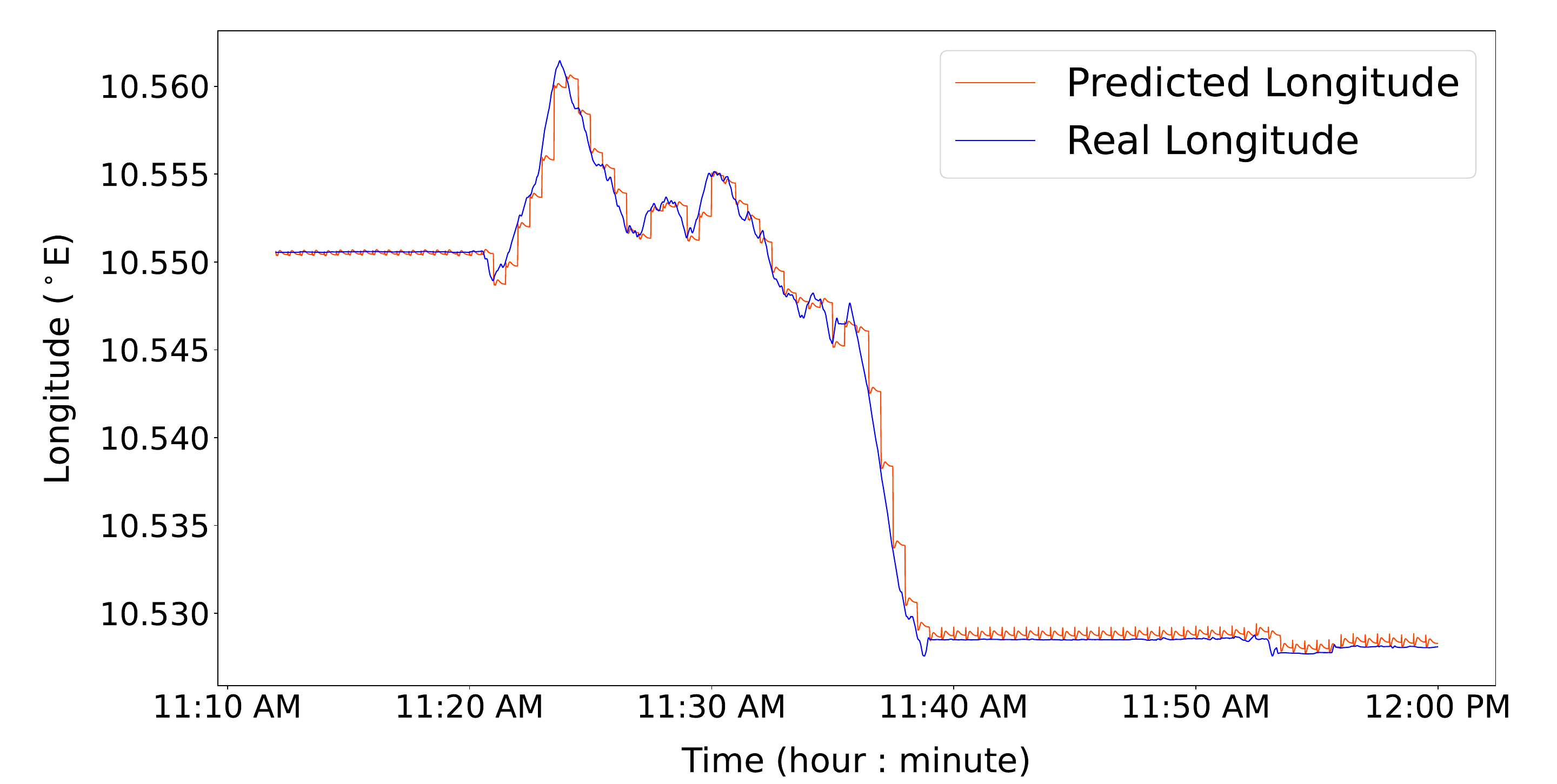}
         \caption{Longitude prediction during the first test data set}
         \label{ED-Long1}
     \end{subfigure}
%###########################
     \centering
     \begin{subfigure}[p]{0.495\textwidth}
         \centering
         \includegraphics[width=\textwidth]{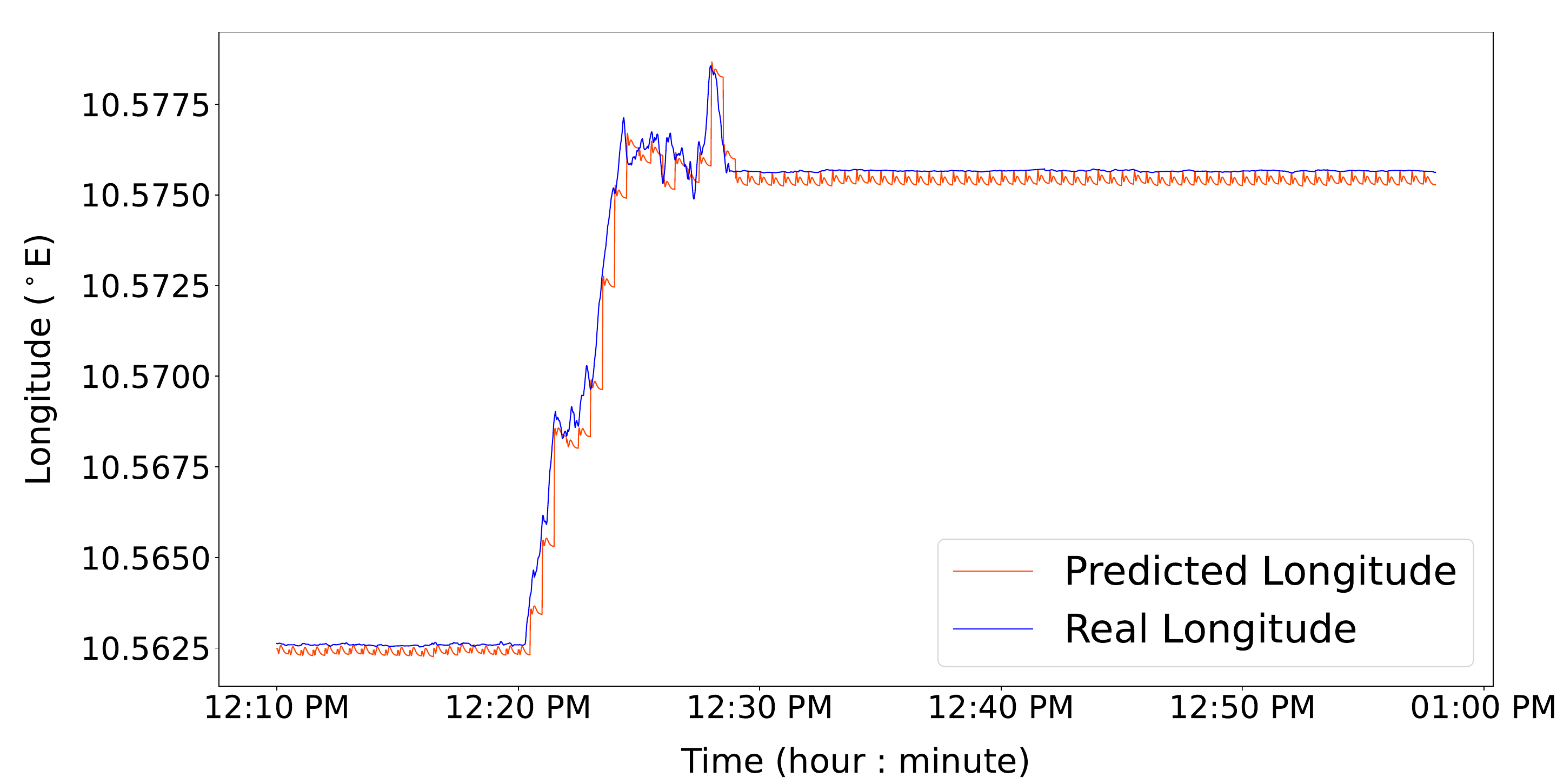}
         \caption{Longitude prediction during the second test data set}
         \label{ED-Long2}
     \end{subfigure}
     \hfill
     \begin{subfigure}[p]{0.495\textwidth}
         \centering
         \includegraphics[width=\textwidth]{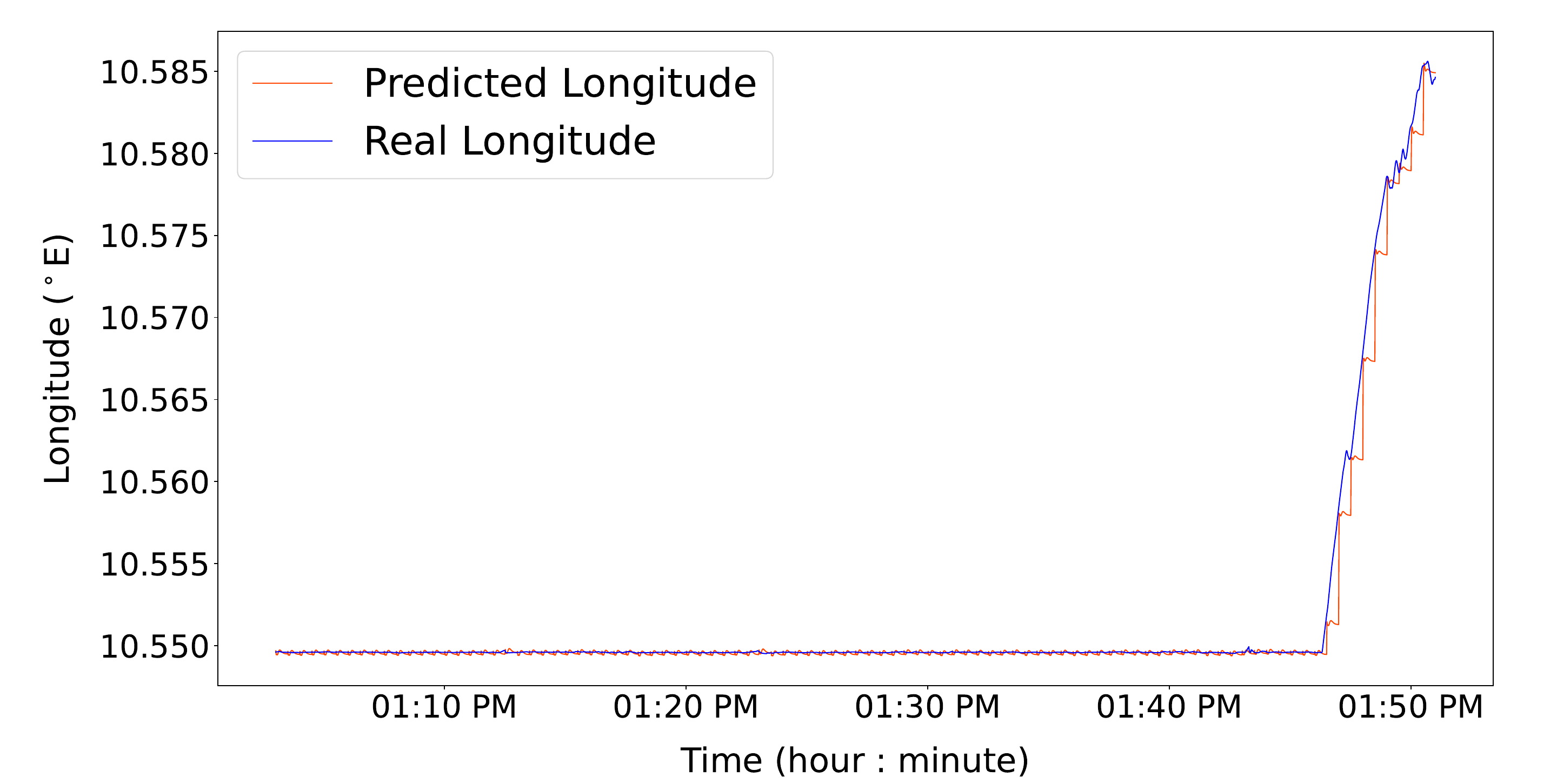}
         \caption{Longitude prediction during the third test data set}
         \label{ED-Long3}
     \end{subfigure}
%###########################
     \centering
     \begin{subfigure}[p]{0.495\textwidth}
         \centering
         \includegraphics[width=\textwidth]{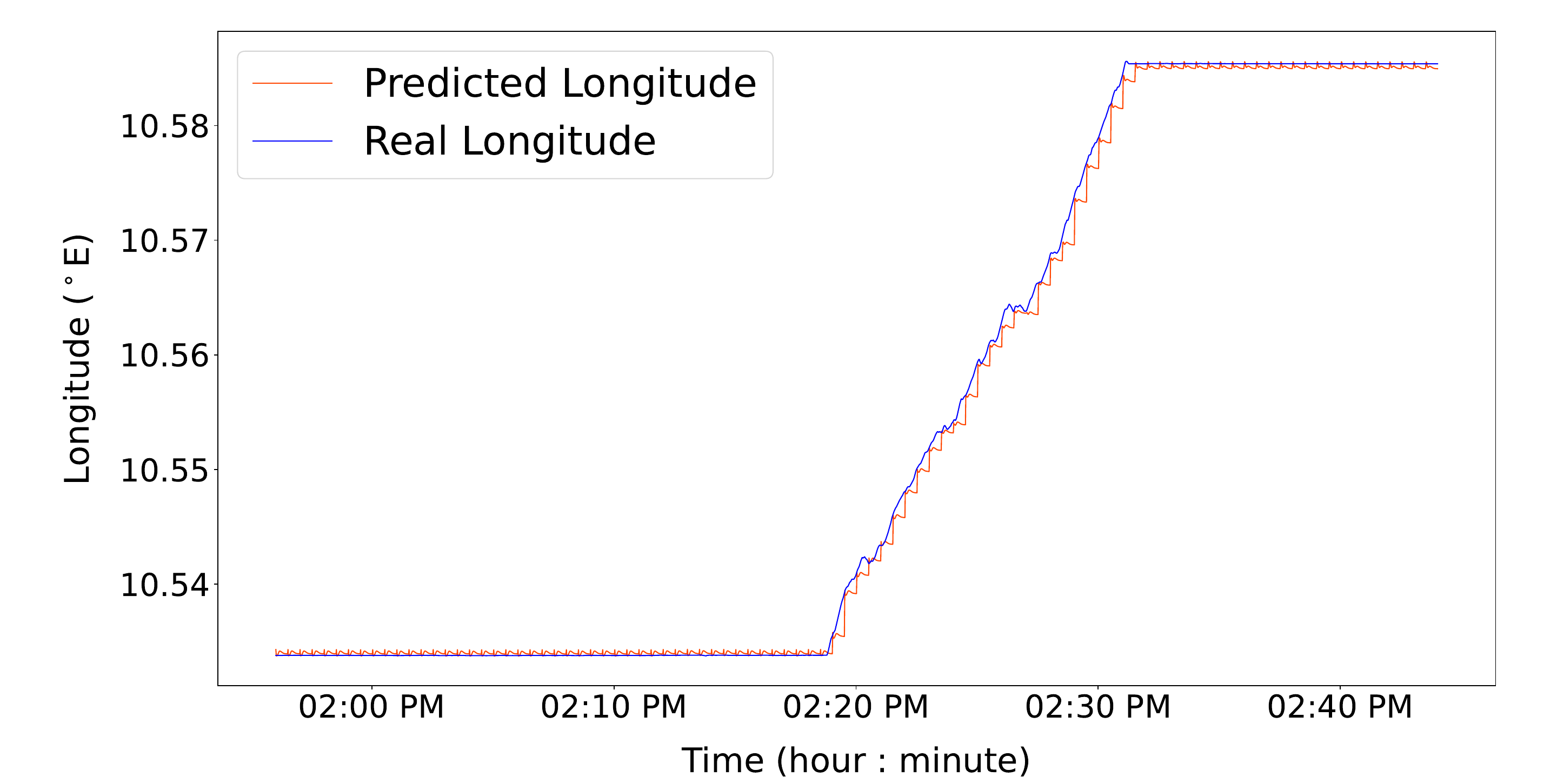}
         \caption{Longitude prediction during the forth test data set}
         \label{ED-Long4}
     \end{subfigure}
     \hfill
     \begin{subfigure}[p]{0.495\textwidth}
         \centering
         \includegraphics[width=\textwidth]{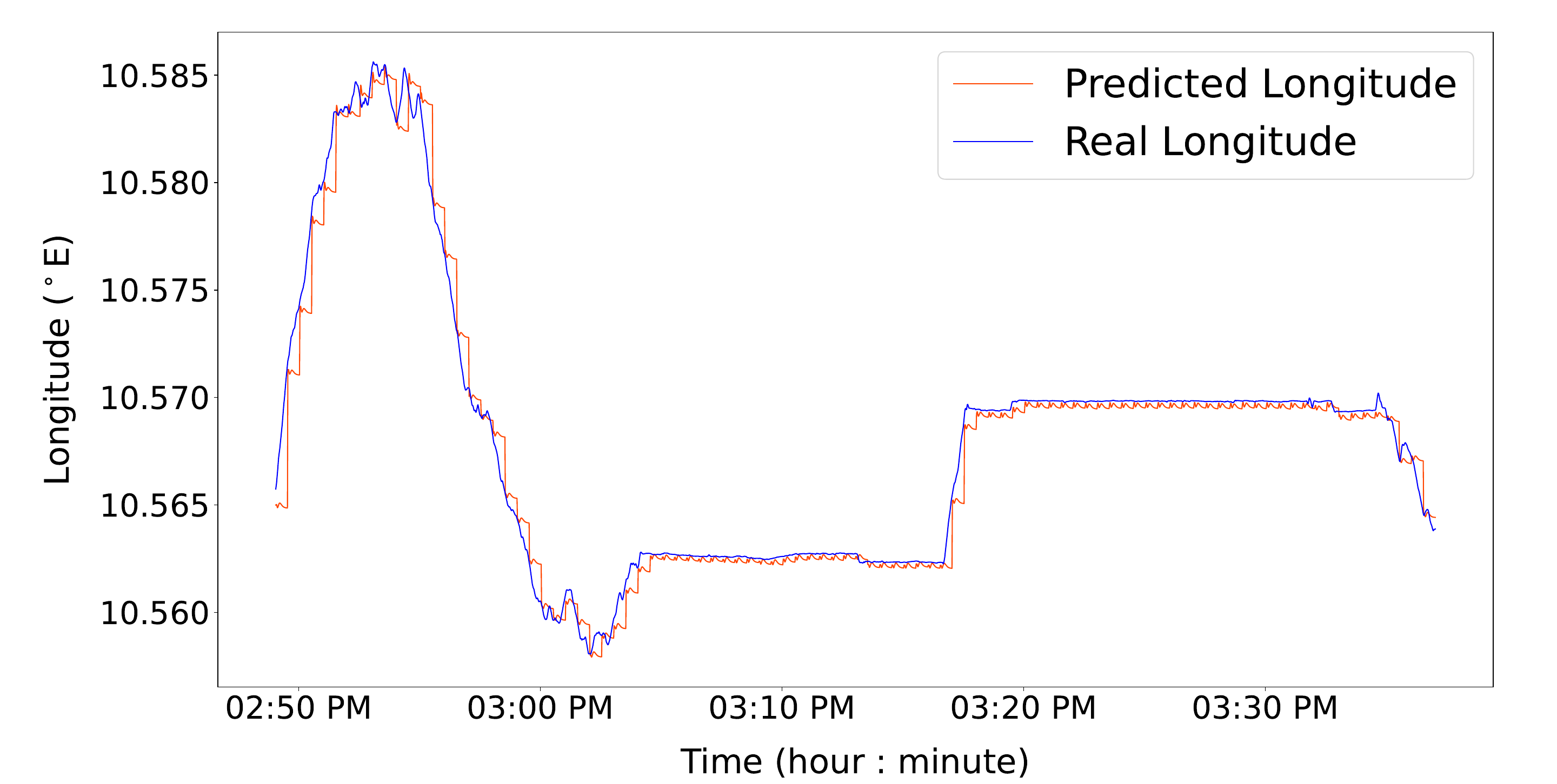}
         \caption{Longitude prediction during the fifth test data set}
         \label{ED-Long5}
     \end{subfigure}
     \caption{Longitude predicted by encoder-decoder LSTM}
     \label{encoder-Longitude}
\end{figure}

In our study, we used two separate encoder-decoder LSTM models for latitude and longitude prediction, each comprising two recurrent neural networks that work as an encoder-decoder pair. Both models had 32 neurons for the encoder and 8 neurons for the decoder. Figs. \ref{ED-Lat-T} and \ref{ED-Long-T} show the training process for latitude and longitude prediction, respectively. We evaluated the performance of the latitude prediction model on five different test data sets, as shown in Fig.\ref{encoder-Latitude}. Similarly, we evaluated the performance of the longitude prediction model on five different test data sets, which are illustrated in Fig. \ref{encoder-Longitude}. The performance of the encoder-decoder LSTM in predicting latitude is worse than that of the vanilla LSTM, while it performed better in longitude prediction. This may be because the model was able to better understand the relationship between the input data points and the output data points during training in longitude prediction.
\floatplacement{figure}{!h}
%################################################Longitude############################################
\floatplacement{figure}{!p}
%######################################################################################################

\end{document}